School of Production Engineering & Management

# Decision Making via Semi-Supervised Machine Learning Techniques

Eftychios Protopapadakis

Thesis submitted in partial fulfillment of the requirements for the degree of Doctor of Philosophy.

Technical University of Crete
University Campus, Akrotiri 73100 Chania

The School of Production Engineering and Management nominated the following persons to serve as the Doctoral Committee:

| | | | | |
|---|---|---|---|---|
| | | **Doctoral Committee** | | |
| **1** | **Nikolaos** | **Matsatsinis** | (supervisor) | |
| | Technical University of Crete | | School of Production Engineering and Management | Full Professor |
| **2** | **Anastasios** | **Doulamis** | (advisor) | |
| | National Technical University of Athens | | School of Rural and Surveying Engineering | Assistant Professor |
| **3** | **Michael** | **Doumpos** | (advisor) | |
| | Technical University of Crete | | School of Production Engineering and Management | Associate Professor |
| **4** | **Nikolaos** | **Grammalidis** | (committee member) | |
| | Center of Research and Technology Hellas | | Informatics and Telematics Institute | Senior Researcher (Grade B) |
| **5** | **Yannis** | **Marinakis** | (committee member) | |
| | Technical University of Crete | | School of Production Engineering and Management | Assistant Professor |
| **6** | **Georgios** | **Stavroulakis** | (committee member) | |
| | Technical University of Crete | | School of Production Engineering and Management | Full Professor |
| **7** | **Stelios** | **Tsafarakis** | (committee member) | |
| | Technical University of Crete | | School of Production Engineering and Management | Assistant Professor |

*Acknowledgements*

This thesis has been supported by IKY fellowships of excellence for post graduate studies in Greece-Siemens Program.

Various chapters demonstrate, partially, research outcomes, obtained during the following projects:

1. POSEIDON: Development of an Intelligent System for Coast Monitoring using Camera Arrays and Sensor Networks in the context of the inter-regional programme INTERREG (Greece-Cyprus cooperation) - contract agreement K1 3 1017/6/2011.
2. 4D-CH-World: Four Dimensional Cultural Heritage World, Marie Curie IAPP project Grant agreement number 324523.
3. ROBO-SPECT: Robotic system with intelligent vision and control for tunnel structural inspection and evaluation, European Union's Seventh Framework Programme under grant agreement no 611145.

I would, also, thank my colleges and professors in School of production engineering and management, whom fields of research provided various opportunities to validate many of this thesis ideas.

At first, I would like to thank my supervisor Prof. Nikolaos Matsatsinis, for his support, tutorship and for letting me run with my own ideas at my own pace. Many years ago, I had been taught that data are easy to obtain; extracting information, in a meaningful way, is the tricky part. I was able to realize the true meaning of this phrase after some time, during my involvement with decision systems, as a student. It was Mr. Matsatsinis research interests and knowledge, in Information and decision support systems development, which facilitated the creation of many sophisticated systems, presented in this thesis.

I would like, also, to thank Mr. Anastasios Doulamis, who introduced me to the word of machine learning, during an undergraduate course introduction to artificial intelligence, more than ten years ago. It was a whole new word of notions and mathematical formulations. At first everything appeared too complex. Yet, all this knowledge had countless application domains; especially during the development phase of any decision support system.

Defining a problem is not an easy task; there are always many variables and constraints that have to be satisfied. On top of that, there are always multiple objectives that contradict each other. The significance of trade-offs, among the objectives, became apparent. Handling such trade-offs, as well as, brief and robust formulation techniques were taught by Mr. Michael Doumpos.

Even if you manage to formulate a problem, there is no guarantee of finding the optimal solution; in that case, you should look for good ones. Thanks to Mr. Yannis Marinakis, I started utilizing genetic optimization tools, in order to find “good” solutions at complex problems. Thankfully, genetic algorithms could be applied anywhere, overriding the need of heuristic approaches.

My introduction to the multidisciplinary field of engineering, called Mechatronics, occurred back in 2008, by Mr. Georgios Stavroulakis. The necessity of synergies, among various machine learning techniques, was engraved to my mind more than seven years ago. Yet, the combination of various techniques is not an easy task; excessive testing, parameters definition and performance impact analysis is a common phenomenon. Facing such challenges force you to think outside of the box.

The involvement with large datasets, occurred while working with Mr. Stelios Tsafarakis. Missing entries exist everywhere, complicating any type of data analysis. Dealing with such issues, forced me to understanding the importance of sampling and handling the missing data.

Last, but not least, I would like to thank Mr. Nikolaos Grammalidis for his acceptance to participate in this thesis committee, despite asking him, literally, the very last moment.

*Special Thanks*

Many thanks to my family and all those who were like a family to me. The life of a student has many unexpected situations, which can become a burden, if you don’t have reliable persons by your side.

Special thanks go to my professor Mr. Anastasios Doulamis who was the “grey eminence” behind many of this thesis’ topics. His expertise and deep knowledge on multiple research fields was a great help, supporting the development of various approaches on real life application scenarios. Thanks to his active evolvement in various projects, I was able to learn, test and validate various techniques related to this thesis subjects. As a character he is a rather calm and kind person, whose advice was crucial and extremely helpful in all kind of situations, in a rather chaotic life of Ph.D. student.

The contribution of Konstantinos Makantasis was, also, extremely important. He has a solid foundation in computer vision field and an even better foundation in machine learning field. Thankfully, many of his research topics and ideas were ideal to apply semi-supervised learning approaches. It has been more than five years sharing the same office; no quarrels, no disputes. Only ideas, conferences, traveling and all kind of deadlines to catch in our daily lives.

*How to cite*

If you would like to cite this work, I recommend using the following bibtex entry:

```
@phdthesis{protopapadakis_decision_2016,
	address = {Greece},
	title = {Decision {Making} via {Semi} {Supervised} {Machine} {Learning} {Techniques}},
	school = {Technical University of Crete},
	author = {Protopapadakis, Eftychios},
	year = {2016}
}
```

This thesis research outcomes contributed to the following publications:

*Author's info*

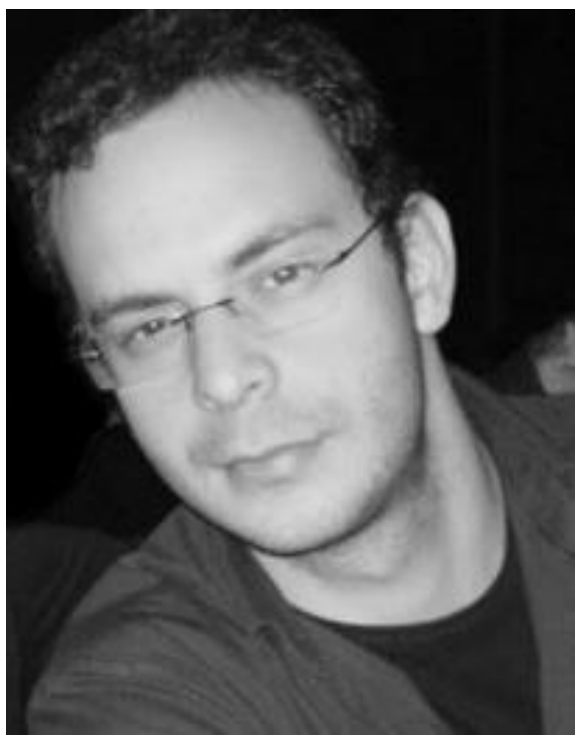

Eftychios Protopapadakis received the Diploma degree in Production Engineering and Management from Technical University of Crete in 2009. In 2011, he received his master degree in management and business administration at the same institute. Since then, he is a PhD student. He has been a research associate in Interreg and European projects since 2011. His research fields are focused on semi-supervised machine learning, decision support systems and genetic optimization, emphasizing on actual life applications.

eft<dot>protopapadakis<youknowwhat>gmail<dot>com
https://gr.linkedin.com/pub/eftychios-protopapadakis/bb/75a/427
https://www.researchgate.net/profile/Eftychios_Protopapadakis

# *Preface*

It was a master course assignment, back in 2010, that led me to the semi-supervised machine learning field. The assignment required the development of an appropriate methodology for visual surveillance of industrial assembly lines, in a user friendly way; i.e. the development of an appropriate Decision Support System (DSS), able to handle complex data, in a cost effective way, avoiding long initialization processes.

DSSs is a broad term utilized in many cases (Geertman and Stillwell, 2012; Knijnenburg et al., 2012; Martínez et al., 2010). Regardless the application scenario, the main goal is always the same: *Provide information to the user, in a meaningful way, supporting the decision making process*. DSSs exploit a variety of methods from the machine learning field. They utilize available data in order to create appropriate structures and inference mechanisms. In the end, created structures produce an outcome for specific data, at a user's request.

It is evident that available data induce DSS performance (H. Chen et al., 2012; Demirkan and Delen, 2013); *the more we have the better the performance,* assuming that we have data of good quality. Nowadays, data availability is not such a problem. However, *data abundance does not imply good features quality, neither explicit information over them*. Semi-supervised learning (SSL) deals directly with these two major disadvantages.

SSL approaches emerge naturally, as the data availability grow bigger over the years. We need approaches that utilize a small portion of data, processed by an expert, with a many times greater portion, available online. SSL exploits experts' knowledge on a minimum amount of data, minimizing both the effort and the cost. Simultaneously, it take into account the rest of the available data in order to create high quality DSSs.

In this thesis, various real life applications are presented, where SSL was necessary for the development of an appropriate DSSs. The first chapter is dedicated in SSL history, for the sake of completeness. If you are interested in the semi-supervised learning field, an excellent start would be the work of (Zhu and Goldberg, 2009a). The rest of the thesis chapters are dedicated to specific real-life application scenarios.

Each of the chapters follows a similar structure. A brief description of the problem is provided at first. Then, the related work section describes the most recent approaches on the field[1]. As such, we can identify weak points and extend current research. Finally, a mathematical formulation is formed and applied. The end of each chapter contains experimental results and conclusions.

[1] To the best of our knowledge, by the time published either as journal or conference paper.

Contents

# *Chapter* I: A Bit of History

***Quality over quantity.***

***Aristotle, Greek philosopher***

## 1 Introduction

In this section we try to grasp, briefly, the notion behind Semi-Supervised Learning (SSL). As we will see, the core idea lies in *the need for evolution, of existing techniques, in a cost effective way*. It is extremely important to handle the data abundance, in a meaningful way, in order to: (a) improve model's performance and (b) minimize experts' interventions. As such, the first steps on the SSL field begun around 1960 when scientists started exploiting new (i.e. unseen so far, unlabeled) data to advance their models' abilities. Since then, and up to date, SSL field continues to involve, motivated by the data abundance of our age.

### 1.1 Definition of semi-supervised machine learning

SSL is the machine learning task of inferring a function from labeled and unlabeled data. Thus, SSL falls between unsupervised learning, e.g. (Ranzato et al., 2007; Konstantinos Makantasis et al., 2013), and supervised learning, e.g. (Doulamis et al., 2003; Kosmopoulos et al., 2011). The main idea lies in the usefulness of unlabeled data (Seeger, 2001); when used, in conjunction with a small amount of labeled data, can considerably improve the model's performance in terms of accuracy, precision, etc.

Unlabeled data exploitation is based on various assumptions, regarding their structure and the feature space properties. There are in total three main assumptions, which can be used either alone or as a combination. Given an assumption, at least one appropriate regularizer over the unlabeled data is formed and utilized. Then, the model is trained and the overall performance is evaluated.

SSL is used, mainly, for classification (Olivier Chapelle et al., 2006; Zhu, 2005). Other approaches involve feature reduction (Cheng et al., 2008) and hybrid hashing techniques (J. Wang et al., 2012), regression (Cortes and Mohri, 2006), and clustering (Anand et al., 2014; Grira et al., 2004) problems. There are, also, cases outside previous categories, e.g. metric learning (Q. Y. Wang et al., 2012); the work of (Hoi et al., 2008) utilized SSL to appropriate calculate a distance metric, applied in an image retrieval scenario.

### 1.2 A brief history

Usage of unlabeled data, in classification problems, is documented in 1965-1970 period. These approaches were using self-training techniques (Agrawala, 1970; Fralick, 1967; Scudder, H., 1965). The concept of such approach was rather simple; train a model, use it over new data to produce results, and use the new results to further train the model. Latter appeared transductive learning (Vapnik and Chervonenkis, 1974; Vapnik and Sterin, 1977). In this case the labeling procedure occurs only on the currently available (unlabeled) data. In contrast to inductive inference, no general decision rule is inferred.

In the 1970s appeared the estimation problem of Fisher linear discriminant rule with unlabeled data (Hosmer, 1973; McLachlan, 1977; McLachlan and Ganesalingam, 1982; O'neill, 1978). The main approach involved mixture of Gaussians and the expectation maximization algorithm into an iterative algorithm (Dempster et al., 1977). The goal was the maximization of likelihood in both labeled and unlabeled data. Such approaches used one-component-per-class setting. Yet, it is possible to use multiple components (Miller and Uyar, 1996; Shahshahani and Landgrebe, 1994) or different mixture models (Cooper and Freeman, 1970).

Theoretical analysis over the SSL field first took place in 1980. The work of (Ratsaby and Venkatesh, 1995), over a mixture of two Gaussians, produced learning rates in a probably approximately correct framework. Additionally, if we have an identifiable mixture, with an infinite number of unlabeled points, the probability of error has an exponential convergence to the Bayes risk (Castelli and Cover, 1995). Theoretical analysis is still an active field (Lafferty and Wasserman, 2007; Nadler et al., 2009).

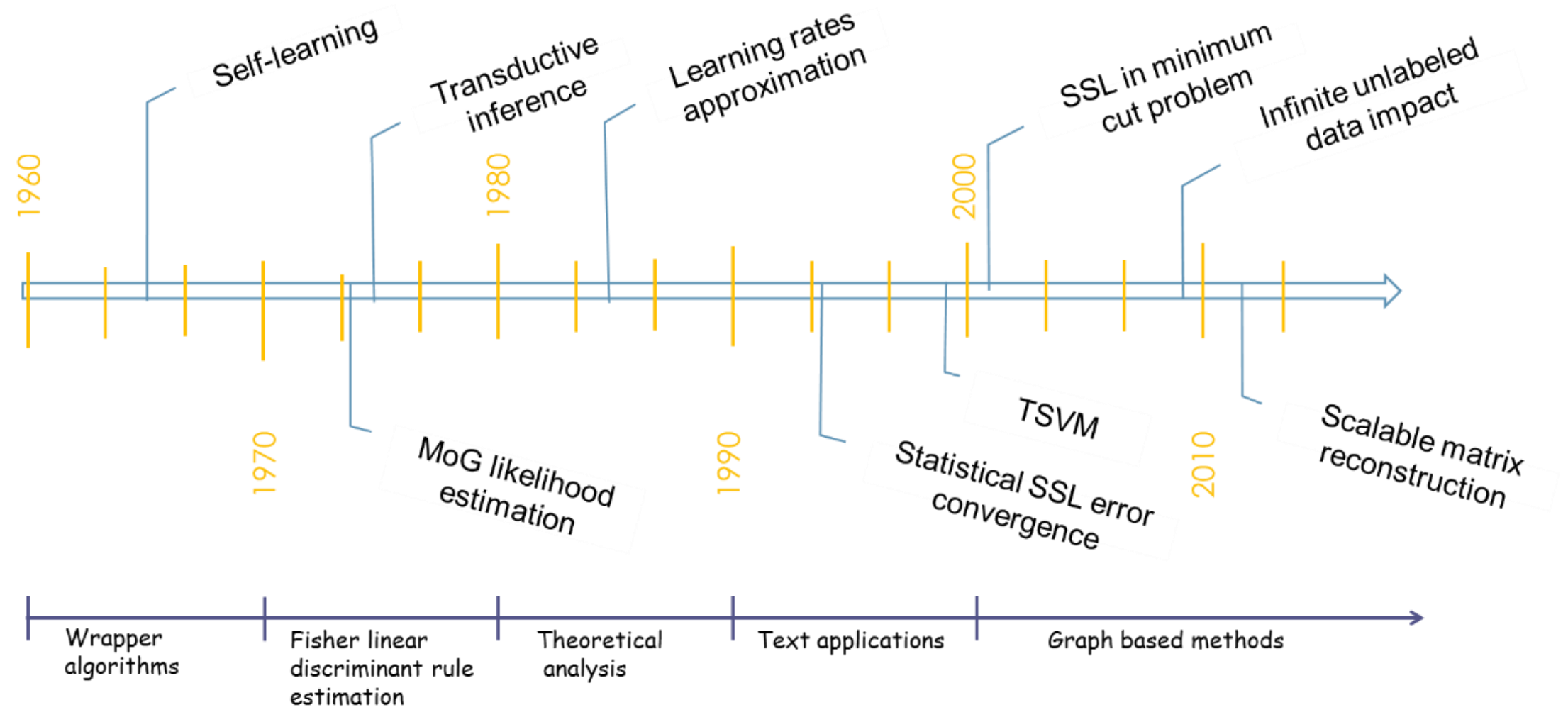

*Figure 1.1. The SSL field evolution through the years..*

After 1990, SSL has been extensively used in natural language problems (Collins and Singer, 1999; Yarowsky, 1995) and text classification (McCallum and Nigam, 1998; Nigam et al., 2000). The co-training approach was introduced by (Blum and Mitchell, 1998); the two classifiers (or hypotheses) must agree on the much larger unlabeled data as well as the labeled data. Evaluating results over generative mixture models and EM are shown in (Nigam and Ghani, 2000).

At the same time emphasis where given in discriminative decision boundaries and their placement away from dense regions. That problem led to transductive support vector machines, a method that is NP-hard. Primary researchers had focused on efficient approximation algorithms. Early algorithms (Demiriz and Bennett, 2001) (Fung & Mangasarian, 1999) either cannot handle more than a few hundred unlabeled examples, or did not do so in experiments. The SVM-light TSVM implementation (Joachims, 1999) is the first widely used software.

In the 2000s graph-based methods (Goldberg and Zhu, 2006) emerged. The data are represented by the nodes of a graph, the edges of which are labeled with the pairwise distances of the incident nodes (and a missing edge corresponds to infinite distance). The SSL concept in a graph minimum cut problem was used by (Blum and Chawla, 2001) for binary classification. Since then, research has focused on different regularizers exploitation (Niu et al., 2013) and effective weight matrix construction (Liu et al., 2010), in order to deal with scalability issues.

The last decade emphasis is given on synergies. At first the methodology of (Ratle et al., 2010) constitutes a general framework for building computationally efficient semi-supervised methods on hyperspectral image classification problems. The work of (Kingma et al., 2014) is based on deep generative models and approximate Bayesian inference, exploiting recent advances in variational methods. Deep hybrid Boltzmann machines and deep noising auto encoders are described in (Ororbia II et al., 2015).

## 1.3 Practical value of SSL

The acquisition of labeled data for a learning problem often requires a skilled human agent (e.g. to transcribe an audio segment, annotate background in an image, etc.) or a physical experiment (e.g. determining the 3D structure of a protein ,determining whether there is oil at a particular location). The cost associated with the labeling process, thus, may render a fully labeled training set infeasible, whereas acquisition of unlabeled data is relatively inexpensive. In such situations, SSL can be of great practical value[2]. SSL is also of theoretical interest in machine learning field and as a model for human learning.

One major advantage is the easy implementation on existing techniques; SSL can be directly or indirectly incorporated in any machine learning task. Semi-supervised SVMs approaches are a classical example of direct usage of SSL assumptions into the minimization function (Qi et al., 2012). Indirect utilization of SSL can be found in multi-objective optimization (MOO) frameworks (Alok et al., 2015; Cheng et al., 2012; Kobayashi et al., 2012). In MOO we have multiple fitness evaluation functions; many of them are based on SSL assumptions. Then, from a large pool of possible solution we peak those over the Pareto front. Thus, SSL is involved in the best individual selection procedure.

In real life, there are literally countless fields of testing, assuming that there is data availability. Some examples, further developed in the following chapters, are provided: The work of (E. Protopapadakis et al., 2015) evaluates the foundation piles structural situation using graph based approaches. A scalable graph based approach was used in (Makantasis et al., 2015c) for the initialization of a maritime surveillance system. The SSL cluster assumption was used in (Makantasis et al., 2015b) for the initialization of a fall detection system for elder people. A self-training approach is adopted by (Protopapadakis et al., 2012) for industrial workflow surveillance purposes in Nissan factories. In cultural heritage, SSL has been exploited by (Protopapadakis and Doulamis, 2014) in order to develop image retrieval schemes suitable to the user preferences.

[2] There are, of course, other approaches dealing with the labeling cost; active learning (Demir et al., 2014), noisy labelers (Ipeirotis et al., 2013), etc.

# *Chapter* II: The Basics

*You can't cross the sea merely by standing and staring at the water.*
*Rabindranath Tagore, Bengali polymath*

## 2 Understanding the SSL field

In this section, we move deeper into the world of SSL using mathematical notations and formulation. Additionally, we provide further information regarding the main assumptions of the field, the various regularizers and the techniques' taxonomy. In order to understand the utilized techniques and proposed approaches of this thesis, someone needs to grasp the basic formulations on the field[3]. At very fist, we start with some basic notations, most of them are in accordance with (Zhu and Goldberg, 2009a).

Let us denote as $\boldsymbol{\mathcal{X}} \in \mathbb{R}^d$ the space of input values (i.e. features originating from available data) and $\mathcal{Y} \in \mathbb{R}^m$, the space of output values. $\boldsymbol{X} = \{x_1, \dots, x_n\}$. The space of input values is divided in two sub sets $\boldsymbol{X}_L = \{x_1, \dots, x_l\}$ and $\boldsymbol{X}_U = \{x_{l+1}, \dots, x_n\}$. We also have $\boldsymbol{Y}_L \subset \boldsymbol{Y}$ available outputs. Generally, the decision mechanism is a function $f: \mathbb{R}^d \to \mathbb{R}^m$, that maps any given $x_i \in \boldsymbol{\mathcal{X}}$ to an appropriate $\widehat{y_i} \in \mathbb{R}^m$. We want $\widehat{y_i} \cong y_i, \forall i = 1, \dots, n$, i.e. the models' outputs be similar to an expert's decision. At this point, let us explain why we prefer $\widehat{y_j} \cong y_j$, rather than $\widehat{y_j} = y_j, \forall j = 1, \dots, l$. *There is always the possibility of labeling errors among the labeled data set*.

### 2.1 Inductive vs. transductive learning

Please note a very important detail at this point; SSL may refer to either transductive learning or inductive learning. If $\boldsymbol{X} \subset \boldsymbol{\mathcal{X}}$ then we may have transductive or inductive learning (depending on the method). If $\boldsymbol{X} = \boldsymbol{\mathcal{X}}$ we have transductive learning: we have a set of observations and we do not expect any more. Thus, we utilize all of them. The goal of transductive learning is to infer the correct labels for the given unlabeled data $x_{l+1}, \dots, x_{l+u}$ only (i.e. current subspace). The goal of inductive learning is to infer the correct mapping from $X$ to $Y$ (i.e. entire space $\boldsymbol{\mathcal{X}}$). Put it simply, inductive techniques can generalize to unseen data and transductive cannot. Thus, according to (Zhu and Goldberg, 2009a) we have:

**Inductive semi-supervised learning:** Given a training sample, $\{(\boldsymbol{x}_i, \boldsymbol{y}_i)\}_{i=1}^{l}, \{\boldsymbol{x}_j\}_{j=l+1}^{n}$, inductive semi-supervised learning learns a function $f: \boldsymbol{\mathcal{X}} \to \boldsymbol{\mathcal{Y}}$ so that $f$ is expected to be a good predictor on future data, $\{\boldsymbol{x}_k\}_{k=n+1}^{m}$. Like in supervised learning, one can estimate the performance on future data by using a separate test sample $\{\boldsymbol{x}_k, \boldsymbol{y}_k\}_{k=n+1}^{m}$, which is not available during training.

**Transductive learning**: Given the same training sample, as before, transductive learning trains a function $f: \boldsymbol{\mathcal{X}} \to \boldsymbol{\mathcal{Y}}$ so that $f$ is expected to be a good predictor on the unlabeled data, $\{\boldsymbol{x}_j\}_{j=l+1}^{n}$. Note $f$ is defined only on the given training sample, and is not required to make predictions outside. It is therefore a simpler function.

[3] Rest assured these formulations are repetitive and very consistent among the researchers on the SSL field. After a while, you will be able to follow such notations with ease.

## 2.2 Exploitation of unlabeled data

Unlabeled data provide further information during the models initialization phase. Available information is utilized through regularizers (Bousquet et al., 2004). The regularizer is a functional, defined by the user and the adopted SSL assumption(s). As such, there are many alternatives to be used for a given problem.

### 2.2.1 Assumptions

In order to make any use of unlabeled data, we must assume some structure to the underlying distribution of data. SSL algorithms make use of at least one of the following assumptions.

1. Smoothness assumption: Points which are close to each other are more likely to share a label. This is also generally assumed in supervised learning and yields a preference for geometrically simple decision boundaries. In the case of semi-supervised learning, the smoothness assumption additionally yields a preference for decision boundaries in low-density regions, so that there are fewer points close to each other but in different classes (Zhou et al., 2004).
2. Cluster assumption: The data tend to form discrete clusters, and points in the same cluster are more likely to share a label (although data sharing a label may be spread across multiple clusters). This is a special case of the smoothness assumption and gives rise to feature learning with clustering algorithms (Li et al., 2008).
3. Manifold assumption: The data lie approximately on a manifold of much lower dimension than the input space. In this case we can attempt to learn the manifold using both the labeled and unlabeled data to avoid the curse of dimensionality. Then learning can proceed using distances and densities defined on the manifold. Manifolds are usually estimated using graphs (Belkin and Niyogi, 2004).

### 2.2.2 Regularization

Regularization, in the fields of machine learning and inverse problems, refers to a process of introducing additional information in order to solve an ill-posed problem or to prevent overfitting. This information is usually of the form of a penalty for complexity, such as restrictions for smoothness or bounds on the vector space norm. This section focuses on gasping the regularization notion by providing few characteristic examples, exploited in SSL. Thus, depending on the assumption, we may have smoothness, cluster or manifold based (Niyogi, 2013) regularizers. There are, also, approaches that utilize multiple regularizers (Chapelle and Zien, 2004; Chen and Wang, 2011).

Smoothness regularization involves around significant changes in $f$ values, for closely located feature vectors $\boldsymbol{x}$. To put it simply, we do not want a function that does too many jumps (Belkin et al., 2004), especially in dense areas. A typical smoothness functional is the following (Belkin et al., 2004):

$$S(f) = \sum_{i \sim j} \boldsymbol{W}_{ij} \left(f_i - f_j\right)^2 \tag{2.1}$$

In eq. (2.1) the sum is taken over the adjacent vertices of a given graph. For “good” functions $f$ the functional $S(\cdot)$ takes small values. A natural extension of this idea is the cluster assumption. Cluster regularization involves around decision boundaries. The boundaries are forced away from high density regions. As such, a cluster based regularizer is also, expected to create a smooth function.

Cluster based approaches assume that the data contains clusters, which have homogeneous labels, and the unlabeled observations are used to identify these clusters. This idea can be put in practice in several ways, giving rise to various methods. The simplest is: estimate the clusters, then label each cluster uniformly. Most of these methods (Hartigan and Wong, 1979) use definition of clusters , namely the connected components of the density level sets. However, they use a parametric -usually mixture- model to estimate the underlying density which can be far from reality. An investigation on generalization error bounds has been provided by (Rigollet, 2006).

The cluster assumption can be interpreted in another way, that is, as the requirement that the decision boundary has to lie in low density regions. This interpretation has been widely used in learning since it can be used in the design of standard algorithms such as Boosting (d'Alché-Buc et al., 2002; Hertz et al., 2004) or SVM (Chapelle and Zien, 2004), which are closely related to kernel methods mentioned above. In these algorithms, a greater penalization is given to decision boundaries that cross a cluster.

Manifold regularization is a different approach from the previous two, supported by a much larger collection of algorithms. Manifold regularization involves two steps: the creation of an appropriate manifold and the use of a regularizer over it. The most common approach for manifold creation is the data adjacency graph, which can be calculated in many ways (Belkin and Niyogi, 2004; Luo et al., 2013). Any regularizer applied over the graph is a manifold regularizer.

Manifold learning has been widely used for capturing the local geometry (Fan et al., 2011) and conducting low-dimensional embedding (L. Chen et al., 2012). In manifold regularization, the data manifold is characterized by a nearest-neighbor-graph $\mathcal{W}$, which explores the geometric structure of the compact support of the marginal distribution. The Laplacian $\mathcal{L}$ of $\mathcal{W}$ and the prediction $\mathbf{f} = [f\,(x_1), \ldots, f(x_n)]$ are then formulated as a smoothness constraint $\|f\|_I^2 = \mathbf{f}^{\mathrm{T}}$. The manifold regularization framework minimizes the regularized loss:

$$\underset{f \in \mathcal{H}_k}{\operatorname{argmin}} \frac{1}{l} \sum_{i=1}^{l} L(f, x_i, y_i) + \gamma_A \|f\|_k^2 + \gamma_I \|f\|_I^2 \tag{2.2}$$

where $L$ is a predefined loss function, $k$ is the standard scalar valued kernel, i.e., $k : \mathcal{X} \times \mathcal{X} \rightarrow \mathbb{R}$, and $\mathcal{H}_k$ is the associated reproducing kernel Hilbert space (RKHS). Here, $\gamma_A$ and $\gamma_I$ are trade-off parameters to control the complexities of $f$ in the ambient space and the compact support of the marginal distribution. The representer theorem (Belkin et al., 2006) ensures the solution of eq. (2.2) takes the form $f^*(x) = \sum_{i=1}^{n} a_i k(x, x_i)$, where $a_i \in \mathbb{R}$ is a coefficient. A pair of close samples means that the corresponding conditional distributions are similar, so that the manifold regularization $\|f\|_I^2$ helps the function learning.

While many semi-supervised algorithms have been derived from this perspective and many have enjoyed empirical success, there are few theoretical analyses that characterize the class of problems on which manifold regularization approaches are likely to work (Niyogi, 2013). In particular, there is some confusion on a seemingly fundamental point. Even when the data might have a manifold structure, it is not clear whether learning the manifold is necessary for good performance (de Sousa et al., 2015).

### 2.2.3 The importance of feature selection

Feature extraction is a special form of dimensional reduction. Transforming the input data into the set of features is called feature extraction. This procedure involves reducing the amount of resources required to describe a large set of data. When performing analysis of complex data one of the major problems stems from the number of variables involved. Analysis with a large number of variables generally requires a large amount of memory and computation power or a classification algorithm which overfits the training sample and generalizes poorly to new samples. Feature extraction is a general term for methods of constructing combinations of the variables to get around these problems while still describing the data with sufficient accuracy.

In other words, bad features harm the accuracy of the model. The work of (Makantasis et al., 2015a) demonstrates that low level features are not sufficient for complex visual recognition tasks. Yet, the selection of good features does not guarantee a good performance. The work of (E. E. Protopapadakis et al., 2015) on credit risk assessment demonstrates the impact of labeled data selection. Although the features utilized were statistically significant, the random selection of few as representative ones, could jeopardize models' performance. Actually, depending on the core SSL method, a different sampling approach performed best. It is therefore crucial not only the quality of features, but also the selection of the most descriptive data, to serve as a training set.

### 2.2.4 Possible risks

SSL techniques facing various threats from the usage of unlabeled data. The possible problems depend on the technique utilized and can be categorized as theoretical foundation problems and implementation problems.

1. Model's related assumption(s) correctness. It is a common approach to make some assumptions regarding the unlabeled data distributions (especially when we are working with generative models (Vandewalle et al., 2013).
2. Degeneration in non- informant function: (Nadler et al., 2009) have shown that graph Laplacian methods (and more specific the regularization approach (Zhu, 2003) and the spectral approach (Belkin and Niyogi, 2002)) are not well posed in spaces $\mathbb{R}^d, d > 2$, and as the number of unlabeled points increases the solution degenerates to a non-informative function.
3. High dimensionality related problems: or simply stated as the curse of dimensionality. In few words, if we have a feature space of many dimensions we will face a significant decline in the performance. Let us explain that using a paradigm (Hein et al., 2005). Let $\boldsymbol{n}$ and $\boldsymbol{m}$ be points drawn from a d-dimensional Gaussian distributions, so that $\boldsymbol{n} \sim N(\mu_1, \sigma_1^2 \cdot \boldsymbol{I})$ and $\boldsymbol{m} \sim N(\mu_2, \sigma_2^2 \cdot \boldsymbol{I})$. Then their expected distance satisfies:
$$E\{\|\boldsymbol{n}-\boldsymbol{m}\|^2\} = E\left\{\sum_{i=1}^{d} |n_i - m_i|^2\right\}$$
$$= \sum_{i=1}^{d} \{Var(n_i - m_i) + E\{n_i - m_i\}^2\} \tag{2.3}$$
$$= d(\sigma_1^2 + \sigma_2^2) + \|\mu_1 - \mu_2\|^2$$
Thus, if $d$ is large, the noise term $d(\sigma_1^2 + \sigma_2^2)$ will always dominate the "informative term" $\|\mu_1 - \mu_2\|^2$; i.e. the model will not perform well.
4. Scalability: most SSL methods scale badly with the data size $n$. The classical TSVM (Joachims, 1999) scales exponentially with $n$. CCCP-TSVM approach (Collobert et al., 2006) has the lowest complexity, but it scales as at least $O(n^2)$. Graph-based SSL usually has a cubic time complexity $O(n^3)$ since the inverse of the $n \times n$ graph Laplacian is needed, thus blocking widespread applicability to real-life problems that encounter growing amounts of unlabeled data. To temper the cubic time complexity, recent studies seek to reduce the intensive computation upon the graph Laplacian manipulation or deal with other possible issues (Delalleau et al., 2005; Fergus et al., 2009; Karlen et al., 2008; Tsang and Kwok, 2006; Zhu and Lafferty, 2005). However, these methods require filed knowledge or make assumptions that is not always applicable.

## 2.3 Taxonomy of techniques

Starting from early 1960 and up to date, many techniques were conceived, evaluated and improved. It is most fortunate that the majority of them can be categorized using some basic filters: learning inference and model basis. In the following we present a brief taxonomy of these techniques. If necessary, we will provide further mathematical notations and formulations.

### 2.3.1 Generative models

Models that randomly generate observable data, typically given some hidden parameters, are called generative. Such model specifies a joint probability distribution over observation and label sequences, $P(\boldsymbol{x}, y)$. Generative models are used in machine learning for either modeling data directly (i.e., modeling observations drawn from a probability density function), or as an intermediate step to forming a conditional probability density function. Recent examples, in the fields of computer vision and text analysis, is the work of (Beecks et al., 2011; Kang et al., 2012; Lücke and Eggert, 2010; Rauschert and Collins, 2012; F. Zhuang et al., 2012).

Generative models recognize the semi-supervised learning problem as a specialized missing data imputation task for the classification problem. Existing generative approaches based on models such as Gaussian mixture or hidden Markov models (Zhu and Ghahramani, 2002), have not been very successful due to the need for a large number of mixtures components or states to perform well. Yet, there is also evidence that generative models can converge faster than discriminative, as shown by (Ng and Jordan, 2002)Ng and Jordan (2002), and so are valuable when dealing with small data sets.

Assuming that there is a data set $\boldsymbol{X}_L = \{(\boldsymbol{x}_1, \boldsymbol{y}_1), \dots, (\boldsymbol{x}_l, \boldsymbol{y}_l)\}$, we wish to fit a model parameterized by some set of parameters $\theta$ to the set's distribution, using maximum likelihood method:

$$\theta^* = arg\max_{\theta} \sum_{i=1}^{l} \log\big(P(\boldsymbol{x}_i, y_i|\theta)\big) \tag{2.4}$$

However, there is no guarantee that eq. (2.4) produces a good solution; i.e. which generalize well to unseen data, $\boldsymbol{X}_U = \{\boldsymbol{x}_{l+1}, \dots, \boldsymbol{x}_{l+u}\}$, especially if the model is rich or the feature space $\boldsymbol{\mathcal{X}}$ high dimensional (Fox-Roberts and Rosten, 2014). Thus, incorporating the unlabeled data, we have the following equation:

$$\theta_S^* = arg\max_{\theta} \left[\sum_{i=1}^{l} \log\big(P(\boldsymbol{x}_i, y_i|\theta)\big) + \sum_{i=n+1}^{l+u} \log\big(P(\boldsymbol{x}_i|\theta)\big)\right] \tag{2.5}$$

Yet, eq. (2.5) has proven to give mixed results, sometime improving model fitting, other times worsening it. Various solutions have used non-parametric density models, either based on trees (Kemp et al., 2004) or Gaussian processes (Adams and Ghahramani, 2009), but scalability and accurate inference for these approaches is still lacking. Variational approximations for semi-supervised clustering have also been explored previously (Li et al., 2009; Wang et al., 2009). In order to guarantee the decrease of the misclassification risk the distribution should be identifiable (Castelli and Cover, 1996).

The asymptotic behavior of semi-supervised learning where the model is miss-specified has been further studied by (O. Chapelle et al., 2006), where no assumptions are made about the parametric model being close to the underlying distribution. In particular, they show that the limiting value of the optimum parameters, when performing ML semi-supervised learning in such a scenario are:

$$\theta^* = arg\max_{\theta}\big[(1-\lambda)E_{P(X,Y)}(\log P(\boldsymbol{x}, y|\theta)) + \lambda E_{P(x)}\big] \tag{2.6}$$

where $\lambda$ is the probability of a sample being unlabeled and $E(\cdot)$ is a combination of objective functions. If $\lambda$ varies (say by adding unlabeled samples) then this will likely change the optimal parameters $\theta^*$, and so the associated error rate. In the limit, as $\lambda \to 1$, we will tend towards the solution found training entirely on unlabeled data. They argue that with a few assumptions on the modelling densities, $\theta^*$ is a continuous function of $\lambda$. They also show that an instance where the asymptotically optimal parameters are not changed by $\lambda$ comes, as might be expected, when the model is "correct" and can be fitted exactly to the underlying distribution; i.e., the true distribution $P(\boldsymbol{x}, y)$ is a member of the family of distributions that can be modelled by $P(\boldsymbol{x}, y|\theta)$.

### 2.3.2 Self-training and multi-view learning

In self-training a model is first trained with the small amount of labeled data and then is used to classify the unlabeled data. Typically the most confident unlabeled points, together with their predicted labels, are added to the training set. The classifier is re-trained and the procedure repeated by using its own predictions. Mathematical foundation on convergence was given for specific learners in (Culp and Michailidis, 2008; Haffari and Sarkar, 2012).

Self-training has been applied to several natural language processing tasks: word sense disambiguation (Yarowsky, 1995), identification of subjective nouns (Riloff et al., 2003), Self-training has also been applied to parsing and machine translation as shown in work of (Rosenberg et al., 2005) who applied self-training to object detection systems from images. The work of (Protopapadakis et al., 2012), on industrial surveillance, utilized a similarity based mechanism to refine self-trained classifier's outputs.

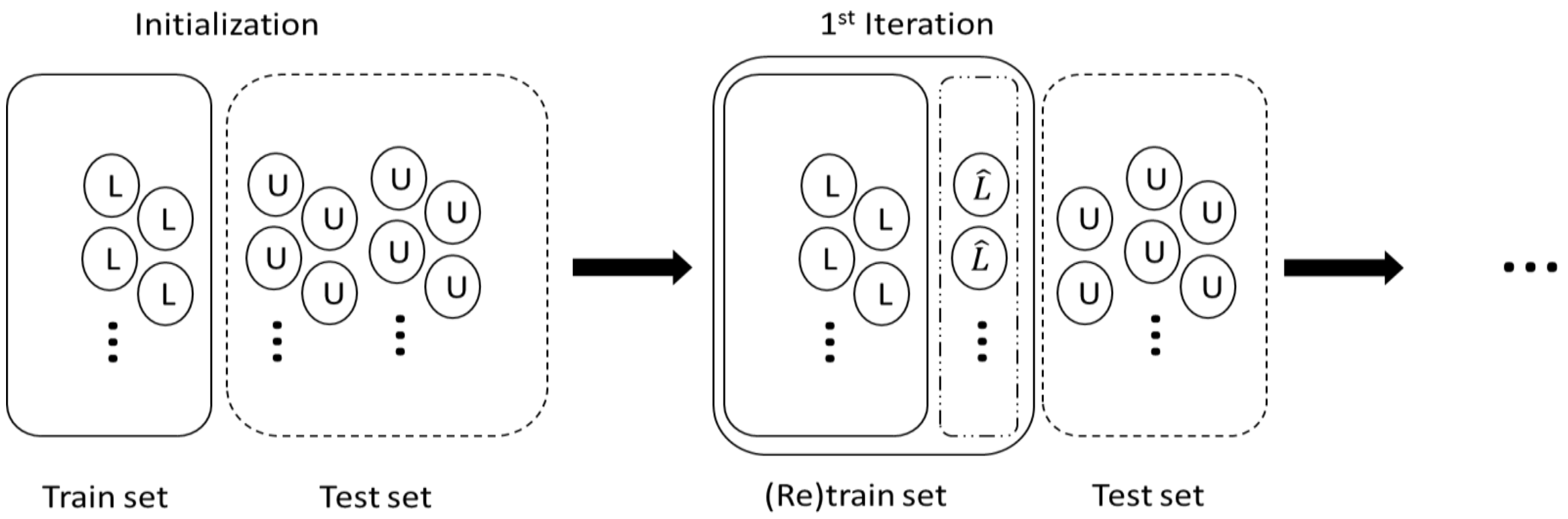

*Figure 2.1. A typical illustration of the self-training approach. At first the model is trained using only the few labeled data. Then, it actuates over the unlabeled ones, selects the most confident predictions and retrains itself. That procedure repeats until certain criteria are met.*

The demand for redundant views of the same input data is a major difference between multi-view and single-view learning algorithms. Thanks to these multiple views, the learning task can be conducted with abundant information. However if the learning method is unable to cope appropriately with multiple views, these views may even degrade the performance of multi-view learning. Through fully considering the relationships between multiple views, several successful multi-view learning techniques have been proposed (Xu et al., 2013). many of them have been modified in order to use SSL assumptions (Sun, 2013). A very common approach in Multi-view learning is co-training.

Co-training has been proposed by (Blum and Mitchell, 1998), in which the description of each samples can be partitioned into two distinct views. The basic idea is to train two learners separately on each view, and then each learner predicts unlabeled samples to enlarge the training set for the other. However, there is a drawback in co-training algorithms: the repeated loop of learning process reduces the learning speed and reinforces the error. Co-training makes a strong assumption that the two feature sets involved should be conditionally independent given a class. Although some studies have been done to relax this assumption with weaker ones (Balcan et al., 2004), a basic intuitive requirement, that the involved classifiers should be different enough from each other, should be met so that they can complement each other (S. Li et al., 2011).

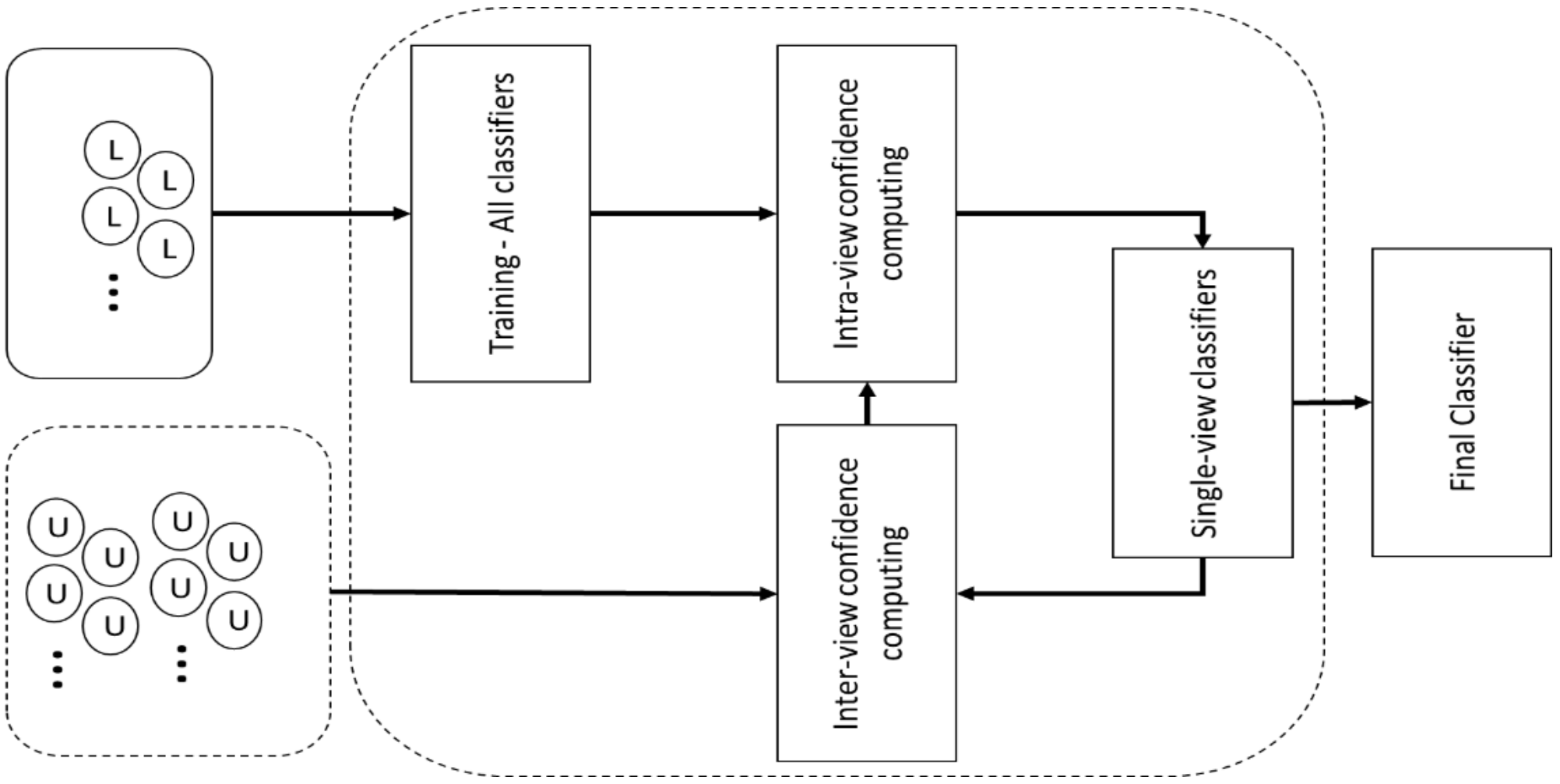

*Figure 2.2. Illustration of the multi-view training. Over the initial set of labeled data various classifiers are formed. Their performance is evaluated for a sub set of features, in both labeled and unlabeled data. Then, what appears to be a robust decision for one of the classifiers, is used to further train the rest.*

How to select basic learners is extremely important in co-training (Li et al., 2013). Extreme Learning Machine (ELM) is a new supervised learning method, proposed by (Huang et al., 2006) for single-hidden layer feed forward networks. ELM has received considerable attention in computational intelligence and machine learning communities, and a few variants have been proposed, e.g., fully complex ELM, online sequential ELM, incremental ELM and ELM ensembles (Huang et al., 2011). Some researchers have paid attention to semi-supervised ELM and there have been appeared graph-based semi-supervised ELM method (J. Liu et al., 2011) and ternary reversible Extreme Learning Machines (Tang and Han, 2009).

### 2.3.3 Low-Density separation

The low-density separation assumption pushes the decision boundary in regions where there are few data points (labeled or unlabeled). The most common approach to achieving this goal is to use a maximum margin algorithm such as support vector machines. The method of maximizing the margin for unlabeled as well as labeled points is called the transductive SVM (TSVM). However, the corresponding problem is non-convex and thus difficult to optimize (Singla et al., 2014).

The training process is actually an iterative algorithm (Sindhwani and Keerthi, 2006). Starting from the SVM solution as trained on the labeled data only, the unlabeled points are labeled by SVM predictions, and the SVM is retrained on all points. This is iterated while the weight of the unlabeled points is slowly increased. Another interesting fact is that, despite the name, TSVM are used for inductive reasoning (Pang and Kasabov, 2004). They can handle unseen data because they are defined over the whole problem space (Qi et al., 2012). An illustration of the $S^3VM$ is shown in Figure 2.3.

Low density separation (LDS) is a combination of transductive SVMs (Bruzzone et al., 2006), trained using gradient descend, and traditional SVMs using appropriate kernel defined over a graph using SSL assumptions (Chapelle and Zien, 2004). Similar to the SVM approach the TSVM need to maximize then margin (i.e. the minimum distance between a hyperplane and the closest example vectors in $\boldsymbol{X}$. Thus the following formulation is adopted:

$$\min_{w,b}\left[\sum_{i=1}^{l}\max(1-y_i(\boldsymbol{w}^T\boldsymbol{x}_i+b),0)+\lambda_1\|\boldsymbol{w}\|+\lambda_2\sum_{j=l+1}^{l+u}\max\big(1-\big(\boldsymbol{w}^T\boldsymbol{x}_j+b\big),0\big)\right] \tag{2.7}$$

where $\boldsymbol{w} \in \mathbb{R}^n$is the parameter vector that specifies the orientation and scale of the decision boundary and $b \in \mathbb{R}$ is an offset parameter. The above formulation exploits both labeld, $\boldsymbol{X}_L$, and unlabeled $\boldsymbol{X}_U$ data, leading to a non-convex optimization problem.

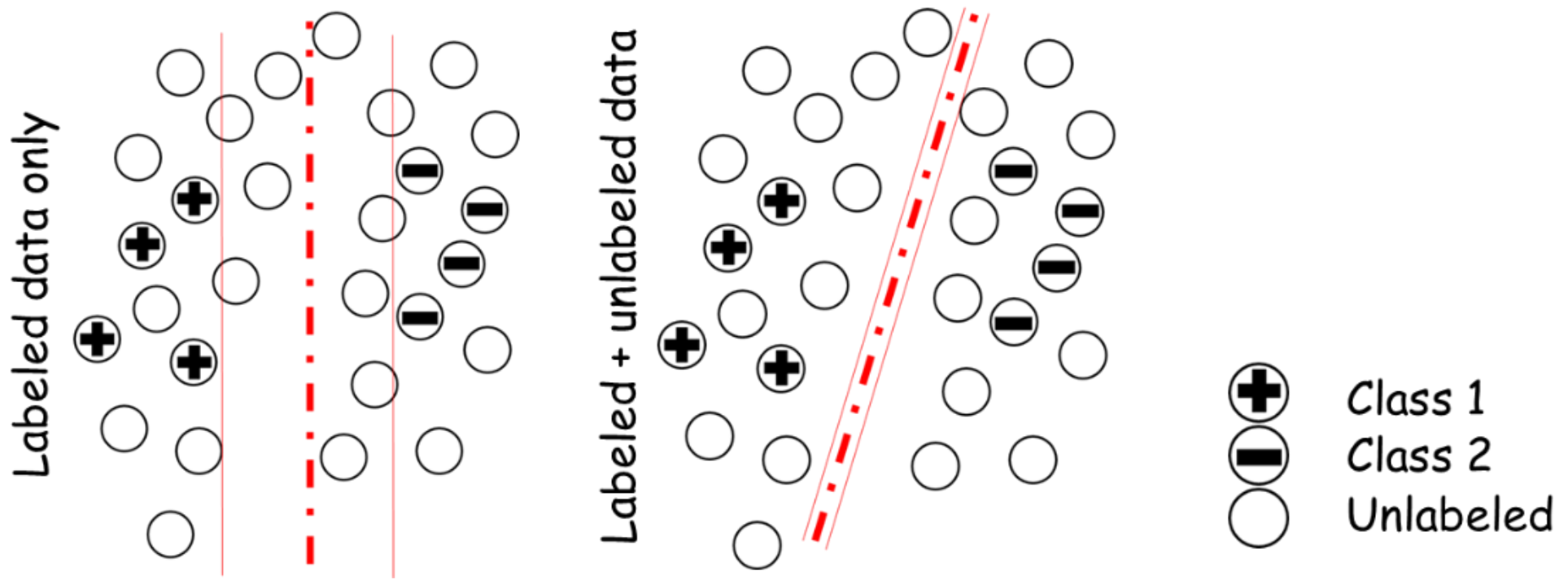

*Figure 2.3. The impact of unlabeled data in the creation of the decision boundaries. Unlabeled data guide the plane towards non dense areas, if possible. As such, there are few, if any, points between the margins of the separation boundaries (image right), in contrast to the traditional SVM approach (left image).*

The problem can be rewritten in the following form, in order to perform a standard gradient based approach:

$$\min_{\boldsymbol{w},b} \left[ \frac{1}{2}\boldsymbol{w}^2 + C\sum_{i=1}^{l} L^2\big(y_i(\boldsymbol{w}^T\boldsymbol{x}_i + b)\big) + C^* \sum_{j=l+1}^{l+u} L^*\big(\left|\boldsymbol{w}^T\boldsymbol{x}_j + b\right|\big) \right] \tag{2.8}$$

where $L(t) = \max(0,1-t)$ and $L^*(t) = \exp(-3t^2)$.

Such formulation allow the use of a non-linear kernel (i.e. a mapping procedure to a different space). The kernel creation requires the computation of the $\rho$-distances. Then a fully connected matrix,$\boldsymbol{W}$, is formed as $w_{ij} = \exp\big(\rho - dist(i,j)\big) - 1$. Dijkstra's algorithm (Dijkstra, 1959) is employed to compute the shortest path lengths, $d_{SP}(i,j)$ for all pairs of points. The matrix $\boldsymbol{\mathcal{D}}$ of squared $\rho$ -path distances is calculated, for all pairs of points, as:

$$\mathcal{D}_{ij} = \left(\frac{1}{\rho}\log\big(1 + d_{SP}(i,j)\big)\right)^2 \tag{2.9}$$

The final step towards the kernel's creation, involves multidimensional scaling (Cox and Cox, 2008), or MDS, to find a Euclidean embedding of $\boldsymbol{\mathcal{D}}^\rho$(in order to obtain a positive definite kernel). The embedding found by the classical MDS are the eigenvectors corresponding to the positive eigenvalues $\boldsymbol{U\Lambda U}^T = -\boldsymbol{H\mathcal{D}}^\rho\boldsymbol{H}$, where $H_{ij} = \delta_{ij} - \frac{1}{l+u}$. The final representation of $\boldsymbol{x}_i$ is $x_{ik} = U_{ik}\sqrt{\lambda_k}, 1 \le k \le p$.

### 2.3.4 Graph based methods

Graph-based semi-supervised methods define a graph over the entire data set, $\boldsymbol{X} = \boldsymbol{X}_L \cup \boldsymbol{X}_U$, where, $\boldsymbol{X}_L = \{(\boldsymbol{x}_1, \boldsymbol{y}_1), \dots, (\boldsymbol{x}_l, \boldsymbol{y}_l)\}$, is the labeled data set and $\boldsymbol{X}_U = \{\boldsymbol{x}_{l+1}, \dots, \boldsymbol{x}_{l+u}\}$ the unlabeled data set. Feature vectors, $\boldsymbol{x}_i \in \mathbb{R}^m, i = 1, \dots, l+u$, are available for all the observations and $\boldsymbol{y}_i \in \mathbb{R}^k, i = 1, \dots, l$, are the corresponding classes of the labeled ones, in a vector form; $k$ denotes the available classes. The nodes represent the labeled and unlabeled, examples in the dataset; edges reflect the similarity among examples. These methods usually assume label smoothness over the graph. That is, if two instances are connected by a strong edge, their labels tend to be the same.

Graph methods are non-parametric (i.e., number of parameters grows with data size), discriminative, and transductive in nature. Intuitively speaking, in a graph that various data points are connected, greater the similarity greater the probability of having similar labels. Thus, the information (of labels) propagates from the labeled points to the unlabeled ones. Researchers utilized such methods because some datasets are naturally represented by a graph (e.g. web, citation networks, and social networks). An illustration of data connection in graphs is shown in Figure 2.4.

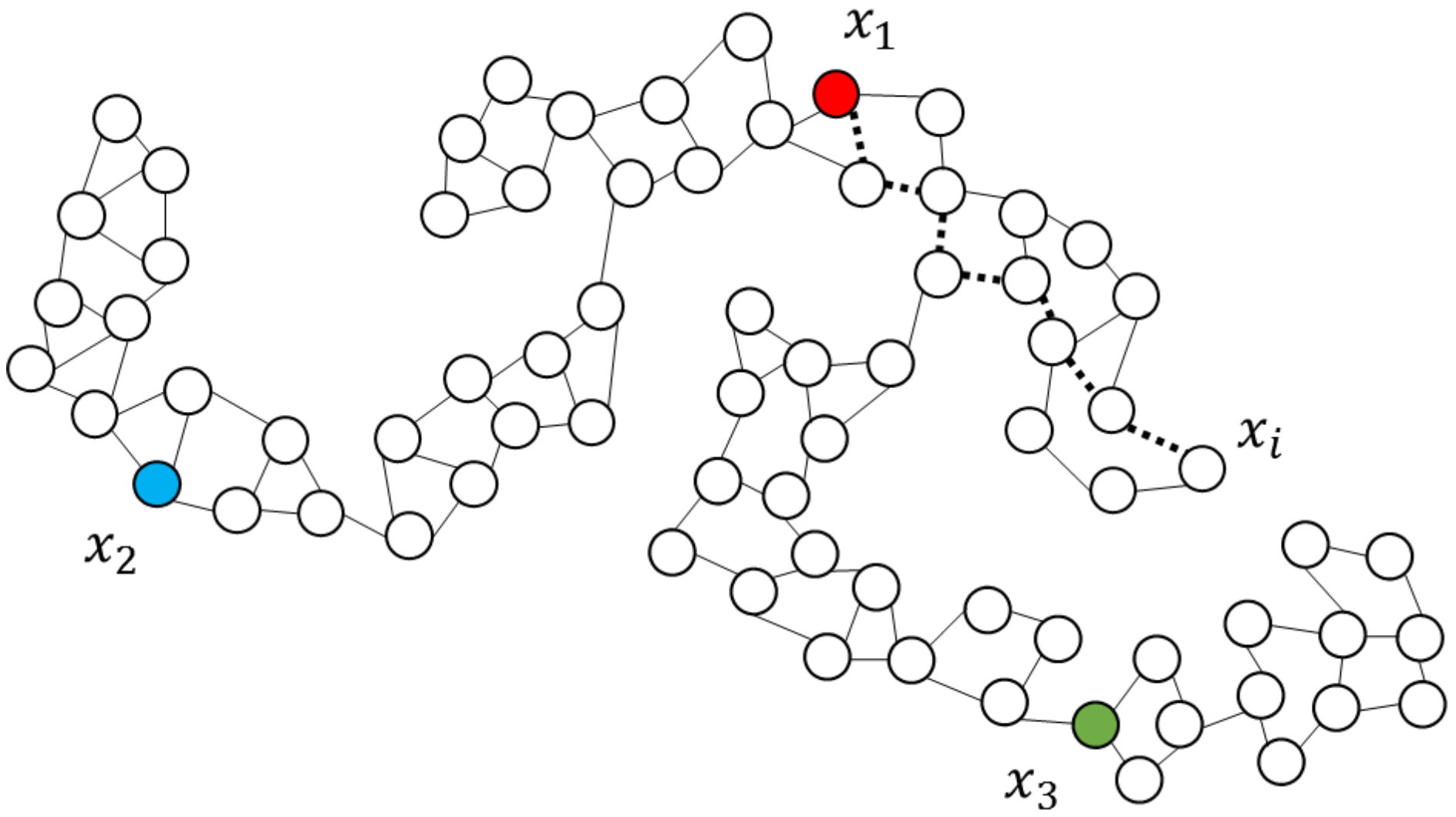

*Figure 2.4. A graph constructed from labeled instances $x_1$, $x_2$, $x_3$ and unlabeled instances. The label of unlabeled instance $x_i$ will be affected more by the label of $x_1$, which is closer in the graph, than by the labels of $x_2$ or $x_3$, which are farther in the graph, even though $x_3$ is closer in Euclidean distance.*

An indicative paradigm of graph based SSL is the harmonic function approach (Zhu, 2003). This approach estimates a function $f$ on the graph which satisfies two conditions. Firstly, $f$ has the same values as given labels on the labeled data, i.e. $f(\boldsymbol{x}_i) = \boldsymbol{y}_i, i = 1, \dots, l$. Secondly, $f$ satisfies the weighted average property on the unlabeled data:

$$f(\boldsymbol{x}_j) = \frac{\sum_{k=1}^{l+u} w_{jk} f(\boldsymbol{x}_j)}{\sum_{k=1}^{l+u} w_{jk}}, j = l+1, \dots, l+u \tag{2.10}$$

where $w_{ij}$ denotes the edge weight. Those two conditions lead to the following problem:

$$\min_{f:f(\boldsymbol{x})\in\mathbb{R}} \sum_{i,j=1}^{l+u} w_{ij} \left(f(\boldsymbol{x}_i) - f(\boldsymbol{x}_j)\right)^2 \\ s.t.\, f(\boldsymbol{x}_i) = \boldsymbol{y}_i, i = 1, \dots, l \tag{2.11}$$

In the following, we introduce some notations in order to present the close form solution of (2.11). Let $\boldsymbol{W}$ be an $(l + u) \times (l + u)$ weight matrix, whose $i, j$ -th element is the edge weight $w_{ij}$. Let $D_{ii} = \sum_{j=1}^{l+u} w_{ij}$ be the weighted degree of vertex i, i.e., the sum of edge weights connected to $i$. Then we create a diagonal matrix $\boldsymbol{D} \in \mathbb{R}^{(l+u)\times(l+u)}$ by placing $D_{ii}$ on the diagonal. The unnormalized graph Laplacian matrix $\boldsymbol{L}$ is defined as: $\boldsymbol{L} = \boldsymbol{D} - \boldsymbol{W}$. Matrix $\boldsymbol{L}$ is rearranged in the form:

$$\boldsymbol{L} = \begin{bmatrix} \boldsymbol{L}_{ll} & \boldsymbol{L}_{lu} \\ \boldsymbol{L}_{ul} & \boldsymbol{L}_{uu} \end{bmatrix} \tag{2.12}$$

Let $\mathbf{f} = \left(f(x_1), \dots, f(x_{l+u})\right)^T$ be the vector of $f$ values on all vertices arranged in a way that $\mathbf{f} = (\mathbf{f}_\mathrm{l}, \mathbf{f}_\mathrm{u})$ and let $\boldsymbol{Y}_l = (\boldsymbol{y}_1, \dots, \boldsymbol{y}_l)^T$. The harmonic solution is:

$$\mathbf{f}_\mathrm{l} = \boldsymbol{Y}_l \tag{2.13}$$

$$\mathbf{f}_{\mathrm{u}} = \boldsymbol{L}_{uu}^{-1}\boldsymbol{L}_{ul}\boldsymbol{Y}_{l}$$

Thus, we are able to estimate (soft label) output vectors for all the edges of the graph. Each labeled edge, $i$, is guaranteed to have the output vector, $\boldsymbol{y}_l$, as it was provided by the expert.

*2.3.4.1 Points of interest*

In graph based approaches, weight matrix, $\boldsymbol{W}$, has to be well-defined so that the graph Laplacian matrix, $\boldsymbol{L}$, will be invertible. In order to create an appropriate weigh matrix we first need to define the graph among the available points. The following heuristic approaches have been used extensively:

1. **Fully connected graph**, where every pair of vertices $\boldsymbol{x}_i, \boldsymbol{x}_j$ is connected by an edge. The edge weight decreases as the Euclidean distance $\|\boldsymbol{x}_i - \boldsymbol{x}_j\|$ increases. One popular weight function is: $w_{ij} = \exp\left(-\frac{\|x_i - x_j\|^2}{2\sigma^2}\right)$ where $\sigma$ is known as the bandwidth parameter and controls how quickly the weight decreases. This weight has the same form as a Gaussian function. It is also called a Gaussian kernel or a Radial Basis Function (RBF) kernel. The weight is 1 when $\boldsymbol{x}_i = \boldsymbol{x}_j$, and 0 when $\|\boldsymbol{x}_i - \boldsymbol{x}_j\|$ approaches infinity.
2. $\boldsymbol{k}$**NN graph**. Each vertex defines its $k$ nearest neighbor vertices in some distance. Note if $\boldsymbol{x}_i$ is among $\boldsymbol{x}_j$ 's kNN, the reverse is not necessarily true: $\boldsymbol{x}_j$ may not be among $\boldsymbol{x}_i$ 's kNN. We connect $\boldsymbol{x}_i, \boldsymbol{x}_j$ if one of them is among the other's kNN. This means that a vertex may have more than $k$ edges. If $\boldsymbol{x}_i, \boldsymbol{x}_j$ are connected, the edge weight $w_{ij}$ is either the constant 1 (i.e. unweighted graph), or a function of a distance (e.g. RBF kernel). If $\boldsymbol{x}_i, \boldsymbol{x}_j$ are not connected, $w_{ij} = 0$.
3. $\boldsymbol{\epsilon}$**NN** graph. We connect $\boldsymbol{x}_i, \boldsymbol{x}_j$ if $\boldsymbol{x}_i - \boldsymbol{x}_j \leq \epsilon$. The edges can either be unweighted or weighted. If $\boldsymbol{x}_i, \boldsymbol{x}_j$ are not connected, $w_{ij} = 0$. Generally, $\epsilon$NN graphs are easier to construct than kNN graphs.

A comparative study in graph creation can be found in (L. Zhuang et al., 2012). There is no dominant methodology. Empirically talking, the matrix creation approach depends on the given problem and the computational cost. Usually graph construction employs the Euclidean distance. Also, keep in mind that $k$ should be small (e.g. 3 or 5). Otherwise, we will have a bad estimation of the manifold and the label propagation will cause many errors. $k$NN graph automatically adapts to the density of instances in feature space: in a dense region, the kNN neighborhood radius will be small; in a sparse region, the radius will be large.

Another point of interesting is the nature of the graph based approaches. Thus we have two major issues:

1. Data correctness; what if some labeled data are by mistake misclassified? That is a common case, especially when we have manually annotation of image collections; e.g. as in (Makantasis et al., 2015c).
2. New (unseen) data handling; how can we handle new data, without having to recreate the entire graph each time?

Those two issues led to manifold regularization approach; we can have an inductive learning algorithm defining $f$ in the whole feature space: $f: \mathcal{X} \to \mathbb{R}$, allowing $f(x_i) \neq y_i$, $i = 1, \ldots, l$ in some cases; we may skip any constraints and try to optimize a generalized (manifold) problem of the form:

$$\min_{f:\mathcal{X}\to\mathbb{R}} \left( \sum_{i=1}^{l} c(f(x_i), y_i) + \lambda_1 \|f\|^2 + \lambda_2 \boldsymbol{f}^T \boldsymbol{L} \boldsymbol{f} \right) \quad (2.14)$$

where $c(f(x), y)$ is a convex loss function, $\|f\|^2 = \int_{x\in\mathcal{X}} f(x)^2 \, dx$ is a regulization term over the entire space $\mathcal{X}$, and $\lambda_2 \boldsymbol{f}^T \boldsymbol{L} \boldsymbol{f}$ is the traditional smoothness regularizer over the existing manifold. Equation (2.14) is just a specific case of eq. (2.2).

# *Chapter* III: Techniques of Reference

*To know, is to know that you know nothing. That is the meaning of true knowledge.*

*Socrates, classical Greek philosopher*

## 3 Other machine learning, clustering & sampling approaches

This thesis is application oriented; the SSL approaches are utilized synergistically to other machine learning approaches. Thus, for the sake of completeness, various machine learning techniques are briefly described in this section. Additionally, various sampling schemes are explained. Sampling is used in all SSL methodologies, since all techniques require a small data set of labelled instances; rather than providing a random selection we employed various sampling schemes; their core mechanisms are also explained here. Finally, most of the case studies correspond to a classification problem. Consequently, an analytic description of the traditional performance metrics is provided.

### 3.1 Machine learning techniques

In this section we will provide a brief description of various supervised/ unsupervised machine learning techniques. These techniques are utilized, primary, for performance comparisons through this thesis. They were also utilized in conjunction with SSL approaches for the creation of advanced DSS, depending on the application scenario.

#### 3.1.1 Linear Regression

Linear regression (*LinReg*) analysis is the study of linear, additive relationships between variables. Let $y_i$ denote the "dependent" variable whose values you wish to predict, and let $\boldsymbol{x}_i = [x_1, \dots, x_m]$, $i = 1, \dots, n$ denote the "independent" variables from which we wish to predict $y_i$. Then the equation for computing the predicted value of $y_i$ is:

$$\hat{y}_i = b_0 + \sum_{j=1}^{m} b_j x_{i,j} \tag{3.1}$$

This formula has the property that the prediction for $y_i$ is a straight-line function of each of the $x_{i,j}$ variables, holding the others fixed, and the contributions of different variables to the predictions are additive. The slopes of their individual straight-line relationships with $y_i$ are the constants $b_1, \dots, b_k$, the so-called coefficients of the variables. That is, $b_i$ is the change in the predicted value of $\hat{y}_i$ per unit of change in $x_{i,j}$, other things being equal. The additional constant $b_0$, the so-called intercept, is the prediction that the model would make if $\boldsymbol{x}_i = 0$ (if that is possible). The coefficients and intercept are estimated by least squares, i.e., setting them equal to the unique values that minimize the sum of squared errors within the sample of data to which the model is fitted.

#### 3.1.2 $k$ nearest neighbors

In pattern recognition, the $k$-nearest neighbors ($k$nn) algorithm is a non-parametric method used for classification (Bhatia and Vandana, 2010). Input consists of the $k$ closest training examples in the feature space. Output is a class membership. An object is classified by a majority vote of its neighbors, with the object being assigned to the class most common among its $k$ nearest neighbors (k is a positive integer, typically small). If $k = 1$, then the object is simply assigned to the class of that single nearest neighbor. $k$nn is a type of instance-based learning, or lazy learning, where the function is only approximated locally and all computation is deferred until classification.

With previously labeled samples as the training set $S$, the $k$nn algorithm constructs a local subregion $R(\boldsymbol{x}) \subseteq R^{m\times m}$ of the input space, which is situated at the estimation point $\boldsymbol{x}$. The predicting region $R(\boldsymbol{x}_i)$, which contains the closest $k$ training points to $\boldsymbol{x}_i$, is written as follows:

$$R(\boldsymbol{x}_i) = \{\widehat{\boldsymbol{x}} | d(\boldsymbol{x}_i, \widehat{\boldsymbol{x}}\,) \leq d_{thrs}\} \tag{3.2}$$

where $d_{thrs}$ is a predefined threshold. Given all points $\widehat{\boldsymbol{x}}_i \in R(\boldsymbol{x})$, $i = 1, \dots, k$ and their corresponding outputs $\hat{y}_i$, point $\boldsymbol{x}_i$ is assigned with classification label $y$ that has smallest expected misclassification cost among the values $\hat{y}_i$.

### 3.1.3 Decision trees

Decision tree learning uses a decision tree as a predictive model which maps observations about an item to conclusions about the item's target value. In classification tree structures, leaves represent class labels and branches represent conjunctions of features that lead to those class labels. Each internal (non-leaf) node is labeled with an input feature. The arcs coming from a node labeled with a feature are labeled with each of the possible values of the feature. Each leaf of the tree is labeled with a class or a probability distribution over the classes.

Algorithms for constructing decision trees usually work top-down, by choosing a variable at each step that best splits the set of items (Rokach and Maimon, 2005). Different algorithms use different metrics for measuring "best". These generally measure the homogeneity of the target variable within the subsets. These metrics are applied to each candidate subset, and the resulting values are combined (e.g., averaged) to provide a measure of the quality of the split.

Given a set of items, suppose $i \in \{1, \dots, m\}$, and $p_i$ the portion of the items labeled with $i$ among the $m$ alternatives. The most common algorithm for split evaluation is Gini impurity:

$$I_G = 1 - \sum_{i=1}^{m} p_i^2 \tag{3.3}$$

### 3.1.4 Adaptive boosting

Adaptive boosting (*AdaBoost*) is an ensemble learning algorithm, which is more resistant to over-fitting, but it is often sensitive to noisy data and outliers (Woźniak et al., 2014). AdaBoost creates a strong learner (a classifier that is well-correlated to the true classifier) by iteratively adding weak learners (a classifier that is only slightly correlated to the true classifier). During each round of training, a new weak learner is added to the ensemble and a weighting vector is adjusted to focus on examples that were misclassified in previous rounds. The result is a classifier that has higher accuracy than the weak learners' classifiers. A boost classifier is a classifier in the form:

$$H_T(\boldsymbol{x}_i) = \sum_{t=1}^{T} h_t(\boldsymbol{x}_i) \tag{3.4}$$

where each $h_t$ is a weak learner that takes an object $\boldsymbol{x}$ as input and returns a real valued result indicating the class of the object. The sign of the weak learner output identifies the predicted object class and the absolute value gives the confidence in that classification. Each weak learner produces an output, hypothesis $h(\boldsymbol{x}_i)$, for each sample in the training set. At each iteration $t$, a weak learner is selected and assigned a coefficient $\alpha_t$ such that the sum training error $E_t$ of the resulting $t$-stage boost classifier is minimized:

$$E_t = \sum_i E[H_{t-1}(\boldsymbol{x}_i) + \alpha_t h(\boldsymbol{x}_i)] \tag{3.5}$$

Term $H_{t-1}(\boldsymbol{x})$ is the boosted classifier that has been built up to the previous stage of training, $E(H)$ is some error function and $f_t(x) = \alpha_t\, h(\boldsymbol{x})$ is the weak learner that is being considered for addition to the final classifier. At each iteration of the training process, a weight is assigned to each sample in the training set equal to the current error $E(H_{t-1}(x_i))$ on that sample. These weights can be used to inform the training of the weak learner, for instance, decision trees can be grown that favor splitting sets of samples with high weights.

### 3.1.5 Support vector machines

Support vector machines (SVMs) are supervised learning models with associated learning algorithms that analyze data and recognize patterns, used for classification analysis (Abe, 2010). An SVM model is a representation of the examples as points in space, mapped so that the examples of the separate categories are divided by a clear gap that is as wide as possible. New examples are then mapped into that same space and predicted to belong to a category based on which side of the gap they fall on. The mappings used by SVM schemes are defined through a kernel function $k(x, y)$ selected tos suit the problem.

Given a training set of $N$ data points $\{\boldsymbol{x}_1, \dots, \boldsymbol{x}_k\}_{k=1}^N$, where $\boldsymbol{x}_k \in \mathbb{R}^n$ is the *k*-th input pattern and $y_k \in \mathbb{R}$ is the *k*-th output pattern, the classifier can be constructed using the support vector method in the form:

$$y(x) = sign\left[\sum_{k=1}^{N} \alpha_k y_k K(\boldsymbol{x}, \boldsymbol{x}_k) + b\right] \tag{3.6}$$

where $\alpha_k$ are called support values and $b$ is a constant. The $K(x, x_k)$ is the kernel, which can be either $K(\boldsymbol{x}, \boldsymbol{x}_k) = \boldsymbol{x}_k^T \boldsymbol{x}$ (linear SVM); $K(\boldsymbol{x}, \boldsymbol{x}_k) = (\boldsymbol{x}_k^T \boldsymbol{x} + 1)^d$ (polynomial SVM of degree d); $K(\boldsymbol{x}, \boldsymbol{x}_k) = \tanh[\kappa \boldsymbol{x}_k^T \boldsymbol{x} + \theta]$ (multilayer perceptron SVM), or $K(\boldsymbol{x}, \boldsymbol{x}_k) = \exp\{-\|\boldsymbol{x} - \boldsymbol{x}_k\|_2^2 / \sigma^2\}$ (RBF SVM), where $\kappa, \theta$, and $\sigma$ are constants.

The kernel parameters, i.e. $\sigma$ for RBF kernel, can be optimally chosen by optimizing an upper bound on the VC dimension. The support values $\alpha_k$ are proportional to the errors at the data points in the LS-SVM case, while in the standard SVM case many support values are typically equal to zero. When solving large linear systems, it becomes needed to apply iterative methods.

A common SVM formulation is the 2 class linear separation problem. Let us assume that the classes of negative and positives samples, described by feature vectors $\boldsymbol{f}$, are linear separable. This means that there exists a hyperplane $\mathcal{P} = \boldsymbol{w} \cdot \boldsymbol{f} - b = 0$ that separates the two classes ($\boldsymbol{w}$ is the normal vector to the hyperplane). SVM classifier tries to estimate and maximize the distance between two other hyperplanes, $\mathcal{P}_p = \boldsymbol{w} \cdot \boldsymbol{f} - b = 1$ and $\mathcal{P}_n = \boldsymbol{w} \cdot \boldsymbol{f} - b = -1$, that separate the two classes with no sample existing between them. This can be expressed by the following constraints:

$$\begin{gathered} \boldsymbol{w} \cdot \boldsymbol{f}_i - b \geq 1 \; if \; l_i = 1 \\ \boldsymbol{w} \cdot \boldsymbol{f}_i - b \geq 1 \; if \; l_i = -1 \end{gathered} \tag{3.7}$$

Exploiting the value of labels the pair of constraints, eq. (3.8) can be rewritten as:

$$l_i(\boldsymbol{w} \cdot \boldsymbol{f}_i - b) \geq 1 \; if \; l_i \geq 1, i =, \dots, n \tag{3.8}$$

The equality of constraint of eq. (3.8) holds for the samples that lie on the hyperplanes $\mathcal{P}_p$ and $\mathcal{P}_n$. These samples are called support vectors. The distance between these two hyperplanes is $2/\|\boldsymbol{w}\|$ , which implies that SVM try to solve the following optimization problem:

$$\min_{\boldsymbol{w},b} \frac{1}{2}\|\boldsymbol{w}\|^2$$
$$s.t.\, l_i(\boldsymbol{w} \cdot \boldsymbol{f}_i - b) \geq 1\; for\; i = 1, \dots, n \tag{3.9}$$

This formulation ensures that the maximum margin classifier classifies each example correctly, which is possible since we assumed that the data is linearly separable. In cases where the two classes are not linearly separable, to allow classification errors, the optimization problem of eq. (3.9) is transformed to (Cortes and Vapnik, 1995):

$$\min_{\boldsymbol{w},b,\xi} \frac{1}{2}\|\boldsymbol{w}\|^2 + c\sum_{i}^{n} \xi_i$$
$$s.t.\, l_i(\boldsymbol{w} \cdot \boldsymbol{f}_i - b) \geq 1\; for\; i = 1, \dots, n\; and\; \xi_i \geq 0 \tag{3.10}$$

where $\xi_i \geq 1$ are variables that allow a sample to be in the margin or to be misclassified and $c$ is a constant that weights these errors.

### 3.1.6 Artificial neural networks

Artificial neural networks are non-linear mapping structures, inspired by animal central nervous systems that are capable of machine learning and pattern recognition (Bahrammirzaee, 2010; Haykin, 1994; H. Li et al., 2011). ANNs are universal approximators which however have multiple local minima (i.e. solutions), due to their structure; they are composed from multiple hierarchical layers of interconnected nodes. Their structure consists of weights, biases and activation functions, imitating the real brain's neurons and synapses. Therefore simple computational units called neurons, which are highly interconnected are used. The work of (Hagan et al., 1996) provides a clear and detailed survey of basic neural network architectures and learning rules. The most widely used learning algorithm in an ANN is the back-propagation algorithm or its variations (Chakraborty and Ghosh, 2012).

Let us denote as $f$ a non-linear function (relationship) that indicates the status of a given datum (e.g. structural capacity of a pile, corresponding class, system's response, etc.). This non-linear relationship is approximated by a Feed Forward Neural Network (FFNN) architecture. Denote as $\hat{f}_{\boldsymbol{w}}$ the approximated function of $f$ as has been produced by the FFNN structure, where $\boldsymbol{w}$ denotes the neural network weights. Actually, function $f$ maps to a compact subset $A$ of $n$-dimensional Euclidean space, $\mathbb{R}^n$ to a bounded subset, $f[A]$, of $m$-dimensional Euclidean space, $\mathbb{R}^m$. Denote, also, as $S = \{(\boldsymbol{x}_1, \boldsymbol{y}_1), \dots, (\boldsymbol{x}_k, \boldsymbol{y}_k)\}$ a training set of $K$ elements used to find appropriate parameters to approximate the unknown function $\hat{f}_{\boldsymbol{w}}$ by estimating the weights $\boldsymbol{w}$.

The weights $\boldsymbol{w}$ are initially set as random numbers; they are adjusted, during training, in order to generate a mapping between input-output training patters. To estimate these weights, a reliable training set $S$ is needed. It has been shown in (Hornik et al., 1989) that a feed-forward neural network can approximate any non-linear function within any degree of accuracy. One of the most important problems, during network training, is overfitting, a situation in which the network can memorize training samples, providing a very small error on data of training set, without being able to generalize to new situations, i.e. bad generalization performance (Doulamis et al., 2000). One method for addressing this problem is to use the cross validation technique.

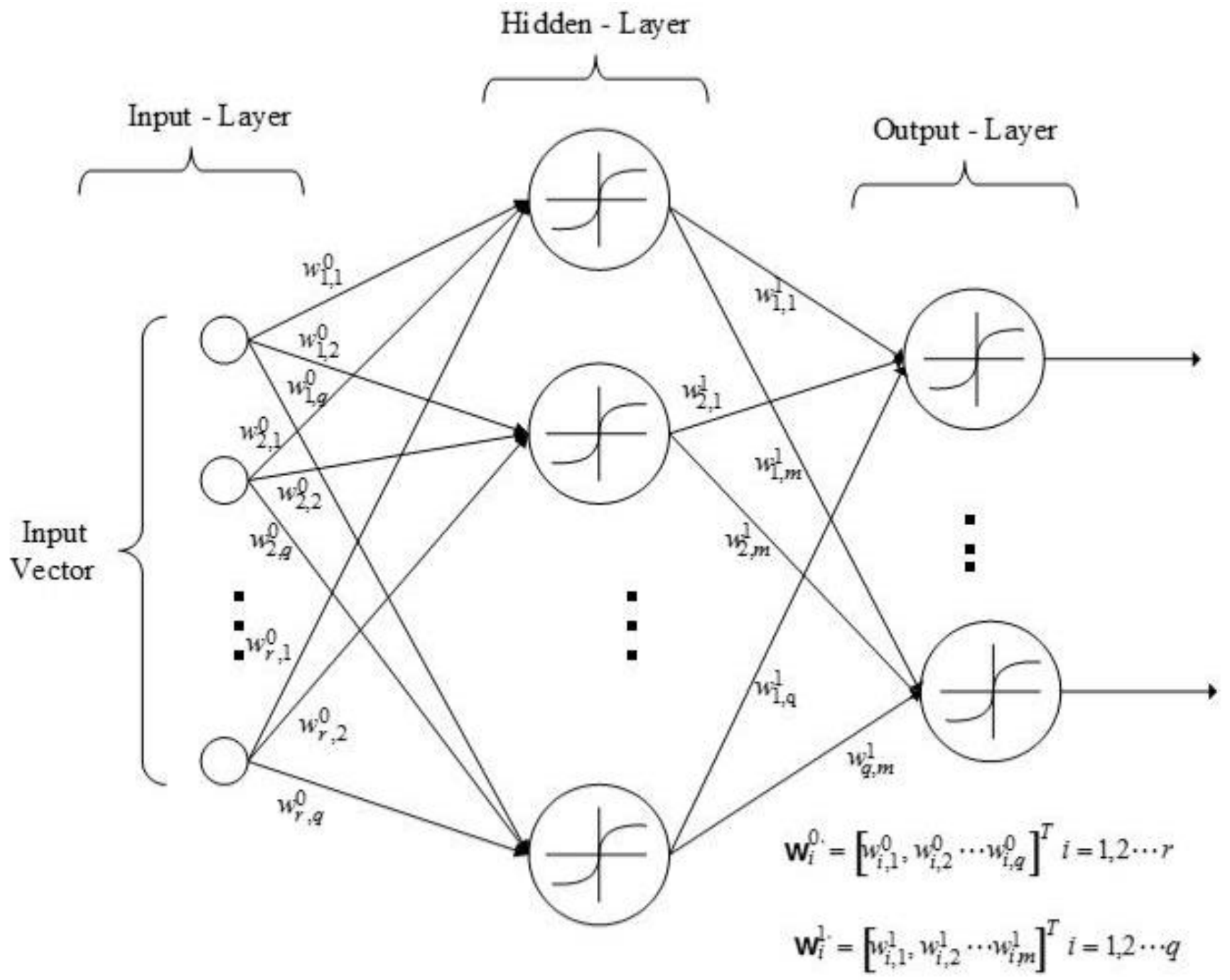

*Figure 3.1. An illustration of artificial neural network unit.*

The network performance, however, depends on a number of parameters, such as the network size, the number of neurons and the attributes used as inputs to the network. There is, also, the adopted training-algorithm impact. Apparently, proper selection of the network size is "art" in the sense that there are no concrete mathematical rules to define the structure, apart from proposing worst bounds. A typical approach during topology set up is the use of genetic algorithms, as shown in (Protopapadakis et al., 2016b).

## 3.2 Sampling & clustering techniques

The main purpose of data sampling is the selection of appropriate representative samples in order to provide a good training set and, thus, improve the classification performance of risk assessment models. The most important factor in data selection is the distance metric function definition. For any two given data points $\boldsymbol{x}_i$ and $\boldsymbol{x}_j$, let $d(\boldsymbol{x}_i, \boldsymbol{x}_j)$ denote the distance between them. In order to compute the distance, let $\boldsymbol{A} \in R^{m\times m}$ be a symmetric matrix, we define the formula of distance measure as:

$$d_{\boldsymbol{A}}(\boldsymbol{x}_i, \boldsymbol{x}_j) = \sqrt{(\boldsymbol{x}_i - \boldsymbol{x}_j)^T \boldsymbol{A} (\boldsymbol{x}_i - \boldsymbol{x}_j)} \tag{3.11}$$

The majority of the proposed approaches are Euclidean based (i.e. $\boldsymbol{A} = \boldsymbol{I}$). Sampling algorithms are used over the entire data set $\boldsymbol{\mathcal{X}}$ and create a new set, $\boldsymbol{\mathcal{X}}_r \subset \boldsymbol{\mathcal{X}}$ , of the most representative samples. In this thesis, we need at least one sample for every possible class, in order for the smooth functionality of the applied techniques. As such, $\boldsymbol{\mathcal{X}}_r$ is examined by an expert and additional data are used if necessary.

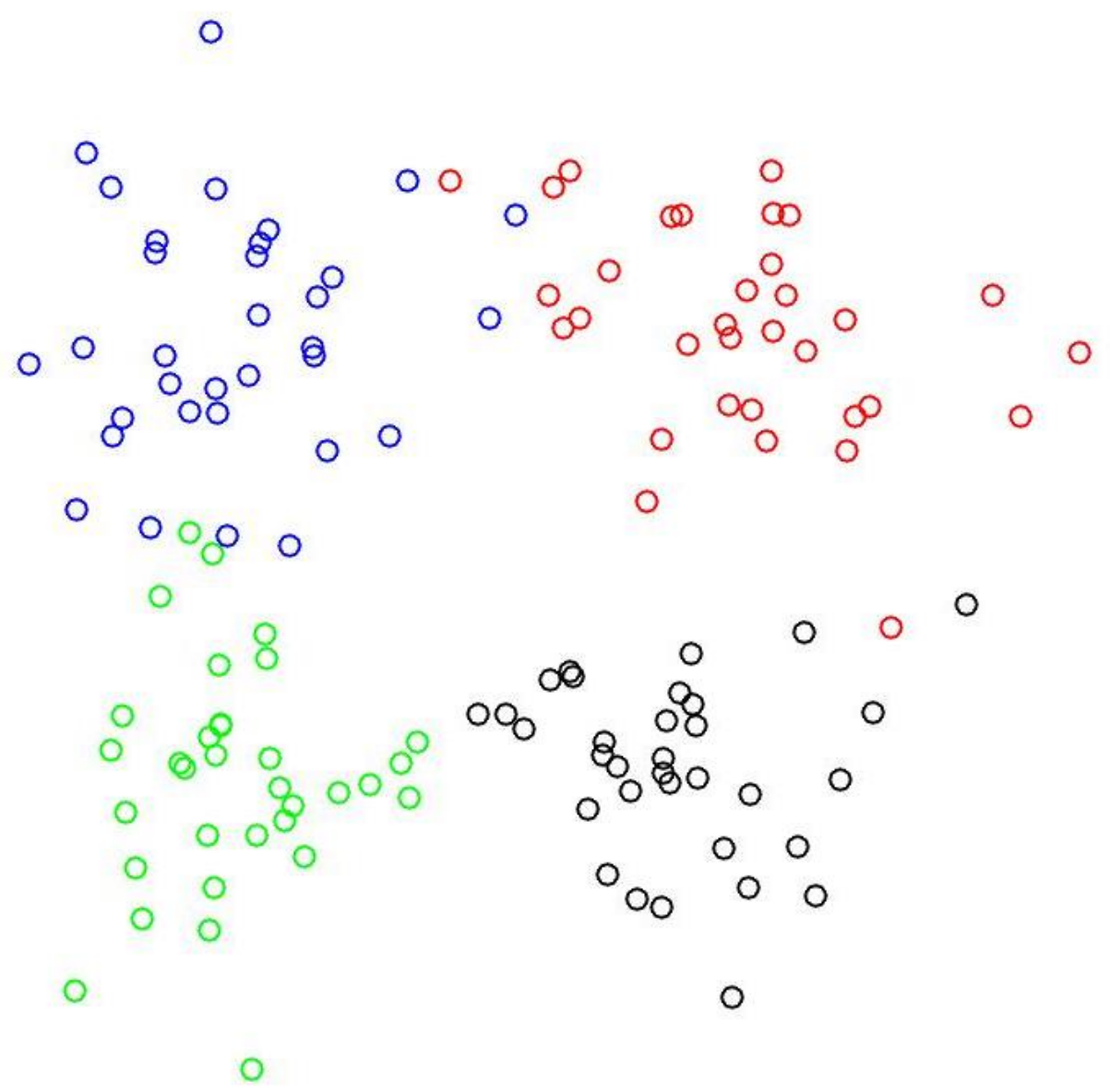

*Figure 3.2. A random dataset of four classes. It will serve for illustration purposes when describing various approaches in this chapter. The set consists of randomly generated data (a), distributed around 4 points: (0.45.0.45). (0.45.-0.45). (-0.45.0.45). (-0.45.-0.45). Data follow the normal distribution, $N(\mu.\sigma)$; mean and variance were set as $\mu = 0.293$ and $\sigma = 0.212$, respectively.*

A typical dataset is presented in Figure 3.2. The set consists of randomly generated data (a), distributed around 4 points: (0.45.0.45). (0.45.-0.45). (-0.45.0.45). (-0.45.-0.45). Data follow the normal distribution, $\mathrm{N}(\mu.\sigma)$; mean and variance were set as $\mu = 0.293$ and $\sigma = 0.212$, respectively. Such set contains outliers of all kinds; away from any distribution center, closer to a different class, or overlapping with different class data.

### 3.2.1 OPTICS algorithm

Ordering points to identify the clustering structure (*OPTICS*) is an algorithm for finding density-based clusters in spatial data (Ankerst et al., 1999); i.e. detect meaningful clusters in data of varying density. In order to do so, the points of the database are (linearly) ordered such that points which are spatially closest become neighbors in the ordering. A typical output of the algorithm is shown in Figure 3.3.

OPTICS requires two parameters: ε, which describes the maximum distance (radius) to consider, and $MinPts$, describing the number of points required to form a cluster. A point $p$ is a core point if at least $MinPts$ points are found within its ε - neighborhood, $N_\varepsilon(p)$. Once the initial clustering is concluded, we may proceed with any sampling approach (e.g. random selection among clusters).

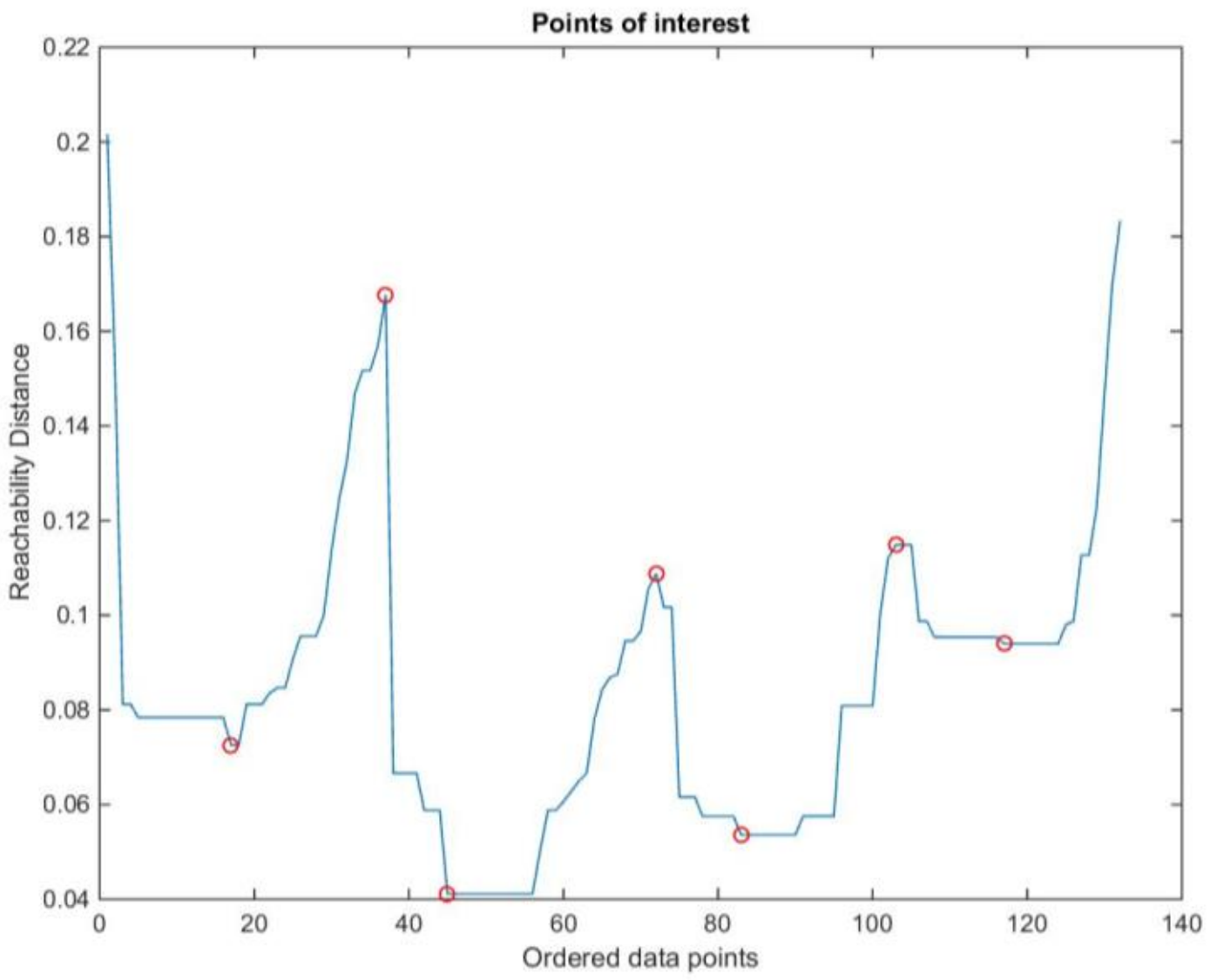

*Figure 3.3. OPTICS output illustration for a given data set of four classes (Figure 3.2). Points marked with 'o' describe local extrema. In this case, four valleys are formed among local maxima, indicating that there are four main classes among the data. Local fluctuations correspond to subclasses according to the data density.*

### 3.2.2 $k$-means algorithm

$k$-means clustering (Wu, 2012) aims to partition $n$ observations into $k$ clusters in which each observation belongs to the cluster with the nearest mean, serving as a prototype of the cluster. It is a classical approach that can be implemented in many ways and for various distance metrics. The main drawback is that the number of clusters should be known a priori. A typical mathematical formulation of such problem is the following:

$$\begin{aligned} &\min \sum_{l=1}^{k} \sum_{i=1}^{n} w_{il} d(\boldsymbol{x}_i, Q_l) \\ &s.t. \sum_{l}^{k} w_{il} = 1, 1 \leq i \leq n \\ &w_{il} \in \{0,1\}, 1 \leq i \leq n, 1 \leq l \leq k \end{aligned} \tag{3.12}$$

where $\boldsymbol{W}$ is an $n \times k$ partition matrix, $\boldsymbol{Q} = \{Q_1, \dots, Q_k\}$ is a set of objects in the same domain, and $d(\cdot)$ a distance measure as in eq. (3.11). A brief survey of $k$-means extensions can be found in (Huang, 1998).

### 3.2.3 Sparse representative selection

In order to extract the most important, i.e. descriptive, data, the work of (Elhamifar et al., 2012) around sparse modeling, is employed. Sparse representative selection (*Sparse)* focus on the identification of representative objects. Their work is summarized through the following formulation:

$$\begin{aligned} &\min \lambda \|\boldsymbol{C}\|_{1,q} + \frac{1}{2} \|\boldsymbol{X} - \boldsymbol{X}\boldsymbol{C}\|_F^2 \\ &s.t. \boldsymbol{1}^T \boldsymbol{C} = \boldsymbol{1}^T \end{aligned} \tag{3.13}$$

where $\boldsymbol{X}$ and $\boldsymbol{C}$ refer to data points and coefficient matrix respectively. This optimization problem can also be viewed as a compression scheme, where we want to choose a few representatives that can reconstruct the available data set. A typical illustration, for the data set of Figure 3.2, is shown in Figure 3.4.

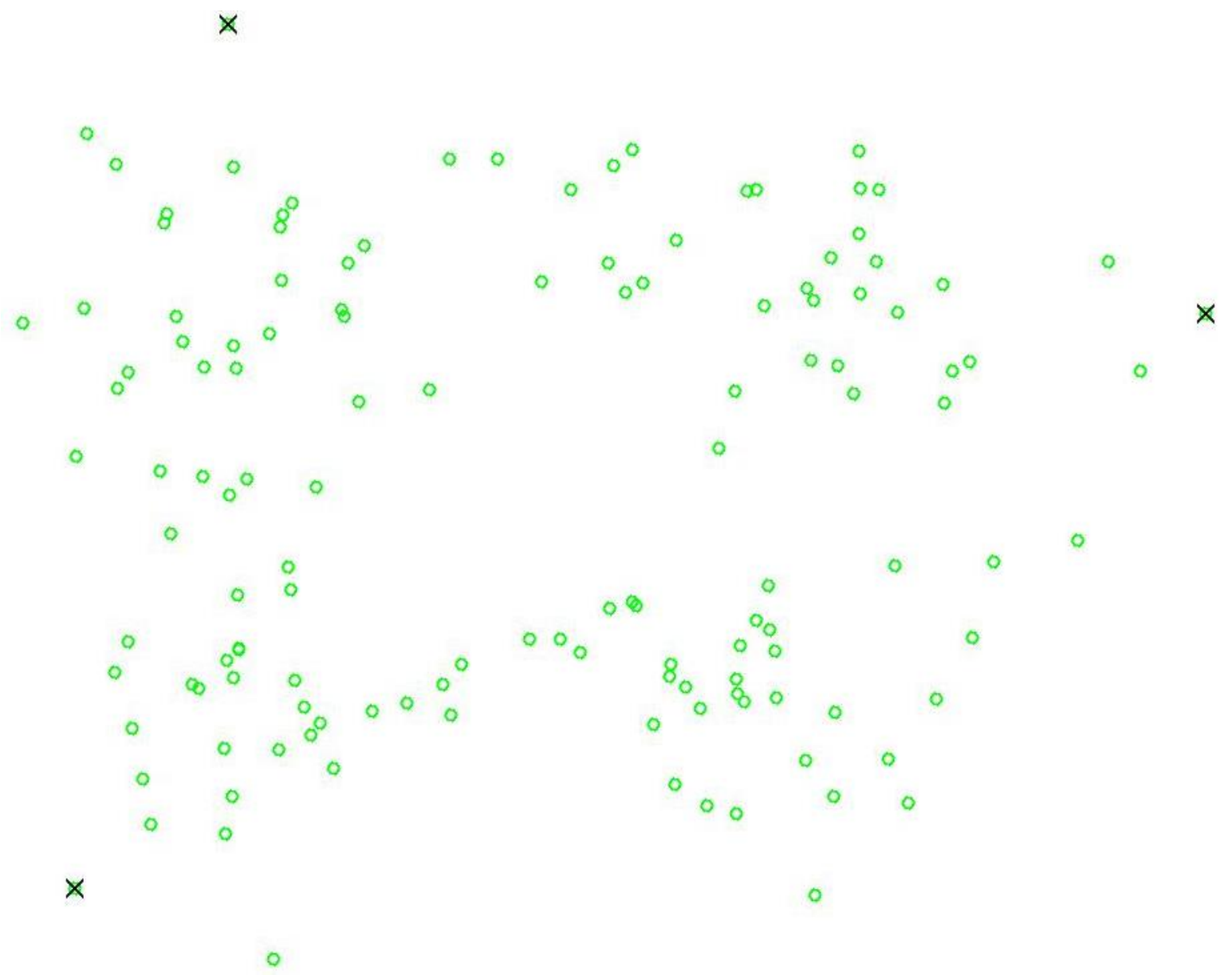

*Figure 3.4. SMRS descriptive data selection. Selected data are the on the edge of spanning volume.*

### 3.2.4 Kennard–Stone algorithm

The classic *KenStone* algorithm (Kennard and Stone, 1969) is a uniform mapping algorithm; it yields a flat distribution of the data. It is a sequential method that should cover the experimental region uniformly. The procedure consists of selecting as the next sample (candidate object) the one that is most distant from those already selected objects (calibration objects). As starting points we either select the two objects that are most distant from each other, or preferably, the one closest to the mean.

From all the candidate points, the one is selected that is furthest from those already selected and added to the set of calibration points. To do this, we measure the distance from each candidate point $\boldsymbol{x}_0$ to each point $\boldsymbol{x}$, which has already been selected and determine which is smallest, i.e. $\min_i d(\boldsymbol{x}, \boldsymbol{x}_0)$. From these we select the one for which the distance is maximal:

$$d_{selected} = \max_{\mathrm{i}_0}\left(\min_i d(\boldsymbol{x}, \boldsymbol{x}_0)\right) \tag{3.14}$$

In the absence of strong irregularities in the factor space, the procedure starts first by selecting a set of points close to those selected by the D-optimality method, i.e. on the borderline of the data set (plus the center point, if this is chosen as the starting point). It then proceeds to fill up the calibration space. A typical illustration, for the data set of Figure 3.2, is shown in Figure 3.5.

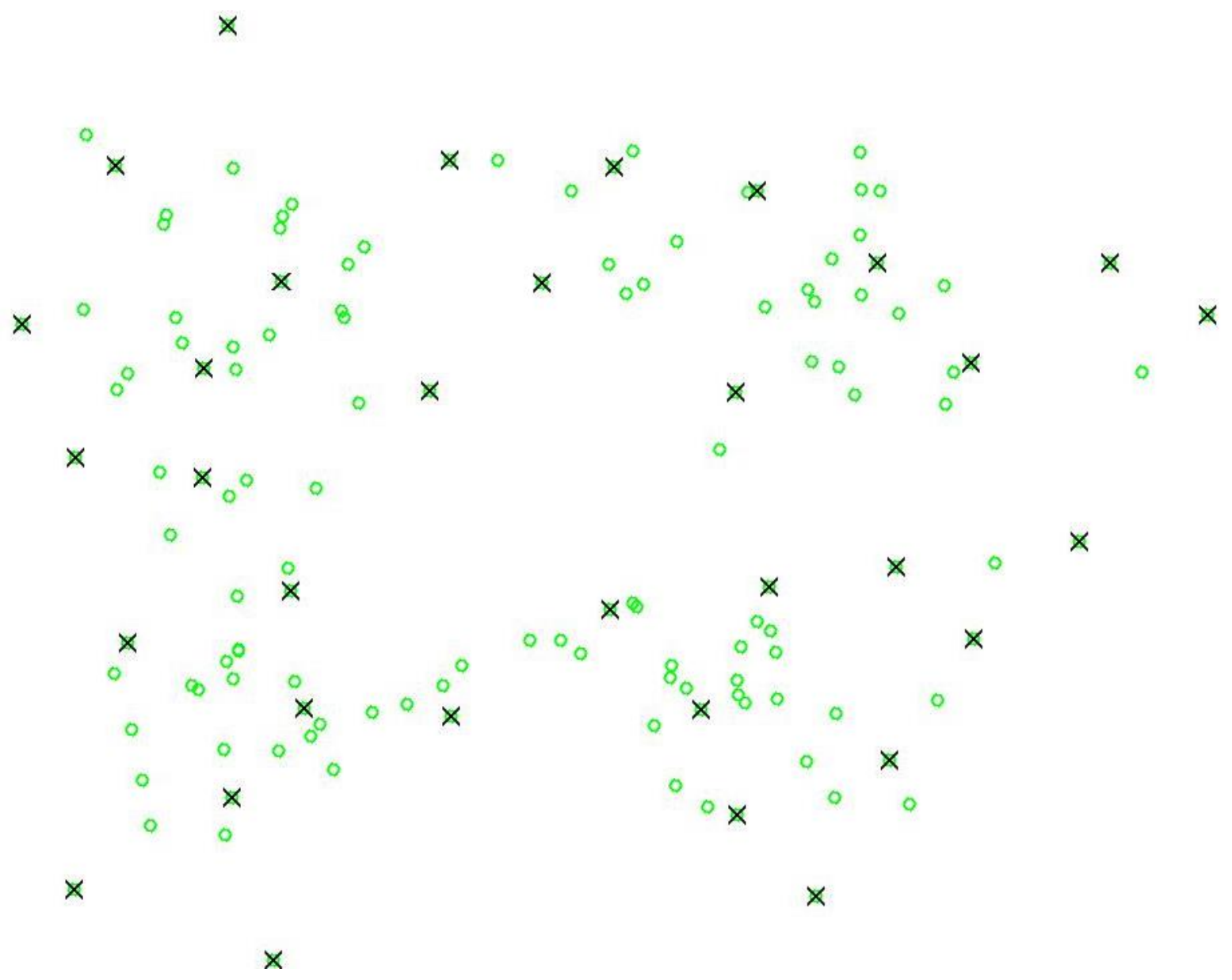

*Figure 3.5. KenStone sampling results. Selected data span a uniform area over the originally distributed samples.*

## 3.3 Performance metrics

Most of the application scenarios, presented in this thesis, were actually classification problems. As such, the primary tool for performance evaluation was the confusion matrix. Other, significant performance metrics were the execution time, algorithms' complexity and features' dimensionality and separability. In all of the developed systems ordinary desktop computers were used.

Usually we have two possible classes; defaulted / non-defaulted companies, defects / non-defects, detection / non-detection, etc. named positive (P) and negative (N) class, respectively. Given the outputs[4], we form the table of confusion, which is a 2 × 2 matrix that reports the number of false positives (FP), false negatives (FN), true positives (TP), and true negatives (TN). Given these values we are able to calculate various performance metrics regarding the defect detection performance. Metrics formulation is shown in Table 3.1. Metrics of special interest are: Sensitivity (proportional to TP) and miss rate (proportional to FN), which are both strongly connected to defect detection.

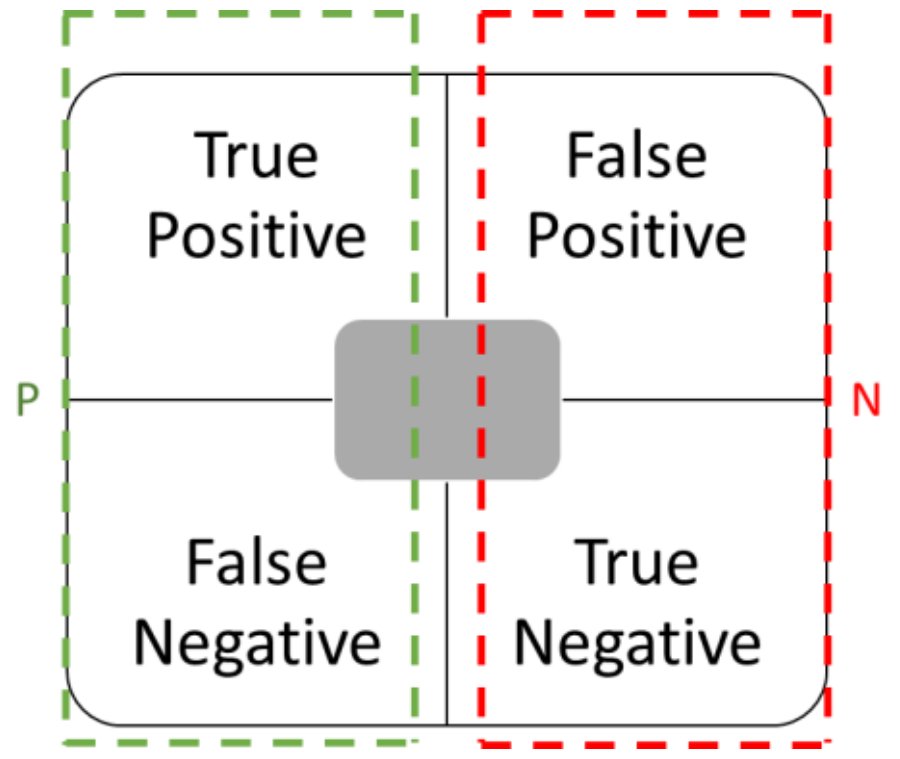

*Figure 3.6. A typical 2 × 2 confusion matrix, for a binary classification problem.*

[4] Integer values in the form {-1,1} or {1,2}. If the techniques produce soft labels (e.g. 0.92, -1.06, etc.) we round them towards the closest integer.

*Table 3.1. Metrics for quantitative performance evaluation.*

| *Metric* | *Formulation* | *Description* |
|---|---|---|
| *Sensitivity (TPR)* | TPR = TP / P | Also known as recall: the fraction of the positive samples that are relevant to the query that are successfully retrieved. |
| *Specificity (SPC)* | SPC = TN / N | It can be looked at as the probability that a non-relevant class is correctly identified by the model. |
| *Precision (PPV)* | PPV = TP / (TP + FP) | In binary classification known as positive predictive value. How many correct positive predictions we have. |
| *Negative predictive value (NPV)* | NPV=TN/(TN+FN) | How many correct negative predictions we have. |
| *False pos. rate (FPR)* | FPR = FP / N | It is the probability that a non-relevant class is incorrectly identified by the model. |
| *False discovery rate (FDR)* | FDR=FP/(FP+TP) | It conceptualize the rate of type I errors. |
| *Miss Rate (FNR)* | FNR = FN / P | It is the probability that a relevant class is incorrectly identified by the model. |
| *Accuracy (ACC)* | ACC = (TP + TN) / (P + N) | Percentage of correct classification for ALL classes. |
| *F1 score (F1)* | F=2TP/(2TP+FP+FN) | The weighted harmonic mean of precision and recall. |

Additionally, when we deal with a binary classifier concept, a receiver operating characteristic (ROC), or ROC curve, could provide further information over the performance. ROC is a graphical plot that illustrates the performance as classifier's discrimination threshold is varied. The curve is created by plotting the true positive rate against the false positive rate. However, if there are many alternative model combinations jeopardizing any graphical illustrations, we use the area under the ROC curve (AUC).

# *Chapter* IV: The Sampling Impact

*Beware of little expenses. A small leak will sink a great ship.*
*Benjamin Franklin, one of the Founding Fathers of the U.S.*

## 4 Labeled data selection impact in credit risk assessment

In this chapter, we emphasize on credit risk assessment of Greek firms (commercial sector), over the period 2006–2009. Suggesting approach involves synergistically models from both sampling and classification fields. In particularly, we demonstrate the sampling influence over the models' performance. A great variety of sampling approaches are used to evaluate the descriptive abilities of small training sets, given a classifier raging from traditional models, e.g. logistic regression, to advanced soft computing techniques, e.g. artificial neural networks.

Results provide an extensive joint performance evaluation of sampling-classification models' synergies. Comparisons are based on various quantitative performance metrics, including the initialization time required. Simulation outcomes suggest that **no optimal choice, regarding the data sampling, neither for the classification approach, exists**. There is a variety of synergistic assessment models; the best alternative is always defined by the user preferences (e.g. execution time, accuracy, precision, etc.).

Finally, due to the abundance of unlabeled data, SSL approaches were investigated, regarding the applicability on the credit risk assessment field. There was no limitations at the selection of the labelled data, except from labeled/unlabeled ratio. However, there are some points where attention is required:

1. We need representative samples. The labeled samples should be able to describe (reproduce) the original data set as good as possible.
2. At least one sample per classification category is required. We need an instance per class, so that model will be able to adjust at the class properties.
3. Outlier consideration. Most data sets contain outliers which could lead to poor performance especially when used as labeled data (all by themselves)

### 4.1 Introduction

Credit risk analysis is a fundamental and challenging issue in financial risk management, and has been the major topic of financial and banking industry in the last decades. Credit scoring models have been extensively used to evaluate the credit risk of enterprises, and they can classify the applicants as either accepted or rejected according to the examined criteria. Over the past decades, a great sum of credit risk decision models have been proposed and evaluated (Marqués et al., 2013; Thomas, 2000).

Existing approaches in credit risk assessment, mainly, include linear discriminant analysis and logistic regression (Vojtek and Kocenda, 2006; Zeng and Gao, 2009), nearest neighbor analysis (Henley and Hand, 1996), Bayesian networks (Pavlenko and Chernyak, 2010), artificial neural networks (Baesens et al., 2003), decision trees (Zhang et al., 2010), genetic algorithms (Abdou, 2009), multiple criteria decision making (J. Li et al., 2011; Niklis et al., 2013; Yu et al., 2009; Zhang et al., 2014), support vector machines (Bellotti and Crook, 2009; Martens et al., 2007) and so on. Recently, increasing interests in the synergies of optimization and data mining can be observed (Corne et al., 2012; Meisel and Mattfeld, 2010; Niklis et al., 2013; Olafsson et al., 2008).

Regardless the core method for the credit assessment, all possible sources of variation, which will be encountered later, must be included in the training data set; the model will be used for the prediction of new samples. In other words, the training data should have a greater variation in feature attributes than the data to be analyzed. As such, before applying any assessment model, we should provide with a good data set; important factors affecting the overall performance of such models can be the class imbalance, outliers and low quality features.

(Abdou and Pointon, 2011) review 214 previous studies on credit scoring applications, emphasizing on the statistical techniques used for evaluation. Results indicate that there is not an overall best technique for building assessment models. It appears that assessment models setup is usually heuristically defined, depending on the data availability, application field and various other factors. Current literature provides extensive comparison results among credit scoring methods; e.g. the review of (Marqués et al., 2013) advocates the benefits of using evolutionary computation for credit scoring.

However, due to the great variation of factors in model assessment performance there is always room for further analysis and comparative studies. Thus, in section 4.4.2 we perform a comparative study among various, well-known, sampling techniques and predictive models, trying to identify the best possible combination for the credit risk assessment problem in Greek commercial sector.

## 4.2 Related work

The rapid development in information and computer technology has created new techniques, which are appearing under the name of data mining. Advanced data analysis techniques are currently used to evaluate risk in credit approval (Huang et al., 2004) and fraud detection (Ngai et al., 2011). Recent literature in the search of trends, in data mining applications, for the banking industry can be found in (Moro et al., 2015).

Data mining methods, especially pattern classification, using real-world historical data, is of paramount importance in building such predictive models (Yu et al., 2008). Other approaches involve hybrid data mining techniques, filtering algorithms, attribute relevance, etc. in the following lines we present a brief description over recent approaches.

For example, (W. Chen et al., 2012) proposed a hybrid data mining technique which contains two processing stages. The proposed model was a two-stage approach: k-means cluster, support vector machines classification and computation of feature importance. Experimental results based on the credit data set provided by a local bank in China showed that by choosing a proper cut-off point, super classification accuracy of the good and the bad credit is obtained. In (Yap et al., 2011) data mining techniques were used to improve the assessment of credit worthiness of credit scoring models. More specifically, three models were examined: credit scorecard model, logistic regression model and decision tree model. Results show the performances of the three models are quite similar. Scorecards are relatively much easier to deploy in practical applications.

The work of (García et al., 2012), investigated whether the application of filtering algorithms leads to an increase in accuracy of instance-based classifiers in the context of credit risk assessment. The experimental results show that the filtered sets perform significantly better than the non-preprocessed training sets when using the nearest neighbor decision rule. (Khashman, 2011), investigate the efficiency of emotional NNs and compare their performance to conventional NNs when applied to credit risk evaluation. Experimental results suggest that both neural models can be used effectively for credit risk evaluations, however the emotional models outperform their conventional counterparts in decision making speed and accuracy, making them ideal for implementation in fast automatic processing of credit applications.

A very popular problem, among these studies, is the attribute relevance by using new and existing feature selection algorithms (Chen and Li, 2010; Liu and Schumann, 2005; Shukai et al., 2010). Few studies address the problem of credit data with noise and outliers. (Kotsiantis et al., 2006) present a survey of data preprocessing techniques for financial prediction, including discretization, feature selection and instance selection. (Tsai and Chou, 2011) use a genetic algorithm to perform feature selection and data filtering for bankruptcy prediction. (Tsai and Cheng, 2012) explore the performance of artificial neural networks, decision trees, logistic regression and support vector machines after removing different amounts of outliers from credit data sets.

## 4.3 Proposed methodology

In this chapter, a two-step process is adopted, in order to assess commercial firms' credit risk. The first step involves data sampling; i.e. the selection of the most descriptive representatives in the available data set. The second step, employs popular data mining algorithms; i.e. predictive models are trained over the descriptive subsets of the previous step. Generated models' performance is calculated for a set of evaluation metrics, plus the computational cost (in terms of time spend during initialization process), in order to analyze both the efficiency of sampling schemes and the synergies' outcomes.

The selection of a subset of individuals, from within a currently available data set, to estimate characteristics of newly appeared data, is a crucial step. Such an approach is able to identify the most representative cases and handle class imbalances in an effective manner, as the infrequent nature of defaults affects the performance of predictive models. There are many sampling approaches, varying from random sample selection to more complicated techniques. Sampling can be used in many ways (i.e. outliers removal (Janssens et al., 2012), feature subset selection (Daszykowski et al., 2002a), representative data selection (Elhamifar et al., 2012), etc.). In our case, sampling approach moderates the effects of two common problems; i.e. class-imbalanced data sets and training/evaluation ratio.

*Table 4.1. Proportion of training / evaluation datasets (%).*

| Related work | Train | Test |
|---|---|---|
| (Abdou et al., 2008) | 80 | 20 |
| (Boros et al., 2000) | 71 | 29 |
| (Hsieh, 2005) | 68 | 32 |
| (Baesens et al., 2003; Kim and Sohn, 2004; Tsai and Wu, 2008) | 70 | 30 |
| (Šušteršič et al., 2009) | 69 | 31 |
| (Setiono et al., 2008) | 67 | 33 |
| (Atiya, 2001) | 62 | 38 |
| (Sakprasat and Sinclair, 2007) | 50 | 50 |
| (Khashman, 2011) | 44 | 56 |

Class-imbalance is a major and very common problem, many studies show that a classifier tends to over-fit the observations of majority-class and simultaneously under-fit the observations of minority-class (Akbani et al., 2004; Yang et al., 2008; Zeng and Gao, 2009). Additionally, even in class-balanced data sets, there are no explicit rules regarding the data amount allocated between training and test sets. As we can see in Table 4.1, there is no census over the proportion for the training set over the currently available data. However, there is a general trend in utilizing around 70% of the data for the training process and evaluating the performance with the remaining 30%.

Regarding the classification models, a variety of them was utilized, ranging from traditional approaches (e.g. logistic regression) to soft computing techniques (e.g. artificial neural networks). These models required set up for at least one parameter. These parameters were defined using heuristic approaches, or they were remain intact at their original values, as provided by the software developers. The SSL techniques were low density separation (Chapelle and Zien, 2004) and Harmonic functions (Zhu et al., 2003).

## 4.4 Experimental setup

We are interesting in evaluating both the sampling approaches performance as well as the classification accuracy of the models. An efficient sampling technique should accurately select few representative samples, which support a smooth model training. Additionally, we investigate the behavior of various predictive models given different training data sets. Specifically, we examine the predictive performance over a year, $N$ (evaluation year), given a small training set from previous years, $N-1, N-2, \dots$ .

More than 200 experiments were performed using various combinations of sampling techniques and classification approaches, over the data set. Every experiment had three characteristic steps (Gorunescu, 2011), described below:

1. Exploring the data: At first we normalize values in $[0,1]$ according to the minmax approach. Then, a sampling approach, over the normalized values, separates data into training and evaluation (test) sets.
2. Building the model: We select the classification model. Training and validation sets are used for the parameters tuning. We also change the cost function (if applicable) in order to handle the unbalanced classes.
3. Applying the model: The trained model predicts the labels over the evaluation (test) data set. Various performance metrics are calculated.

The data mining performance is evaluated regarding both the sampling techniques and the classification approaches. Emphasis is given over the classification performance indexes over the defaulted companies. All applied techniques were implemented in MatLab. A standard quad core, 8GB ram, desktop computer was used.

### 4.4.1 Dataset description

The sample consists of 10716 firm-year observations for non-listed Greek firms from the commercial sector, over the period 2006–2009. The data were obtained from the financial database of ICAP, as described in (Niklis et al., 2014). All observations in this sample are classified as default or non-default on the basis of the definition of default employed by ICAP, which considers a range of default-related events such as bankruptcy, protested bills, uncovered cheques, payment orders, etc. Table 4.2 presents the number of observations per year and category.

*Table 4.2. Sample observations by year and category.*

| Years | Defaulted | Non-defaulted | Total |
|---|---|---|---|
| **2006** | 2,748 | 52 | 2,800 |
| **2007** | 2,846 | 53 | 2,899 |
| **2008** | 2,731 | 99 | 2,830 |
| **2009** | 2,143 | 44 | 2,187 |
| **Total** | 10,468 | 248 | 10,716 |

On the basis of data availability and the relevant literature seven financial ratios are used to describe the financial status of the firms in both samples. The selection of the appropriate financial ratios is a challenging issue. In fact, there is a big variety of ratios that can be used as proxies of the same financial dimensions (leverage, liquidity, profitability, etc.). Furthermore, time and cost issues arise when using a large number of ratios and this can also cause multi-colinearity problems. Table 4.3 presents the selected ratios together with their expected relationship (sign) to the creditworthiness of the firms. A positive sign (+) is used to indicate ratios which are positively related to the creditworthiness of the firms, in the sense that higher values in these ratios are expected to improve the creditworthiness of the firms. The rest of the ratios are assigned a negative sign (–) indicating their negative relationship with the performance and viability of the firms (i.e., as these ratios increase the likelihood of default is also expected to increase).

*Table 4.3. Selected financial ratios description.*

| Category | Variables | Short title | Relationship to default |
|---|---|---|---|
| **Management efficiency** | Short-term liabilities*365 / Cost of Sales | STL/CS | + |
| | Accounts receivable*365 / Sales | AR/S | + |
| | Inventories / Cost of sales | I/CS | + |
| **Profitability** | Profit before tax / Total assets | PBT/TA | – |
| | Financial expenses / Sales | FE/S | + |
| **Solvency** | Quick assets / Short-term liabilities | QA/STL | – |
| | Total liabilities/Total assets | TL/TA | + |

We have four categories of financial ratios (efficiency, profitability, liquidity and financial leverage) and four non-financial indicators that are crucial for commercial companies. Management efficiency ratios are typically used to analyze how well a company uses its assets and liabilities. Efficiency ratios are important because an improvement in the ratios usually translates to improved profitability. We have selected three ratios in this category, which are positive related to credit risk, in the sense that the higher their value, the higher is the probability of default.

The profitability indicators are used to assess firm's ability to generate earnings, compared to its expenses and other relevant costs incurred during a specific period of time. The profitability ratios considered in this study include the return on assets ratio and ratio of financial expenses to sales. The first one is negative related to credit risk in contrast with the second one which is positively associated to the probability of default.

Finally, the category of solvency indicators includes two ratios related to the liquidity and the financial leverage of the firms. Liquidity determines a company's ability to pay off its short-term debt obligations. In this study the quick ratio (Current assets-Inventories/Short-term liabilities) is used which is negative related to credit risk. On the other hand, financial leverage provides an indication of the long-term solvency of a firm. Here the ratio of total liabilities to total assets is used which is positive related to credit risk.

Apart from financial indicators there should be a consideration of other factors that affect the operation of a firm. In our case, two factors are examined:

1. Logarithm of employees (LOGE). This is an indicator of the size of a company, which has been shown in past studies to be negatively associated to the probability of default.
2. Activity indicator. For commercial companies it is important to take into consideration the type of their activities. In this study, in accordance with ICAP's modeling approach, the activities of the companies in the sample are characterized by the following three binary indicators: exports indicator (EXP), imports indicator (IMP), and Representations indicator (REPR); i.e., companies that are local resellers of products of foreign companies.

Table 4.3 summarizes the numerical values of the selected financial ratios, for each group of observations (i.e., default, non-default). The differences between the defaulted and non-defaulted firms from the sample of non-listed companies, they are all found significant at the 1% level under the Mann-Whitney test. For the significance of the binary attributes regarding the business activity of the firms was tested with a $\chi^2$ test and all three indicators were found significant at the 1% level[5].

*Table 4.4. Selected financial ratios description numerical attributes.*

| | Non Defaulted | | | | | | Defaulted | | | | | |
|---|---|---|---|---|---|---|---|---|---|---|---|---|
| | Average values | | | Standard deviation | | | Average values | | | Standard deviation | | |
| | **2007** | **2008** | **2009** | **2007** | **2008** | **2009** | **2007** | **2008** | **2009** | **2007** | **2008** | **2009** |
| **STL/CS** | 425.68 | 458.06 | 466.89 | 368.28 | 400.95 | 404.59 | 529.86 | 647.85 | 731.53 | 384.29 | 451.51 | 443.96 |
| **AR/S** | 221.42 | 213.11 | 214.82 | 209.87 | 220.26 | 236.82 | 324.90 | 255.48 | 188.25 | 324.53 | 246.97 | 281.78 |
| **I/CS** | 104.47 | 112.77 | 117.14 | 164.31 | 175.70 | 183.70 | 154.68 | 198.47 | 239.15 | 233.74 | 294.97 | 319.81 |
| **PBT/TA** | 0.04 | 0.04 | 0.03 | 0.13 | 0.12 | 0.11 | -0.03 | -0.03 | -0.07 | 0.17 | 0.12 | 0.13 |
| **FE/S** | 0.03 | 0.03 | 0.03 | 0.03 | 0.04 | 0.04 | 0.06 | 0.07 | 0.08 | 0.06 | 0.06 | 0.06 |
| **QA/STL** | 1.13 | 1.14 | 1.23 | 0.85 | 0.87 | 0.93 | 0.87 | 0.86 | 0.84 | 0.76 | 0.56 | 0.69 |
| **TL/TA** | 0.73 | 0.80 | 0.70 | 0.27 | 0.39 | 0.27 | 0.84 | 0.85 | 0.89 | 0.27 | 0.30 | 0.16 |

### 4.4.2 Sampling approaches

Given the financial assessment problem, the primary goal of sampling approaches is the redundant data removal. Thus, such approaches exclude a large amount of non-defaulted companies. The proposed sampling approaches utilize many of the previously described algorithms in various ways.

When many samples are available, we can first measure their spectra and select a representative set that covers the calibration space (x-space) as well as possible. Normally such a set should also represent the y-space well, this should preferable be verified. The chemical analysis with the reference method, which is often the more expensive step, can then be restricted to the selected samples.

Several approaches are available for selecting representative calibration samples. The simplest is random selection, but it is open to the possibility that some sources of variation will be lost. These variation sources are often represented by samples that are less common and have little probability of being selected. A second possibility is based on knowledge about the problem. If one is confident that we are aware of all the sources of variation, samples can be selected on the basis of that knowledge. However, this situation is rare and it is very possible that some source of variation will be forgotten.

Given an evaluation year, $N$, sampling approaches actuate over years $N-1, N-2, \dots$ . Bellow we present a brief description of the proposed approaches. The size of the final train set can be found in Table 4.5. Training data sets consist of all defaulted firms plus an indicative number of non-defaulted ones.

1. OPTICS extrema: We perform Optics algorithm on the entire data set. Then we locate local maxima and minima, over OPTICS calculated distances. All the extrema indexes are considered as labeled instances and the rest as unlabeled. The proposed approach results in a very limited train set; OPTICS approach results in the minimum possible sum of outliers, given a feature set and a distance metric. Selected points location can be anywhere in the data spanning area. The code of this algorithm was provided by the authors of (Daszykowski et al., 2002b).

2. Sparse modeling representative selection (SMRS): we perform the SMRS approach over the entire data set. The proposed approach results in a very limited train set, which is, however, greater than OPTICS extrema approach. In contrast to OPTICS, selected points are located only on the exterior cell of the available data volume. The algorithm has the same implementation as in (Elhamifar et al., 2012).

3. Combination of OPTICS and SMRS (OPTICS SMRS). In this case SMRS is used among the sub-clusters indicated by OPTICS algorithm. This approach is similar to the work of (Protopapadakis et al., 2014). It creates a small subset of representative samples. This approach locate the most descriptive data among each subcluster, as indicated by OPTICS output.

[5] Inventories / Cost of sales ratio was significant at 5%.

4. Combination of $k$-means and SMRS ($k$-means SMRS). We first divide the set into $k$ sub clusters. For each sub cluster we run SMRS algorithm to get the representative samples among each subcluster. As such, outcome results in surrounding points for each of the subclusters.

5. Kennard and Stone (KenStone) sampling data points. We execute KenStone algorithm over years $N-1, N-2, \dots$. Thus, we have data entries spanning uniformly the entire data space in each of the given years. The code was provided by the authors of (Daszykowski et al., 2002a).

6. Random selection. A random selection that picks 40% of the previous years' data as training samples.

7. An alternative approach is the creation of $k$ clusters (using k-means) and a random selection of $m$ samples from each cluster ($k$-means RANDOM). It can be seen as an improvement of random selection, without involving any advanced techniques. Similar instances are, likely, clustered together. Thus, the few random samples from each cluster are expected to provide adequate information, over the data set.

Almost all previously mentioned methods require some parameters as input. These parameters were estimated using heuristic approaches. For instance, the number of clusters, $k$, was defined using the rule: $k = \left\lceil\sqrt{n/2}\right\rceil$ , where $n$ denotes the number of available samples. The minimum number of data within a cluster, $m_c$, was defined as: $m_c = \lfloor n/k \rfloor$. Regardless the selected data instances, at the end of the sampling process, an additional step includes all non-defaulted data entries within the sampling set.

*Table 4.5. Illustration of the training set data size per sampling approach*

| EVALUATION YEAR | ENTIRE SET[6] | KENSTONE | KMEANS RANDOM | KMEANS SMRS | OPTICS EXTREMA | OPTICS SMRS | RANDOM | SMRS |
|---|---|---|---|---|---|---|---|---|
| **2007** | 2800 | 707 | 340 | 310 | 63 | 111 | 1148 | 72 |
| **2008** | 5699 | 1450 | 695 | 648 | 127 | 231 | 2349 | 145 |
| **2009** | 8529 | 2197 | 1084 | 995 | 237 | 399 | 3534 | 262 |

*Table 4.6. Training Set creation time (mins), using the sampling algorithms.*

| EVALUATION | ENTIRE SET | KENSTONE | KMEANS | KMEANS | OPTICS | OPTICS | RANDOM | SMRS |
|---|---|---|---|---|---|---|---|---|
| **2007** | 0.00025 | 0.42472 | 0.00701 | 0.65417 | 0.06346 | 2.16183 | 0.00063 | 14.84835 |
| **2008** | 0.00028 | 0.94954 | 0.01198 | 1.16747 | 0.09685 | 4.97662 | 0.00026 | 36.53702 |
| **2009** | 0.00028 | 1.39297 | 0.02380 | 1.60534 | 0.16518 | 7.48545 | 0.00121 | 52.40785 |

[6] For evaluation year 2007, data from year 2006 are used. Similarly, for evaluation year 2008 models are trained using data of 2006 – 2007; for the evaluation year 2009 data span from 2006 to 2008.

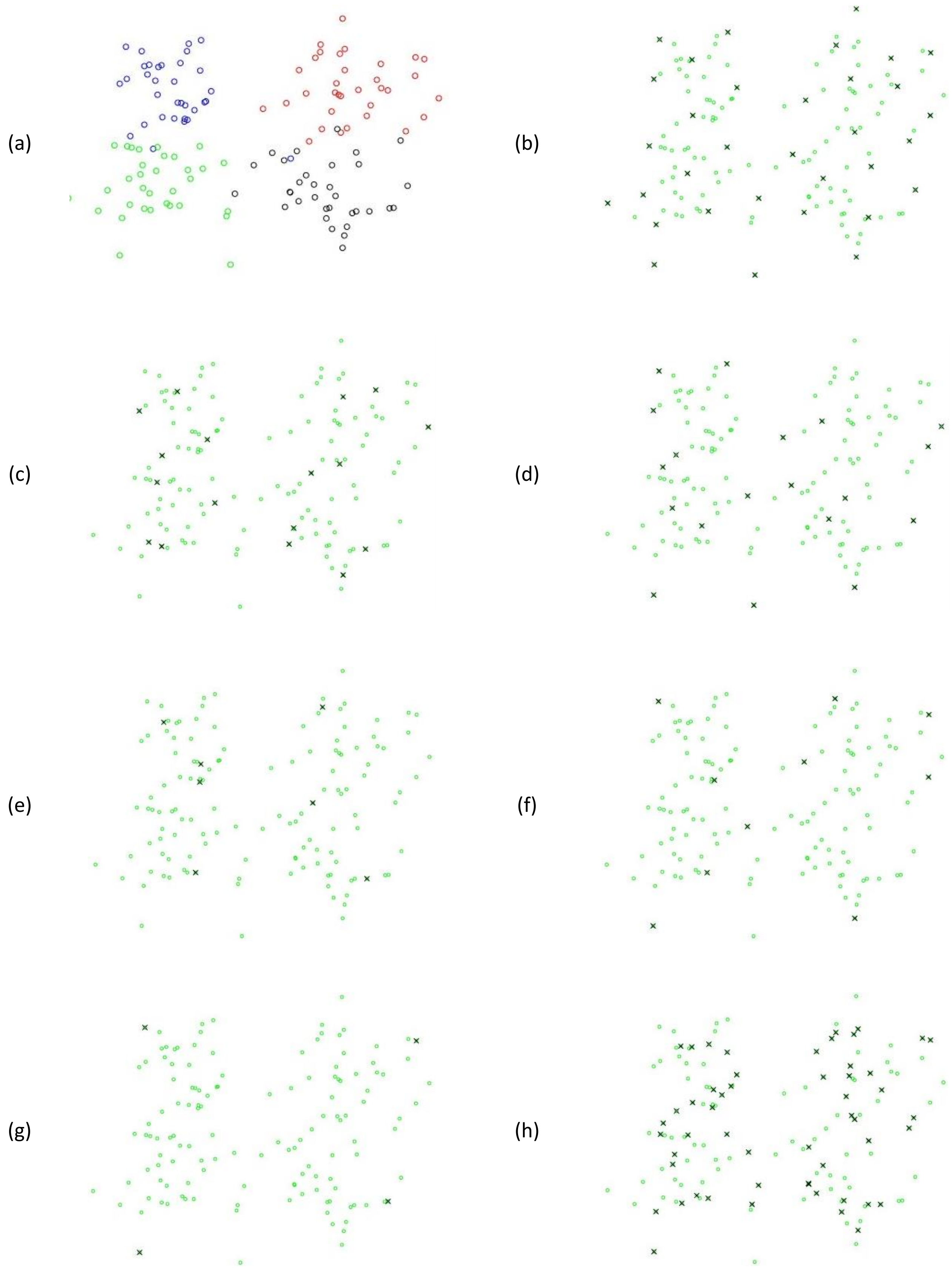

*Figure 4.1. Illustration of sampling approaches for randomly generated data (a), distributed around 4 points: (0.45.0.45). (0.45.-0.45). (-0.45.0.45). (-0.45.-0.45); All data* $\sim N(\mu.\sigma)$. $\mu = 0.293$. $\sigma = 0.212$. *Data are represented with 'o'. Selected samples are marked with 'x'. The sampling approaches are: (b) KenStone, (c) kmeans Random, (d) kmeans SMRS, (e) OPTICS extrema, (f) OPTICS SMRS, (g) SMRS, (h) Random.*

*Table 4.7. Average time for the assessment model training, using various sampling approaches (mins).*

| | LinReg | knn | FFnet | Ctree | Adaboost knn | Lin SVM | Poly SVM | Rbf SVM | LDS | Harmonic |
|---|---|---|---|---|---|---|---|---|---|---|
| EntireSet | 0.0032 | 0.0010 | 0.1615 | 0.0020 | 0.0106 | 0.0253 | 0.0408 | 0.0415 | 26.2818 | 0.2502 |
| Kenstone | 0.0030 | 0.0010 | 0.1016 | 0.0015 | 0.0079 | 0.0030 | 0.0026 | 0.0030 | 2.8251 | 0.1024 |
| kmeansRandom | **0.0026** | 0.0010 | 0.0775 | 0.0012 | 0.0070 | 0.0004 | 0.0006 | 0.0007 | 1.5645 | 0.0873 |
| kmeansSMRS | **0.0026** | 0.0010 | 0.0799 | 0.0013 | 0.0069 | 0.0005 | 0.0005 | 0.0007 | 1.4640 | 0.0743 |
| Optics extrema | 0.0028 | 0.0010 | 0.0914 | 0.0011 | 0.0067 | 0.0008 | 0.0001 | 0.0001 | **0.9235** | 0.0637 |
| Optics SMRS | 0.0625 | 0.0306 | 0.4706 | 0.0112 | 0.0295 | 0.0025 | 0.0011 | 0.0014 | 1.0319 | 0.0709 |
| Random | 0.0078 | 0.0025 | 0.1372 | 0.0029 | 0.0113 | 0.0085 | 0.0070 | 0.0073 | 5.2051 | 0.1264 |
| SMRS | 0.0028 | 0.0010 | **0.0729** | **0.0012** | **0.0068** | **0.0001** | **0.0001** | **0.0002** | 0.9374 | **0.0622** |

## 4.5 Classification results

The following classifiers were implemented using native code included in MatLab toolboxes: linear, $k$NN, trees, adaptive boosting and ANN. For the SVMs the LIBSVM implementation (Chang and Lin, 2011) was utilized. The ANN adopt a two hidden layer topology (Angelini et al., 2008) with 8 neurons per layer. SSL approaches used (Chapelle and Zien, 2004; Zhu, 2003) implementations; for both cases 5 nearest neighbors graphs were constructed.

Under sampling the majority class resulted in less data (see Table 4.5). However, that cannot guarantee a balanced train set. Thus, we had to adjust the weights in cost functions, during training. In particular, we emphasize the default class by reducing the weigh value of the non-defaulted class.

### 4.5.1 Performance evaluation

In this section, extensive results tables are provided, covering all possible aspects of the assessment synergies proposed so far. At first the impact of sampling schemes is investigated, separately for each year. Comparative results are shown in

Table 4.8. Consequently, the same evaluation is performed, for the classification algorithms, as shown in Table 4.9. Table 4.10 was explicitly created to demonstrate the detection abilities over defaulted companies. Finally, in order to facilitate the synergies impact over the assessment performance AUC scores are provided in Table 4.11.

*Table 4.8. Performance scores values per data sampling approach.*

| | *Accuracy* | | *Sensitivity* | | *Specificity* | | *Precision* | |
|---|---|---|---|---|---|---|---|---|
| *2007* | **Train** | **Test** | **Train** | **Test** | **Train** | **Test** | **Train** | **Test** |
| *EntireSet* | 0.882 | **0.870** | 0.726 | 0.311 | **0.888** | **0.880** | 0.611 | 0.042 |
| *Kenstone* | 0.791 | 0.757 | 0.738 | 0.464 | 0.800 | 0.763 | 0.619 | 0.052 |
| *Optics extrema* | 0.823 | 0.748 | 0.836 | 0.581 | 0.819 | 0.751 | 0.717 | **0.055** |
| *Random* | 0.770 | 0.644 | 0.854 | 0.649 | 0.736 | 0.644 | 0.656 | 0.055 |
| *SMRS* | **0.906** | 0.402 | 0.922 | 0.909 | 0.755 | 0.393 | **0.974** | 0.031 |
| *Kmeans SMRS* | 0.799 | 0.429 | 0.820 | 0.753 | 0.761 | 0.423 | 0.877 | 0.022 |
| *Kmeans Random* | 0.837 | 0.825 | 0.755 | 0.368 | 0.844 | 0.833 | 0.542 | 0.051 |
| *Optics SMRS* | 0.893 | 0.249 | **0.924** | **0.909** | 0.735 | 0.237 | 0.950 | 0.022 |
| *2008* | | | | | | | | |
| *EntireSet* | 0.860 | 0.834 | 0.754 | 0.298 | **0.864** | **0.853** | 0.600 | 0.077 |
| *Kenstone* | 0.819 | 0.778 | 0.776 | 0.373 | 0.825 | 0.792 | 0.640 | 0.073 |
| *Optics extrema* | 0.843 | 0.757 | 0.864 | 0.537 | 0.835 | 0.765 | 0.734 | 0.098 |
| *Random* | 0.815 | 0.702 | 0.845 | 0.517 | 0.803 | 0.709 | 0.698 | 0.072 |
| *SMRS* | **0.922** | 0.451 | 0.934 | 0.872 | 0.805 | 0.436 | **0.980** | 0.061 |
| *Kmeans SMRS* | 0.849 | 0.397 | 0.925 | 0.815 | 0.724 | 0.381 | 0.859 | 0.047 |
| *Kmean sRandom* | 0.848 | **0.818** | 0.764 | 0.347 | 0.855 | 0.835 | 0.571 | **0.127** |
| *Optics SMRS* | 0.910 | 0.240 | **0.950** | **0.911** | 0.700 | 0.216 | 0.946 | 0.041 |
| *2009* | | | | | | | | |
| *EntireSet* | 0.864 | **0.860** | 0.766 | 0.323 | **0.869** | **0.871** | 0.573 | 0.080 |
| *Kenstone* | 0.805 | 0.770 | 0.785 | 0.414 | 0.809 | 0.777 | 0.600 | 0.039 |
| *Optics extrema* | 0.831 | 0.751 | 0.883 | 0.625 | 0.807 | 0.753 | 0.729 | 0.059 |
| *Random* | 0.817 | 0.706 | 0.838 | 0.550 | 0.807 | 0.709 | 0.727 | 0.040 |
| *SMRS* | **0.929** | 0.416 | 0.939 | 0.895 | 0.806 | 0.406 | **0.985** | 0.034 |
| *kmeansSMRS* | 0.863 | 0.362 | 0.933 | 0.916 | 0.718 | 0.350 | 0.879 | 0.029 |
| *kmeansRandom* | 0.846 | 0.830 | 0.782 | 0.414 | 0.854 | 0.839 | 0.556 | **0.088** |
| *Optics SMRS* | 0.923 | 0.265 | **0.951** | **0.945** | 0.726 | 0.251 | 0.962 | 0.026 |

Data set compression is crucial especially for large data sets. In our case, data compression was feasible and meaningful, i.e. there is an acceptable trade-off among initialization time, data variance and model performance.

Table 4.8 suggest so, since sampling approaches achieve similar performance to a traditional training approach, by selecting few data. The combination of OPTICS with SMRS provide a good tradeoff between data sample size and models performance.

*Table 4.9. Performance scores values per data classification technique.*

| | *Accuracy* | | *Sensitivity* | | *Specificity* | | *Precision* | |
|---|---|---|---|---|---|---|---|---|
| *2007* | **Train** | **Test** | **Train** | **Test** | **Train** | **Test** | **Train** | **Test** |
| *Ctree* | 0.951 | 0.674 | 0.956 | 0.488 | 0.974 | 0.678 | 0.890 | 0.040 |
| *FFnet* | 0.918 | 0.681 | 0.590 | 0.474 | 0.846 | 0.685 | 0.892 | 0.032 |
| *Harmonic* | 1.000 | 0.672 | 1.000 | 0.550 | 1.000 | 0.674 | 1.000 | 0.077 |
| *knn* | 0.802 | 0.606 | 0.944 | 0.774 | 0.771 | 0.603 | 0.621 | 0.051 |
| *LDS* | 1.000 | 0.688 | 1.000 | 0.528 | 1.000 | 0.691 | 1.000 | 0.039 |
| *Lin SVMs* | 0.730 | 0.668 | 0.748 | 0.788 | 0.721 | 0.665 | 0.528 | 0.045 |
| *LinReg* | 0.848 | 0.671 | 0.438 | 0.450 | 0.713 | 0.675 | 0.716 | 0.044 |
| *Poly SVMs* | 0.465 | 0.300 | 0.836 | 0.833 | 0.312 | 0.291 | 0.460 | 0.019 |
| *Rbf SVMs* | 0.714 | 0.650 | 0.709 | 0.764 | 0.700 | 0.648 | 0.509 | 0.040 |
| *Adaboost knn* | 0.949 | 0.547 | 1.000 | 0.533 | 0.885 | 0.547 | 0.817 | 0.027 |
| *2008* | | | | | | | | |
| *Ctree* | 0.979 | 0.682 | 0.993 | 0.509 | 0.977 | 0.688 | 0.911 | 0.083 |
| *FFnet* | 0.912 | 0.722 | 0.606 | 0.462 | 0.880 | 0.732 | 0.873 | 0.093 |
| *Harmonic* | 1.000 | 0.680 | 1.000 | 0.496 | 1.000 | 0.687 | 1.000 | 0.119 |
| *knn* | 0.854 | 0.640 | 0.950 | 0.634 | 0.859 | 0.640 | 0.666 | 0.075 |
| *LDS* | 1.000 | 0.729 | 1.000 | 0.458 | 1.000 | 0.739 | 1.000 | 0.091 |
| *Lin SVMs* | 0.761 | 0.670 | 0.775 | 0.713 | 0.744 | 0.669 | 0.543 | 0.077 |
| *LinReg* | 0.858 | 0.661 | 0.464 | 0.434 | 0.713 | 0.669 | 0.788 | 0.049 |
| *Poly SVMs* | 0.519 | 0.250 | 0.970 | 0.933 | 0.219 | 0.225 | 0.423 | 0.042 |
| *Rbf SVMs* | 0.752 | 0.647 | 0.760 | 0.687 | 0.753 | 0.645 | 0.539 | 0.069 |
| *Adaboost knn* | 0.946 | 0.541 | 1.000 | 0.511 | 0.870 | 0.542 | 0.792 | 0.045 |
| *2009* | | | | | | | | |
| *Ctree* | 0.963 | 0.673 | 0.977 | 0.585 | 0.973 | 0.674 | 0.899 | 0.058 |
| *FFnet* | 0.885 | 0.673 | 0.641 | 0.534 | 0.830 | 0.676 | 0.778 | 0.045 |
| *Harmonic* | 1.000 | 0.662 | 1.000 | 0.577 | 1.000 | 0.663 | 1.000 | 0.111 |
| *knn* | 0.857 | 0.598 | 0.969 | 0.722 | 0.836 | 0.596 | 0.677 | 0.047 |
| *LDS* | 1.000 | 0.706 | 1.000 | 0.511 | 1.000 | 0.710 | 1.000 | 0.044 |
| *Lin SVMs* | 0.747 | 0.702 | 0.784 | 0.719 | 0.729 | 0.702 | 0.576 | 0.050 |
| *LinReg* | 0.851 | 0.649 | 0.489 | 0.463 | 0.686 | 0.652 | 0.734 | 0.037 |
| *polynomial SVMs* | 0.585 | 0.334 | 0.932 | 0.974 | 0.334 | 0.321 | 0.475 | 0.029 |
| *Rbf SVMs* | 0.768 | 0.658 | 0.804 | 0.739 | 0.752 | 0.656 | 0.581 | 0.045 |
| *Adaboost knn* | 0.943 | 0.545 | 1.000 | 0.528 | 0.856 | 0.545 | 0.792 | 0.029 |

Table 4.9 suggests that SVMs and neural networks perform better than most classification approaches. However, neural network topology was not optimized, which could lead to further classification performance improvement. In this work, emphasis is given over the detection of defaulted companies. Thus, an appropriate performance metric would be the false negative rate, which demonstrates how the misclassification of defaulted firms as non-defaulted. According to Table 4.10 SVMs score better than the other approaches.

*Table 4.10. False negative rates results; the smaller the value the better the model's performance. Average values for the years 2007-2009.*

| | **Entire Set** | | **KenStone** | | ***k means* *Random*** | | ***k* means SMRS** | | **OPTICS extrema** | | **OPTICS SMRS** | | **Random** | | **SMRS** | |
|---|---|---|---|---|---|---|---|---|---|---|---|---|---|---|---|---|
| | **Train** | **Test** | **Train** | **Test** | **Train** | **Test** | **Train** | **Test** | **Train** | **Test** | **Train** | **Test** | **Train** | **Test** | **Train** | **Test** |
| **Ctree** | 0.000 | 0.860 | 0.000 | 0.668 | 0.023 | 0.572 | 0.008 | 0.502 | 0.041 | 0.125 | 0.051 | 0.190 | 0.000 | 0.740 | 0.073 | 0.123 |
| **FFnet** | 0.298 | 1.000 | 0.633 | 0.830 | 0.287 | 0.521 | 0.296 | 0.514 | 0.003 | 0.045 | 0.047 | 0.175 | 0.853 | 0.942 | 0.019 | 0.053 |
| **Harmonic** | 0.000 | 0.914 | 0.000 | 0.734 | 0.000 | 0.596 | 0.000 | 0.501 | 0.000 | 0.042 | 0.000 | 0.037 | 0.000 | 0.822 | 0.000 | 0.026 |
| **Knn** | 0.000 | 0.610 | 0.000 | 0.311 | 0.002 | 0.282 | 0.000 | 0.167 | 0.228 | 0.287 | 0.036 | 0.098 | 0.000 | 0.470 | 0.100 | 0.098 |
| **LDS** | 0.000 | 1.000 | 0.000 | 0.724 | 0.000 | 0.420 | 0.000 | 0.527 | 0.000 | 0.111 | 0.000 | 0.231 | 0.000 | 0.920 | 0.000 | 0.072 |
| **Lin SVMs** | 0.206 | 0.232 | 0.305 | 0.357 | 0.219 | 0.231 | 0.242 | 0.292 | 0.148 | 0.182 | 0.266 | 0.277 | 0.209 | 0.250 | 0.253 | 0.259 |
| **Linreg** | 0.000 | 1.000 | 0.330 | 1.000 | 0.571 | 0.598 | 0.663 | 0.702 | 0.000 | 0.000 | 0.073 | 0.107 | 0.989 | 0.992 | 0.000 | 0.006 |
| **Poly SVMs** | 0.105 | 0.104 | 0.029 | 0.021 | 0.042 | 0.037 | 0.036 | 0.028 | 0.057 | 0.084 | 0.039 | 0.354 | 0.041 | 0.044 | 0.015 | 0.021 |
| **Rbf SVMs** | 0.237 | 0.270 | 0.372 | 0.397 | 0.246 | 0.293 | 0.297 | 0.358 | 0.204 | 0.202 | 0.227 | 0.244 | 0.237 | 0.276 | 0.122 | 0.122 |
| **Adaboost knn** | 0.000 | 0.904 | 0.000 | 0.788 | 0.000 | 0.638 | 0.000 | 0.689 | 0.000 | 0.000 | 0.000 | 0.006 | 0.000 | 0.780 | 0.000 | 0.000 |

*Table 4.11. AUC performance (test set) for the suggested models.*

| | **Entire Set** | **Kenstone** | **k means Random** | **k means SMRS** | **OPTICS extrema** | **OPTICS SMRS** | **Random** | **SMRS** |
|---|---|---|---|---|---|---|---|---|
| Ctree | **0.555** | **0.602** | **0.655** | **0.650** | **0.631** | **0.577** | **0.602** | **0.559** |
| **2007** | 0.553 | 0.651 | 0.612 | 0.635 | 0.636 | 0.535 | 0.557 | 0.521 |
| **2008** | 0.544 | 0.568 | 0.638 | 0.652 | 0.660 | 0.568 | 0.575 | 0.542 |
| **2009** | 0.568 | 0.585 | 0.713 | 0.663 | 0.596 | 0.630 | 0.673 | 0.615 |
| FFnet | **0.712** | **0.753** | **0.795** | **0.739** | **0.747** | **0.670** | **0.787** | **0.640** |
| **2007** | 0.612 | 0.794 | 0.815 | 0.762 | 0.753 | 0.584 | 0.792 | 0.681 |
| **2008** | 0.738 | 0.699 | 0.766 | 0.706 | 0.700 | 0.678 | 0.779 | 0.612 |
| **2009** | 0.785 | 0.767 | 0.805 | 0.748 | 0.790 | 0.747 | 0.790 | 0.626 |
| Harmonic | **0.735** | **0.728** | **0.798** | **0.741** | **0.787** | **0.758** | **0.784** | **0.703** |
| **2007** | 0.832 | 0.798 | 0.850 | 0.797 | 0.828 | 0.814 | 0.844 | 0.745 |
| **2008** | 0.665 | 0.644 | 0.749 | 0.678 | 0.753 | 0.701 | 0.730 | 0.665 |
| **2009** | 0.707 | 0.742 | 0.797 | 0.748 | 0.780 | 0.759 | 0.778 | 0.700 |
| kNN | **0.663** | **0.685** | **0.755** | **0.706** | **0.746** | **0.651** | **0.700** | **0.592** |
| **2007** | 0.711 | 0.764 | 0.795 | 0.751 | 0.767 | 0.652 | 0.743 | 0.592 |
| **2008** | 0.613 | 0.635 | 0.722 | 0.660 | 0.737 | 0.644 | 0.655 | 0.595 |
| **2009** | 0.664 | 0.657 | 0.747 | 0.709 | 0.733 | 0.655 | 0.702 | 0.588 |
| LDS | **0.785** | **0.695** | **0.728** | **0.680** | **0.777** | **0.709** | **0.779** | **0.718** |
| **2007** | 0.822 | 0.721 | 0.722 | 0.657 | 0.808 | 0.740 | 0.773 | 0.693 |
| **2008** | 0.773 | 0.692 | 0.686 | 0.693 | 0.779 | 0.700 | 0.796 | 0.691 |
| **2009** | 0.760 | 0.671 | 0.776 | 0.691 | 0.743 | 0.687 | 0.768 | 0.769 |
| Lin SVMs | **0.826** | **0.765** | **0.827** | **0.765** | **0.811** | **0.718** | **0.825** | **0.680** |
| **2007** | 0.853 | 0.799 | 0.853 | 0.790 | 0.828 | 0.741 | 0.853 | 0.672 |
| **2008** | 0.807 | 0.736 | 0.807 | 0.742 | 0.813 | 0.679 | 0.806 | 0.650 |
| **2009** | 0.817 | 0.762 | 0.820 | 0.762 | 0.793 | 0.734 | 0.815 | 0.717 |
| LinReg | **0.817** | **0.746** | **0.821** | **0.758** | **0.801** | **0.733** | **0.815** | **0.739** |
| **2007** | 0.840 | 0.791 | 0.841 | 0.808 | 0.815 | 0.755 | 0.838 | 0.739 |
| **2008** | 0.800 | 0.707 | 0.803 | 0.719 | 0.795 | 0.699 | 0.798 | 0.686 |
| **2009** | 0.809 | 0.739 | 0.820 | 0.745 | 0.793 | 0.745 | 0.808 | 0.790 |
| Poly SVMs | **0.752** | **0.715** | **0.752** | **0.721** | **0.745** | **0.688** | **0.750** | **0.687** |
| **2007** | 0.787 | 0.764 | 0.786 | 0.775 | 0.762 | 0.734 | 0.786 | 0.706 |
| **2008** | 0.726 | 0.696 | 0.725 | 0.695 | 0.722 | 0.659 | 0.724 | 0.656 |
| **2009** | 0.744 | 0.684 | 0.746 | 0.693 | 0.752 | 0.672 | 0.741 | 0.700 |
| Rbf SVMs | **0.800** | **0.759** | **0.800** | **0.754** | **0.803** | **0.736** | **0.800** | **0.727** |
| **2007** | 0.820 | 0.792 | 0.820 | 0.783 | 0.820 | 0.757 | 0.822 | 0.720 |
| **2008** | 0.781 | 0.729 | 0.779 | 0.735 | 0.801 | 0.705 | 0.781 | 0.687 |
| **2009** | 0.799 | 0.755 | 0.802 | 0.744 | 0.789 | 0.745 | 0.796 | 0.775 |
| Adaboost knn | **0.619** | **0.631** | **0.614** | **0.591** | **0.628** | **0.612** | **0.638** | **0.576** |
| **2007** | 0.641 | 0.695 | 0.651 | 0.617 | 0.621 | 0.554 | 0.666 | 0.595 |
| **2008** | 0.596 | 0.591 | 0.632 | 0.615 | 0.608 | 0.657 | 0.605 | 0.578 |
| **2009** | 0.620 | 0.607 | 0.559 | 0.542 | 0.653 | 0.626 | 0.643 | 0.556 |

## 4.6 Conclusions and future work

Appropriate train data sets greatly improve the classification performance of any model. Sampling approaches results were exceptionally, despite the heuristic rules over parameters' definition. Further research regarding parameters' set up will lead to even better performance. Additionally, there are many possibilities in exploiting unlabeled data, as long as there are good features available.

These positive preliminary results indicate that there is much room for future research that has the potential to provide many new capabilities and insights into credit risk modeling. A first, obvious, direction would be to employ a richer set of predictor attributes taking among others into account variables related to the business sector of the firms, variables related to non-financial characteristics of the firms (e.g., age, board member composition), stock market data, corporate governance indicators, macroeconomic variables, as well as variables indicating the dynamics of the financial data of the firms (e.g., growth ratios).

It is also necessary to examine the applicability of this modeling approach to developed international markets and consider the relationship of the results in comparison to credit ratings issued by major rating agencies. Finally, it is worth to investigate possible additional effects related to the recent debt crisis and other events that had significant impact on the international markets.

# *Chapter* V: Vision Based Tunnel Inspection

*The real problem is not whether machines think but whether men do.*

*B. F. Skinner, American psychologist*

## 5 Deep-learning, vision-based tunnel inspection

In this chapter, we consider the detection of tunnel concrete defections through visual cues. It is a typical two class identification problem, where we investigate the performance of various approaches, using the raw input from a single monocular camera.

In particular, we exploit a Convolutional Neural Network (CNN) to construct high-level features and as a classifier we choose to use a Multi-Layer Perceptron (MLP) due to its global function approximation properties. Following the aforementioned approach, our method achieves real-time predictions due to the feed-forward nature of CNNs and MLPs. The CNN is evaluated against a variety of machine learning approaches, including SSL techniques.

**The tested SSL paradigms were not suitable for the specific task**, as shown below. **There was three major drawbacks: low feature quality, long execution times and hardware requirements.** In particular, for the problem at hand low level features fail to appropriate describe the defected regions on the image. Also, SSL even for a small size image require the construction of large scale matrices, which is time consuming. Even if we have sparse matrices, their inversion requires a significant amount of RAM and a quick processor.

### 5.1 Introduction

Civil infrastructures are progressively deteriorating (e.g. ageing, environmental factors), calling for inspection and assessment. Presently, structural tunnel inspection is predominantly performed through tunnel-wide visual observations by inspectors, who identify structural defects, rate them and then, based on their severity, categorize the liner. This visual inspection (VI) process is slow, labor intensive and subjective (depending on the experience and fatigue). Additionally, it occurs while working in an unpleasant environment and uncomfortable conditions (Yu et al., 2007). Moreover, the liner condition evaluation is empirical and incomplete and lacks any engineering analysis; it is therefore, unreliable.

Approaches that utilize automated procedures for VI of concrete infrastructures aim specifically to defects detection and structure evaluation. Towards this direction, such methods exploit image processing and machine learning techniques. Initially, low-level image features are used towards the construction of complex handcrafted features, which are used to train learning models, i.e. the detection methods. Automated approaches have been applied in practical settings including roads, bridges, fatigues, and sewer-pipes (Kim and Haas, 2000; Sinha and Fieguth, 2006; Tung et al., 2002).

Recent research in robotics and relevant sectors, such as computer vision and sensors, have significantly increased the competitiveness of components needed in automated systems. Such components can perform quick and robust inspection/assessment, in general transportation and tunnel infrastructures. However, such an integrated and automated system is not yet available. The work that will be presented in this chapter, involves the core mechanism of such system; a computer vision scheme, easy to integrate with all the required components (e.g. laser scanners) for the inspection and assessment of tunnels in one pass.

### 5.1.1 Related work

Intensity features and SVMs for crack detections on tunnel surfaces where used in (Liu et al., 2002). Color properties, different non-RGB color spaces and various machine learning algorithms are also investigated in (Son et al., 2012). Edge detection techniques are applied in (Abdel-Qader et al., 2003) for detecting concrete defects. Edge detection algorithms (i.e. Sobel and Laplacian operators) and graph based search algorithms are also utilized in (Yu et al., 2007) to extract crack information.

An image mosaic technology for detecting tunnels surface defects was further extended in (Mohanty and Wang, 2012). A pothole detection system (Koch and Brilakis, 2011), based on histogram shape-based thresholding and low level texture features, has been used in asphalt pavement images. A concrete spalling measurement system for post-earthquake safety assessments, using template matching techniques and morphological operations, has been proposed by (German et al., 2012).

The exploitation of more sophisticated features has also been proposed. Histograms of Oriented Gradient features and SVMs are utilized in the work of (Halfawy and Hengmeechai, 2014), to support automated detection and classification of pipe defects. Shape-based filtering is exploited in the work of (Jahanshahi et al., 2013) for crack detection and quantification. The constructed features are fed as input to ANN or SVM classifiers in order to discriminate crack from non-crack patterns.

The conventional paradigm of pattern recognition consists of two steps: complex handcrafted feature construction and classifiers training. However, variety in defect types makes difficult the feature construction/selection task. Deep learning models (Hinton et al., 2006; Hinton and Salakhutdinov, 2006) are a class of machines that can learn a hierarchy of features by building complex, high-level features from low-level ones, automating the process of feature construction for the problem at hand.

The work of (Makantasis et al., 2015a) exploits a CNN to hierarchically construct high-level features, describing the defects, and a Multi-Layer Perceptron (MLP) that carries out the defect detection task in tunnels. Such an approach offers an automated feature extraction, adaptability to the defect type(s), and has no need for special set-up for the image acquisition. Nevertheless, there is a major drawback regarding the applicability in real life scenarios: resources spend for data annotation. Data annotation is a time consuming job that requires a human expert; it is therefore prone to segmentation errors.

The aforementioned approach has been further enriched by (Protopapadakis and Doulamis, 2015a); they incorporated a prior, image processing, detection mechanism, facilitating the initialization phase. Such mechanism stands as a simple detector and is only used at the beginning of the inspection. Possible defects are annotated and then validated by an expert; after validating few samples, the required training dataset for the deep learning approach has been formed. From this point onwards, a CNN is trained and, then, utilized for the rest of the inspection process.

In their most recent approach (Protopapadakis et al., 2016a) the same initialization mechanism, as in the previous paragraph. However, the CNN detector is directly utilized over raw data. These data are image patches, which allow us to bypass the low level feature extraction process, preserving from wrong feature selection adverse impact. Finally, since most of defects result in crack appearance, the CNN is specifically utilized for crack detection.

## 5.2 Approach overview

We consider the detection of concrete defects in tunnels using monocular camera's RGB images. Seen as an image segmentation problem, the detection of defects entails, to a great extent, the classification of each one of the pixels in the image into one of two classes; defection class and no defection class.

Classification requires the description of pixels by a set of highly discriminative features that fuse visual and spatial information. However, the features extraction is inherently depended on the problem at hand. Such drawback can be eliminated through a hierarchical construction of complex, high-level features, following the deep learning paradigm. Concretely, a CNN is proposed for feature construction, followed by a MLP, which conducts the classification task. While visual information is derived using the RGB values of each pixel, spatial information is obtained by taking into consideration a neighborhood around each pixel.

Fusing spatial and visual information is aligned with the exploitation of CNNs, which, typically, handle 2D inputs. Instead of using as input to the network a single pixel $p_{xy}$, located at $(x, y)$ point on image plane, we feed it with a patch centered at point $(x, y)$. Particularly, in order to classify a pixel $p_{xy}$ and successfully fuse visual and spatial information, we use a square patch of size $s \times s$ centered at pixel $p_{xy}$.

If we denote as $l_{xy}$ the class label of the pixel at location $(x, y)$ and as $b_{xy}$ the patch centered at pixel $p_{xy}$, then, we can form a dataset $D = \{b_{xy}, l_{xy}\}$ for $x = 1, \dots, w$ and $y = 1, \dots, h$; parameters $w$ and $h$ refer to image width and height respectively. Patch $b_{xy}$ is 3D tensor, who is divided into $c$ matrices of dimensions $s \times s$ (2D inputs) which are fed as input into a CNN. Parameter $c$ refers to the channels of visual information, i.e. for a typical RGB image, $c = 3$. Then, the CNN hierarchically builds high-level features that encode visual and spatial characteristics of pixel $p_{xy}$.

The output of the CNN is sequentially connected with a MLP. Thus, obtained features are used as input by the MLP classifier, which is responsible for detecting defections. The overall architecture of the CNN is shown in Figure 5.1.

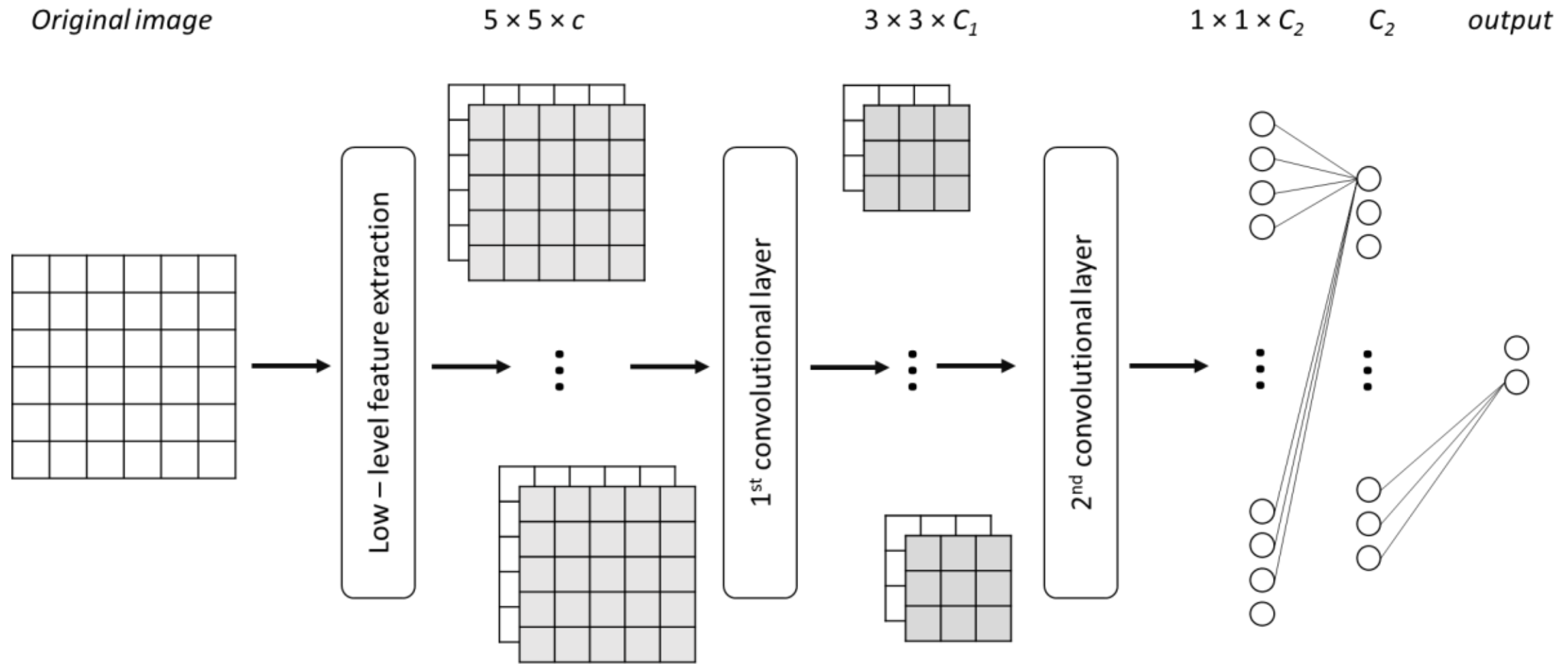

*Figure 5.1. Proposed CNN detector.*

### 5.2.1 Data acquisition and processing challenges

Ideally, during image acquisition, no special setup should take place; we aim towards the development of a generic optic inspection method. In other words, images should be taken from any angle and distance from the tunnel surfaces. However, even if we have the ideal conditions, during acquisition, the defect types make the problem increasingly difficult for the detection mechanism. The term "defect" can be interpreted in many ways; deformations, cracks, surface disintegration, and other defects are widely known and commonly appear.

Discreet, parallel cracks that look like tearing of the surface are caused by shrinkage while the concrete is still fresh, called plastic shrinkage cracks. Fine random cracks or fissures that may only be seen when the concrete is drying after being moistened are called crazing cracks. Cracking that occurs in a three-point pattern is generally caused by drying shrinkage. Large pattern cracking, called map-cracking, can be caused by alkali-silica reaction within the concrete. Structural failure cracking may look like many other types of cracking; however, in slabs they are often associated with subsequent elevation changes, where one side of the crack is be lower than the other.

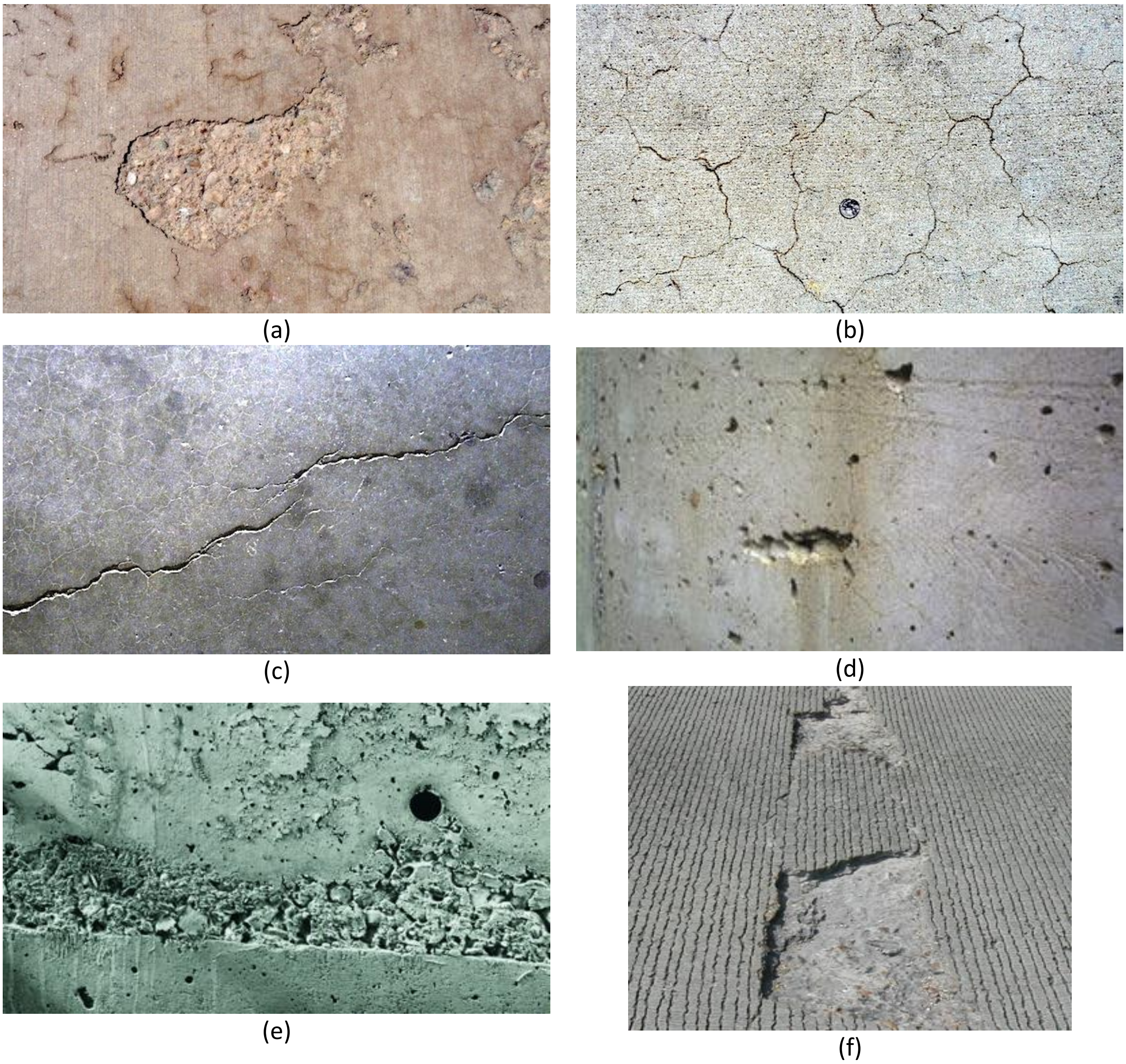

Figure 5.2. Illustration of various defect types: (a) small patches of flaking, (b) map-cracking, (c) shrinkage crack, (d) bugholes, (e) honeycombing, and (f) raveling or spalling at joints. These are some of the major defect types, appear in almost all concrete infrastructures.

Disintegration of the surface is generally caused by three types of distress: (a) dusting, due to carbonation of the surface by unventilated heaters or by applying water during finishing, (b) ravelling or spalling at joints, when pieces of concrete from the joint edges are dislodged and, (c) breaking of pieces from the surface of the concrete, generally caused by delaminations and blistering. Popouts are conical fragments that come off the surface, typically leaving a broken aggregate at the bottom of the hole. Popoffs, or mortar flaking, is similar to popouts, except that the aggregate is not broken and the broken piece is generally smaller. Flaking of the concrete surface over a widespread area is called scaling.

Other defects include discoloration of the concrete; bugholes, which are small voids in the surface of vertical concrete placements, and honeycombing, which is the presence of large voids in concrete caused by inadequate consolidation. At this point, we are able to understand the defect identification problem: it is extremely difficult to extract features suitable for the accurate description of such a large number of defect alternatives, simultaneously.

### 5.2.2 Visual Information Modeling for Tunnel Inspection

In this section we describe the process for encoding visual information. This process takes place exploiting low-level features. There are two main reasons we used such features. Firstly, similar features were used by many researchers (e.g. (Abdel-Qader et al., 2003; German et al., 2012; Halfawy and Hengmeechai, 2014; Koch and Brilakis, 2011; Son et al., 2012)). Secondly, such low-level features are calculated over raw-data and are computationally less-expensive than other high-level features. It has to be mentioned that these features are used by the CNN to hierarchically construct high-level features. They are not used directly for classification purposes.

Using low-level feature extraction techniques, each pixel $p_{xy}$ is described by a feature vector $\boldsymbol{f}_{xy} = \left[f_{1,xy}, \dots, f_{k,xy}\right]^T$, where $f_{i,xy}$, $i = 1, \dots, k$ are scalars correspond to the presence and magnitude of the low level features detected at location $(x, y)$. Feature vectors along with the class labels of every pixel are used to form a dataset for training, validating and testing our learning model. In the following we describe, which features are used to form feature vector $\boldsymbol{f}_{xy}$.

#### *5.2.2.1 Edges*

In order to successfully exploit images edges, on the one hand the system must be able to detect them, in a very accurate way and, on the other, it must preserve their magnitude. For this reason we combined the Canny and Sobel operators. Canny (McIlhagga, 2011) operator is a very accurate image edge detector, which outputs zeros and ones for image edges absence and presence respectively. On the other hand, Sobel (Yasri et al., 2008) operator measures the strength of detected edges.

Multiplying pixel-wise the output of two operators the system is able to detect edges in a very accurate way, while at the same time it preserves their magnitude. If we denote as $C_I$ and $S_I$ the Canny and Sobel operators for image $I$ then the edges $\boldsymbol{\mathcal{E}}_I$ are defined as:

$$\boldsymbol{\mathcal{E}}_I = C_I \cdot S_I \tag{5.1}$$

Matrix $\mathcal{E}_I$ has the same dimensions with image $I$ and its elements $\boldsymbol{\mathcal{E}}_I(x, y)$ correspond to the magnitude of an image edge at $(x, y)$ location.

#### *5.2.2.2 Frequency*

Frequency feature is utilized to emphasize regions of high frequency in the image and at the same time suppress low frequency regions. Frequency components of $I$ are computed as:

$$\boldsymbol{\mathcal{F}}_I = \nabla^2 \cdot I \tag{5.2}$$

The matrix $\boldsymbol{\mathcal{F}}_I$ has the same dimensions with image I and its elements $\boldsymbol{\mathcal{F}}_I(x, y)$correspond to the frequency's magnitude at location $(x, y)$ on image plane.

#### *5.2.2.3 Entropy*

Image entropy quantifies the information coded in an image, i.e. the amount of information which can be coded by a compression algorithm. Image entropy can be interpreted as a statistical measure of randomness, which can be used to characterize the texture of the input image.

Images that depict large homogeneous regions[7], present low entropy, while highly textured images will present high entropy. Entropy $\mathcal{H}_r$ of a region $r$ of an image is defined as:

$$\mathcal{H}_r = \sum_{j=1}^{k} P_i^{(r)} \cdot \log\left(P_i^{(r)}\right) \tag{5.3}$$

where $P_j^{(r)}$ is the frequency of intensity $j$ in an image region $r$. For a grayscale image, variable k is equal to 256. In order to compute entropy for a pixel $p_{xy}$, we apply eq. (5.3) on a square window centered at $(x, y)$. Applying this relation on every point of an image $I$ results to a matrix $\mathcal{H}_I$, which can be interpreted as a pixel-wise entropy indicator.

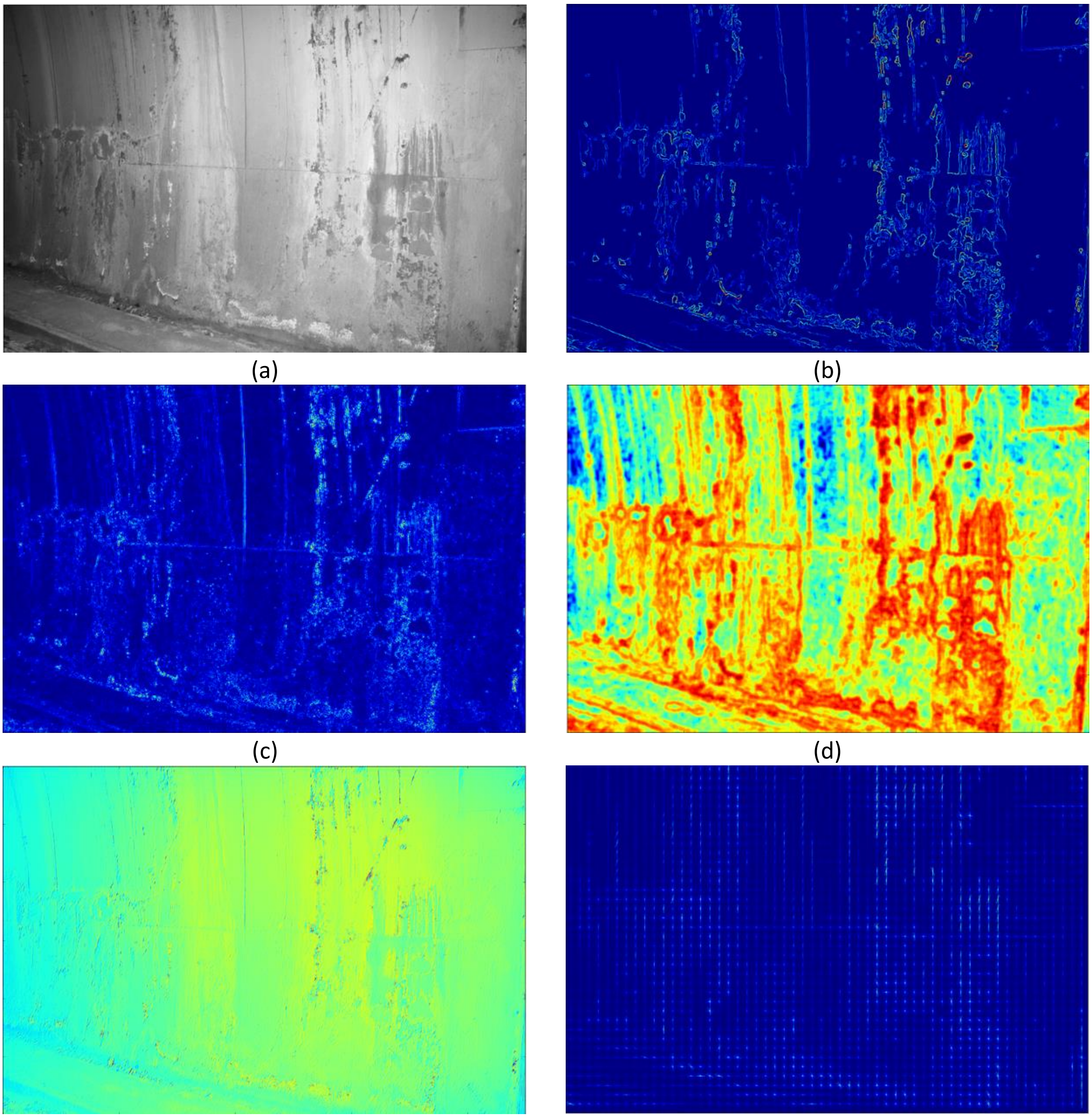

(a) (b)

(c) (d)

[7] Imagine that you have a fresh applied concrete coating; no stripes, holes or other defects. The texture is homogenous and, thus, the image block over this area presents low entropy. Similar examples is a clean sky, or the surface of a lake a calm sunny day.

(e) (f)

*Figure 5.3. Illustration of the extracted low level features: (a) Original image, (b) edges, (c) frequency, (d) entropy, (e) texture and (f) HOG.*

#### *5.2.2.4 Histogram of Oriented Gradients*

HOG is a popular dense feature descriptor (Dalal and Triggs, 2005) used for the task of object detection. It exploits image gradients to capture contour, silhouette and texture information, in order to produce an encoding that is sensitive to local image content while remaining resistant to small changes in pose or appearance.

#### *5.2.2.5 Texture*

For texture identification we used Gabor filters, which is a linear filter used for edge detection. A Gabor filter is characterized by a frequency and an orientation. Frequency and orientation representations of Gabor filters are similar to those of the human visual system, and they have been found to be particularly appropriate for texture representation and discrimination. In the spatial domain, a 2D Gabor filter is a Gaussian kernel function modulated by a sinusoidal plane wave. In our implementation we used Gabor filters with different frequencies and orientations, in order to extract low level features from the image. Specifically, we used twelve Gabor filters with orientations $0°$, $30°$, $60°$and $90°$ and frequencies 0.0, 0.1 and 0.4.

Following the aforementioned procedure we construct 16 low level features for an image. By combining these features with the raw pixels intensity, feature vector $\boldsymbol{f}_{xy}$ takes the form of a $1 \times 17$ vector containing visual information that characterizes each one of the image's pixels.

### 5.2.3 Leaning model

As it has mentioned before, CNNs apply trainable filters and pooling operations on their input resulting in a hierarchy of increasingly complex features. Convolutional layers consist of a rectangular grid of neurons (filters), each of which takes inputs from rectangular sections of the previous layer. Each convolution layer is followed by a pooling layer that subsamples block-wise the output of the precedent convolutional layer and produces a scalar output for each block.

Formally, if we denote the $k$-th output of a given convolutional layer as $h^k$ whose filters are determined by the weights $W^k$ and bias $b^k$, then the $h^k$ is obtained as:

$$h_{ij}^k = \mathrm{g}\left(\left(\mathrm{W^k} * \mathrm{x}\right)_{\mathrm{ij}} + \mathrm{b^k}\right) \tag{5.4}$$

where $x$ stands for the input of the convolutional layer and indices $i$ and $j$ correspond to the location of the input where the filter is applied. Star symbol $(*)$ stands for the convolution operator and $g(\cdot)$ is a non-linear function. Max pooling layers simply take some $k \times k$ region and output the maximum value in that region. For instance, if their input layer is a $N \times N$ matrix, they will then output a $\frac{N}{k} \times \frac{N}{k}$ matrix.

Max pooling layers introduce scale invariance to the constructed features, which is a very important property for object detection/recognition tasks, where scale variability problems may occur. However, For the problem of tunnel defects detection, we involve CNNs to construct features that spatio-visual encode information that indicates the presence or absence of a defection to a specific pixel and thus scale invariance does not consist a significant property for our learning model. Due to this fact, we do not involve pooling layers into learning architecture.

#### *5.2.3.1 Learning model parameterization*

The visual information of each raw image is encoded using the techniques described in sec. 5.2.2, resulting to a 3D tensor of dimensions 17 × $w$ × $h$, where $w$ and $h$ stand for the width and height of the raw input image. This tensor is divided into 1 × 5 × 5 overlapping windows, which are fed as input to the CNN.

The input is convolved with 30 trainable filters of dimension 3 × 3. The output of the first convolutional layer is a 3D matrix of dimension 30 × 3 × 3 since we do not take into consideration the border of the window during convolution. The output of this layer is fed as input to the second convolutional layer and convolved with 60 trainable features of dimension 3 × 3. The output of the second convolutional layer is a 60 dimensional vector, which is fed as input to the MLP. The MLP contains one hidden layer with 60 neurons and an output layer with two neurons (one for each class).

### 5.2.4 Possible limitations

The defects visual complexity varies significantly among inspection sites, as such there is no guarantee that the initial preprocessing (i.e. low level extracted features) suffice as initial (base) descriptors. Also, it requires extra computation effort. Such drawback is addressed in the work of (Protopapadakis and Doulamis, 2015b), where input patches originate from the raw image.

Another issue is the initialization setup; a carefully-annotating image set is required for the detectors initialization, since we have pixel level assessment. Such a process is time consuming and prone to segmentation (human) errors. SSL annotation schemes, as in sec. 8.4 are not applicable in this case due to low descriptive abilities of extracted features.

## 5.3 Performance evaluation

The proposed system was developed on a conventional laptop with i7 CPU, 8GB RAM using MatLab software, and Theano library (Bastien et al., 2012) in Python. The CNN was compared against well-known techniques in pattern recognition (as described in sec. 3.1 ), which are based on handcrafted features. To conduct a fair comparison we used the same features, as in sec. 5.2.2, for each of these techniques.

### 5.3.1 Dataset description

All the images originate from the tunnels of Egnatia motorway in Greece. There was a great diversity of available tunnels for recording. Raw captured tunnel and annotated ground truth images of resolution $800 \times 600$ pixels were provided.

During image acquisition no special setup took place, i.e. images are taken from any angle and distance from the tunnel surface. Since defects spans very few areas, we had to balance the train and test sets. In total, over 100000 samples were classified; evaluation (test) set was 30% of those.

Performance metrics are shown in Table 5.1, for all of the proposed approaches. Metrics description can be found in sec. 3.3, Table 3.1. Illustrations of the models outputs can be found in Figure 5.5. In this case we have two possible classes; cracks or non-cracks, named positive (P) and negative (N) class, respectively. Confusion matrix and calculated performance metrics are explained in sec. 3.3.

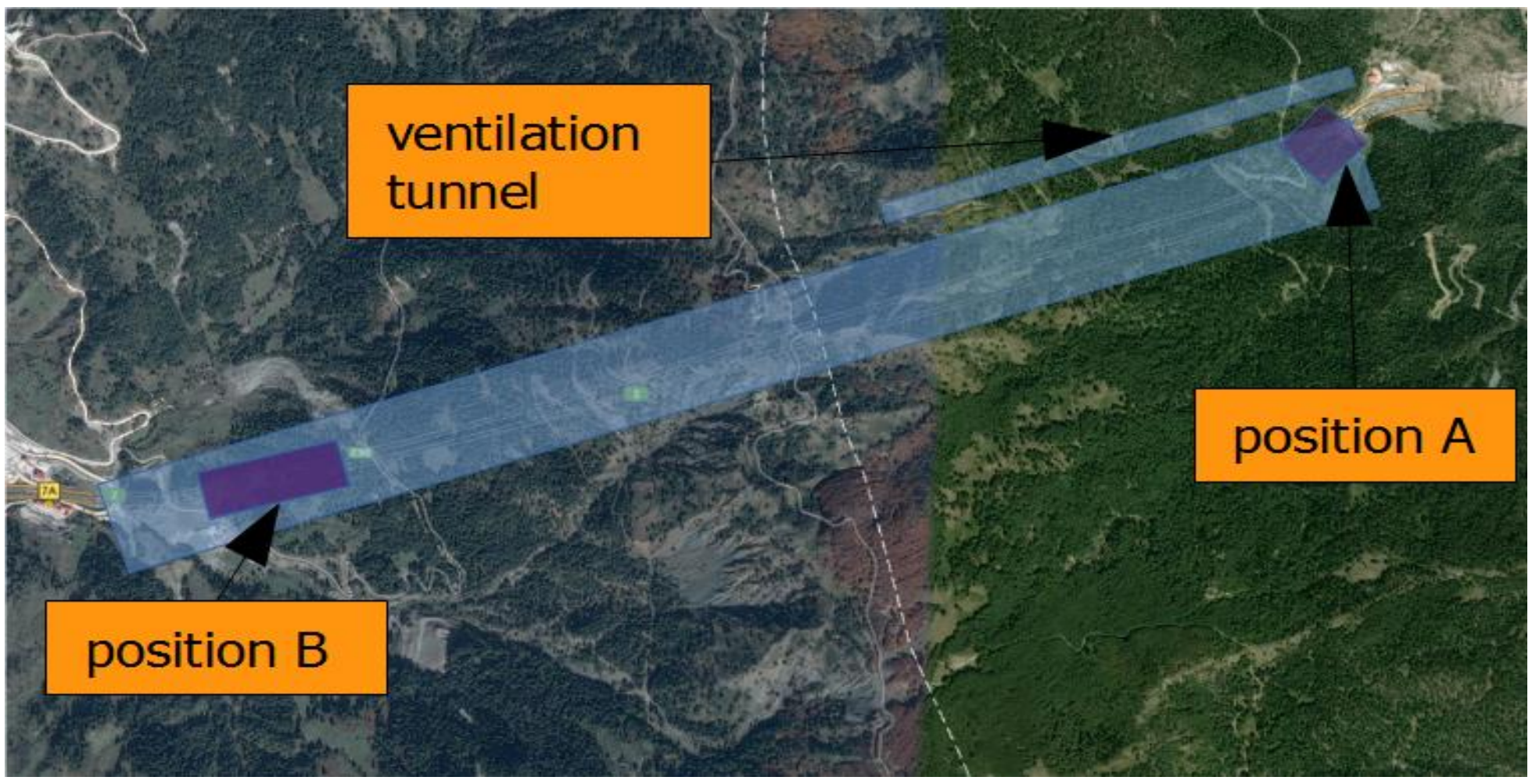

*Figure 5.4. Images from Metsovo tunnels, Egnatia motorway, Greece*

### 5.3.2 Experimental results

Table 5.1 results suggest that for the general defect recognition task low level features are adequate. As such traditional machine learning approaches can support a tunnel inspection process. However, Figure 5.5 suggest against that. The explanation of such contradictory results lies at the features descriptive abilities. Low lever features, which form the training and evaluation sets, are calculated over a set of initial images. Yet, the slightest change in image acquisition process cases significant change in low level feature values. As such, angle, luminosity or focal lenses alters the detector's output.

*Table 5.1. Quantitative performance metrics for the defect recognition problem over RGB images.*

| | Quantitative Performance Metrics | | | | | | | | |
|---|---|---|---|---|---|---|---|---|---|
| | TPR | SPC | PPV | NPV | FPR | FDR | FNR | ACC | F1 score |
| **CNN** | **0,890** | **0,883** | **0,883** | **0,890** | **0,117** | **0,117** | **0,110** | **0,886** | **0,886** |
| **Ctree** | 0,721 | 0,591 | 0,751 | 0,553 | 0,409 | 0,249 | 0,279 | 0,673 | 0,736 |
| **Knn** | 0,845 | 0,575 | 0,773 | 0,685 | 0,425 | 0,227 | 0,155 | 0,746 | 0,807 |
| **Ab kNN** | 0,492 | 0,586 | 0,671 | 0,403 | 0,414 | 0,329 | 0,508 | 0,527 | 0,568 |
| **Ffnn** | 0,854 | 0,554 | 0,766 | 0,689 | 0,446 | 0,234 | 0,146 | 0,743 | 0,808 |
| **Linsvms** | 0,833 | 0,514 | 0,746 | 0,643 | 0,486 | 0,254 | 0,167 | 0,716 | 0,787 |
| **Polysvms** | 0,877 | 0,036 | 0,609 | 0,146 | 0,964 | 0,391 | 0,123 | 0,567 | 0,719 |
| **Rbfsvms** | 0,864 | 0,470 | 0,736 | 0,669 | 0,530 | 0,264 | 0,136 | 0,719 | 0,795 |
| **Harmonic** | 0,668 | 0,534 | 0,710 | 0,485 | 0,466 | 0,290 | 0,332 | 0,619 | 0,689 |
| **LDS** | 0,875 | 0,524 | 0,759 | 0,710 | 0,476 | 0,241 | 0,125 | 0,746 | 0,813 |
| **Anchorgraph** | 0,890 | 0,530 | 0,764 | 0,737 | 0,470 | 0,236 | 0,110 | 0,757 | 0,822 |

A larger data set is not the solution, since the possible combinations of image capturing parameters are infinite; we have not the resources, neither the time required for such a task. Adding additional features does not help since we have the "curse of dimensionality" as illustrated in sec. 2.2.4. The CNN are a good alternative, since the high level features utilized deal with all the aforementioned problems and can be extracted in a rational amount of time. As such they produce far more accurate results.

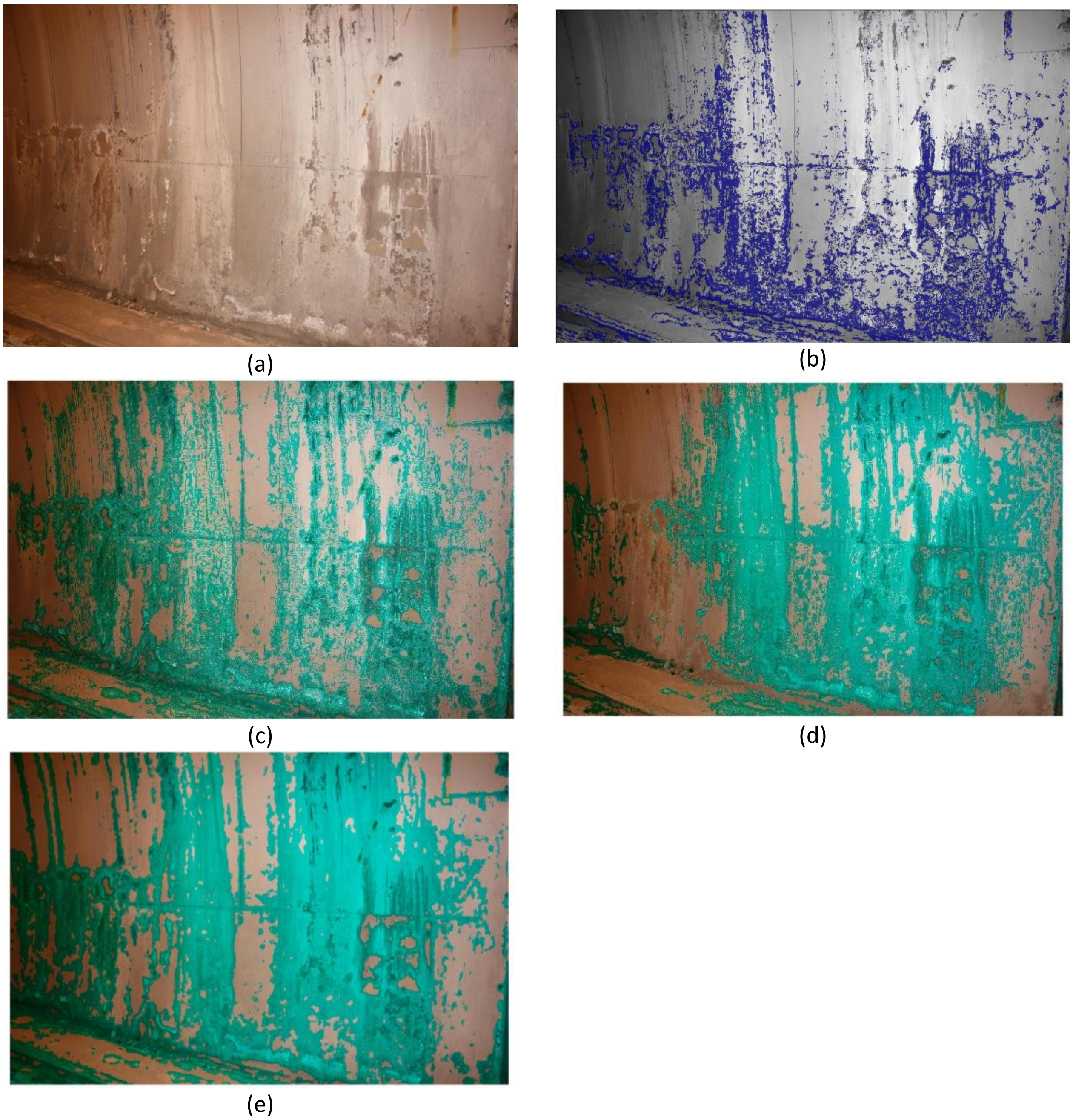

*Figure 5.5. Classification models output. Areas of defect are denoted over the initial image. Top left is the original image (a). The rest are some of the classifiers results: (b) CNN, (c) Ctree, (d) kNN, (e) polynomial SVM.*

## 5.4 Conclusions & future work

In this chapter, we point the suitability for deep learning architectures for the tunnel defect inspection problem. The proposed approach surpass a variety of well-known approaches without making any assumptions on the given images (e.g. minimum resolution, camera angle, etc.). Hierarchical construction of complex high-level features in an automated way result in better classification than the conventional handcrafted features, while at the same time it minimized the feature construction effort, compared to the traditional approaches.

# *Chapter* VI: Industrial Workflow Monitoring

***If everyone is thinking alike, then somebody isn't thinking.***

***George S. Patton, United States Army officer***

## 6 A hybrid, self-trained model for industrial workflow monitoring

In this case, we propose an ANN based scheme for assembly process classification, based on video data taken from Nissan factory. This is a self-trained SSL approach. The procedure is based on (a) a nonlinear classifier, formed using an island genetic algorithm, (b) a similarity-based classifier, and (c) a decision mechanism that utilizes the classifiers' outputs in a semi-supervised way, minimizing the expert's interventions. Such methodology will support the visual supervision of industrial environments by providing essential information to the supervisors and supporting their job.

### 6.1 Introduction

Visual supervision is an important task within complex industrial environments; it has to provide a quick and precise detection of the production and assembly processes. When it comes to smart monitoring of large-scale enterprises or factories, the importance of behavior recognition relates to the safety and security of the staff, to the reduction of bad quality products cost, to production scheduling, as well as, to the quality of the production process.

In most approaches, the goal is either to detect activities, which may deviate from the norm, or to classify some isolated activities (Kim and Ling, 2009; Turaga et al., 2008). Modern techniques are based on supervised training, using large data sets (Kim et al., 2015). The need of a significant amount of labeled data during the training phase makes classifiers data expensive. In addition, that data demands an expert's knowledge that increases further the cost.

Modern industry is based on the flexibility of production lines. Therefore, changes occur constantly. These changes call for appropriate modifications to the supervising systems. A considerable amount of new training paradigms is required in order to adjust the system (Bashir et al., 2007) at the new environment. In order to provide all the training data an expert, whose services will not be at a low-cost, is needed.

A variety of methods has been used for event detection and especially human action recognition, including semi-latent topic models (Wang and Mori, 2009), spatial-temporal context (Hu et al., 2010), optical flow and kinematic features (Ali and Shah, 2010), and random trees and Hough transform voting (Yao et al., 2010). Comprehensive literature reviews regarding isolated human action recognition can be found in (Hu et al., 2004; Poppe, 2010).

In this chapter we focus on creating of a decision support mechanism for the workflow surveillance in an assembly line that would use few training data, initially; as time passes could be self-trained or, if it is necessary, ask for an expert assistance. That way, the human knowledge is incorporated at the minimum possible cost. The innovation can be summarized to the following sentence: *We propose a cognitive system which is able to survey complex, non-stationary industrial processes by utilizing only a small number of training data and using a self-improvement technique through time.*

## 6.2 Proposed methodology

The presented approach employs an innovative self-improvable cognitive system, which is based on a semi-supervised learning strategy as follows: Initially, appropriate visual features are extracted using various techniques. Then, visual histograms are formed, over these features, to address temporal variations in executing different instances of the same industrial workflow. The created histograms are fed as inputs to a non-linear classifier.

The heart of the system is the automatic self-improvable methodology of the classifier. In particular, we start feeding the classifier with a small but sufficient number of training samples (labeled data). Then, the classifier is tested on new incoming unlabeled data. If specific criteria are met, the classifier automatically selects suitable data from the set of the unlabeled data for further training. The criteria are set so that only the most confident unlabeled data will be used on the new training set.

If a vague output occurs, for any of the new incoming unlabeled data, a second classifier, which exploits similarity measure among the in-sampled and the unlabeled data, is used. If classifiers disagree, an expert is called to interweave at the system to improve the classifier accuracy. The intervention is performed, in our case with a totally transparent and hidden way without imposing the user to acquire specific knowledge of the system and the classifier.

### 6.2.1 The island genetic algorithm

The usefulness of the genetic algorithms (GAs) is generally accepted (Whitley et al., 1999). The island GA uses a population of alternative individuals in each of the islands. Every individual is a FFNN. While eras pass networks' parameters are combined in various ways in order to achieve a suitable topology.

A pair of FFNNs (parents) is combined in order to create two new FFNNs (children). Children inherit randomly their topology characteristics from both their parents. Under specific circumstances, every one of these characteristics may change (mutation). The quartet, parents and children, are then evaluated and the two best will remain, updating that way the island's population. An era has passed when all the population members participate in the above procedure. In order to bate the genetic drift, population exchange among the islands, every four eras. The algorithm terminates when all eras have passed. Initially, the parameters' range is described in Table 6.1; the main steps of the genetic algorithm are shown in Figure 6.1. The algorithm is used to parameterize the topology of the non-linear classifier (described in the next session).

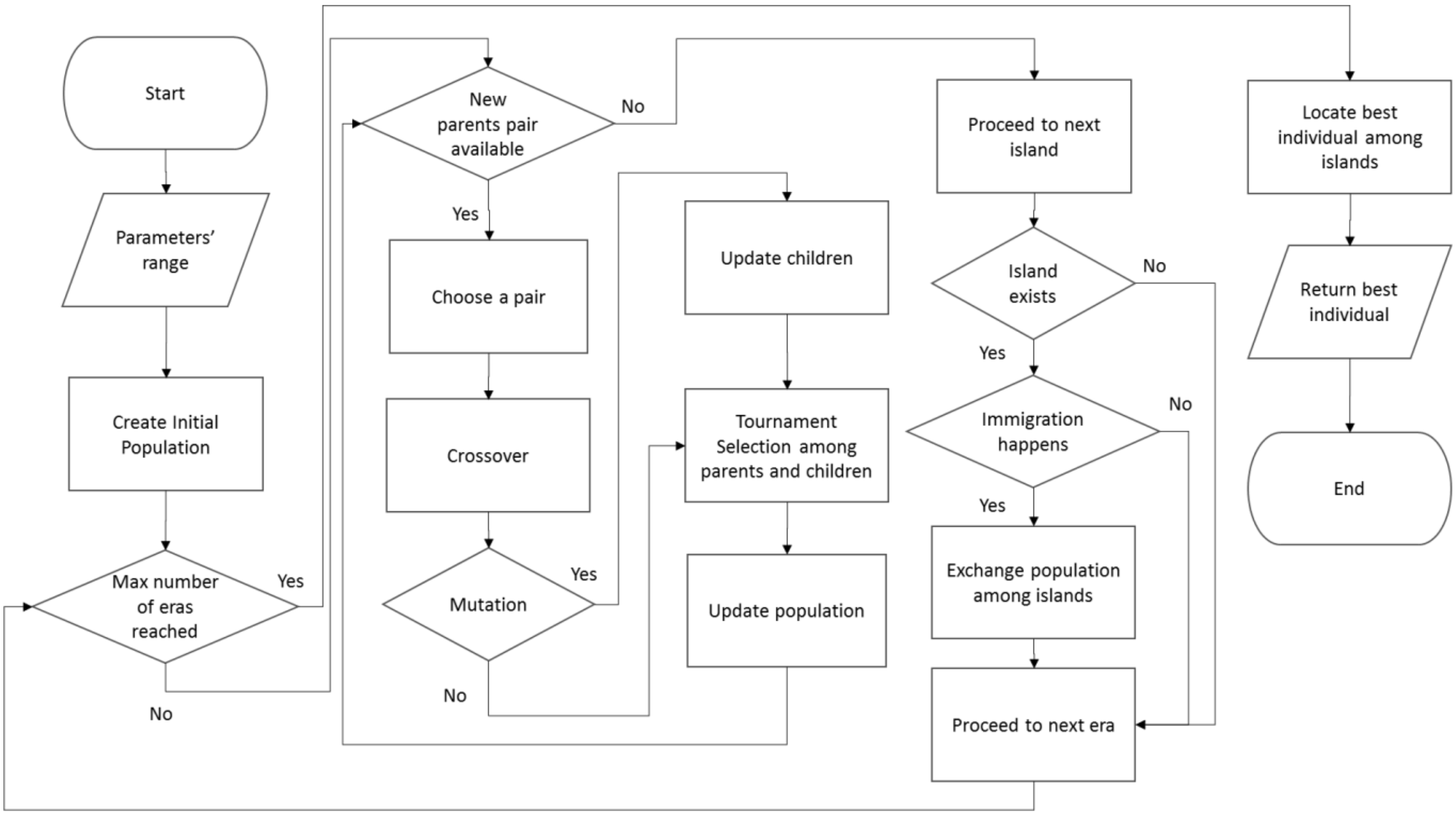

*Figure 6.1. The island genetic algorithm flowchart.*

Regarding the activation functions, the alternatives were five: Hyperbolic tangent or logarithmic sigmoid, saturating linear, hard-limit, and symmetric hard-limit. Individuals may mutate at any era. Mutation can change any of the, previously stated, topology parameters therefore individuals' parameters outside the initially defined range may occur. The fitness of a network is evaluated using the following equation:

$$f_i = \lambda p_i + (1 - \lambda)\alpha \tag{6.1}$$

where $f_i$ denotes the network's fitness score, $p_i$ is the percentage of the correct in-sample classification and $\alpha$ is the average percentage difference, between the two greatest prices, among all the individual's outputs.

*Table 6.1. Island genetic algorithm parameters' range.*

| Parameter | Min value | Max value |
|---|---|---|
| Training epochs | 100 | 400 |
| Number of layers | 1 | 3 |
| Number of neurons (per layer) | 4 | 10 |
| Number of islands | 3 | 3 |
| Number of eras | 10 | 10 |
| Population (per island) | 16 | 16 |

### 6.2.2 The nonlinear classifier

For the proposed approach, the nonlinear classifier is a genetically optimized (i.e. topologically) feed forward neural network, according to the training sample. The neural network's topology is defined by the number of hidden layers, the neurons at each layer, and the activation functions (as described in sec. 3.1.6). All of the above as well as the number of training epochs were optimized using an island genetic algorithm.

Once the training phase is concluded, we start feeding the optimal network unlabeled data. Since the output vector of the classifier contains various values (its actual size is 1×7 as the number of the possible tasks), the output element with the greatest value will be turned into 1 while all the other ones will be set to 0. This is performed only if the greatest value is reliable. The conditions for the reliability are explained at the following section.

### 6.2.3 The semi-supervised approach

The main issue, in order to improve network's performance, is the reliability of labeling the new data, deriving from the pool of the unlabeled ones, exploiting network's performance in the already labeled data. In this approach, output reliability is performed by comparing the absolute value of the greatest output element with the second greatest, according to some criteria. If these criteria are not met, the output is considered vague, otherwise the classifier output is considered as reliable.

An unsupervised algorithm (e.g. $k$-means) is used in case of ambiguous results to support the decision. In particular, the unlabeled input vector that yields the vague output, say $\boldsymbol{u}$, is compared with all the labeled data, say $\boldsymbol{l}_i$, based on a similarity distance and then the distance values are normalized in the range of [0 1] so that all comparisons lie within a pre-defined reference frame, say $d(\boldsymbol{u}, \boldsymbol{l}_i)$. Then, the k-means algorithm is activated to cluster, in an unsupervised way, all the normalized distances $d(\boldsymbol{u}, \boldsymbol{l}_i)$ into a number of classes, equal to the number of available industrial tasks (7 in our case).

In the sequel, the cluster that provides the maximum similarity (highest normalized distance) score, of the unlabeled data that yield the vague output and the labeled ones, is located. Let us denote as *K* the cardinality of this cluster (e.g., the number of its elements). In the following, the neural network output for the given unlabeled datum is linearly transformed according to the following formula:

$$\boldsymbol{n}_f = \boldsymbol{n}_p + \sum_{i=1}^{K} d(\boldsymbol{u}, \boldsymbol{l}_i) \cdot \boldsymbol{v}_i \tag{6.2}$$

where $\boldsymbol{n}_f$ is the modified output vector, $\mathbf{n}_p$ the previous network output before the modification, while $d(\boldsymbol{u}, \boldsymbol{l}_i)$ is the similarity score (distance) for the $i$-th labeled datum $\boldsymbol{l}_i$ and the unlabeled datum $\boldsymbol{u}$ within the cluster of the highest normalized distance, while $\boldsymbol{v}_i$is the neural network output when input is the i-th labeled vector $\boldsymbol{l}_i$ and $K$ is the cardinality of the cluster of the maximum highest similarity.

The modified output vector $\boldsymbol{n}$ which is the base for the decision is created using both manifold, i.e. nearest neighbors similarity mechanism and cluster assumption, i.e. network output indications (Kumar Mallapragada et al., 2009).

### 6.2.4 The decision mechanism

According to the nonlinear classifier output, there are three possible cases (Scenarios):

1. The network made a robust decision that should not be defied. Therefore, the unlabeled data is used for further training but it is not incorporated at the initial training set.
2. The output is fuzzy, in other words, the difference among the two greatest prices does not exceed the threshold values. The similarity-based classifier is activated. If both systems indicate the same then the unlabeled data is used for further training but it is not incorporated at the initial training set.
3. The two classifiers do not agree. Therefore, an expert is called and specifies where the video should be classified. That video is added to the initial training data set.

The combination of these cases leads to a semi-supervised decision mechanism. Threshold values define which from the above scenarios will occur. The threshold value is defined as the percentage of the difference between the two greatest prices at the output vector. The overall process for the decision making is shown in Figure 6.2.

Initially, the first threshold value is set to 0.6. That value means that if the percentage difference of the two greatest values is above or equal to 60% we will be at scenario No 1. The second threshold value is set to 20%. If the percentage difference of the two greatest values is less than that, the system is unable to make a decision and an expert is needed to interfere. Therefore, scenario No 3 will occur. Any value between these two thresholds activates scenario case No 2.

Since the model is self-trained, the first threshold value does not need to be so strict. The model learns through time, thus a reduction at that value would be acceptable. Nevertheless, at the beginning small threshold value could lead the model to wrong learning. Using simulated annealing method, the threshold descents to a 40% through time.

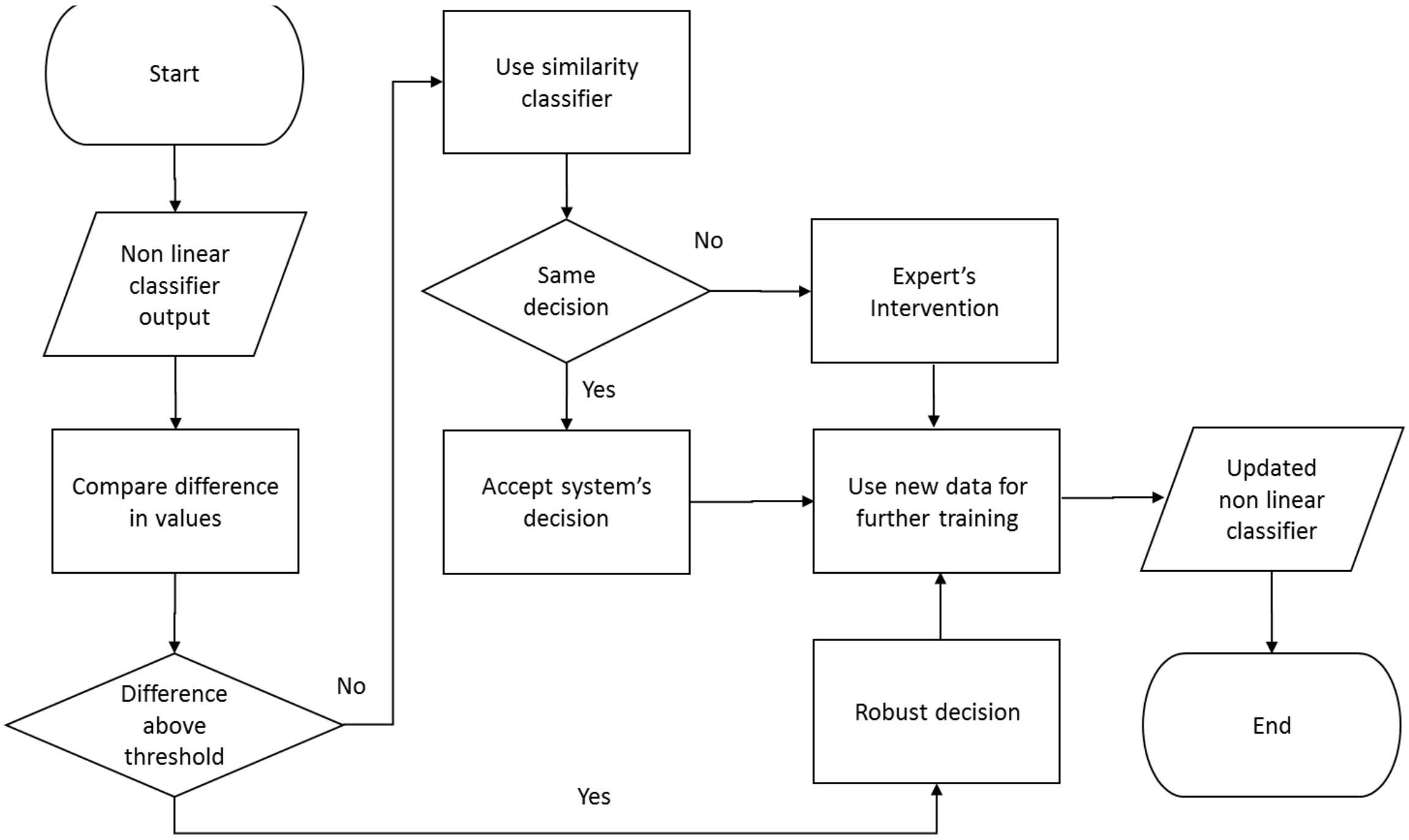

*Figure 6.2. The decision mechanism flowchart.*

### 6.2.5 Possible limitations

Features' characteristics cannot guarantee a smooth performance; due to the similarity in worker movement for many of the activities there will be a trade off in discriminative abilities among tasks. Such a drawback can be solved by either incorporating past frames information (Protopapadakis et al., 2013).

Additionally, system initialization requires descriptive data from all tasks. In different case, the involved classifiers will likely make wrong classification and will retrain the model with them. As such, a significant deterioration in performance is expected. Different camera positions, or data from multiple cameras can further improve the model's performance.

## 6.3 Feature extraction

From all videos, holistic features such as Pixel Change History (PCH) are used. These features remedy the drawbacks of local features, while also requiring a much less tedious computational procedure for their extraction (Ahad et al., 2010). A very positive attribute of such representations is that they can easily capture the history of a task that is being executed. These images can then transformed to a vector-based representation using the Zernike moments (Hwang and Kim, 2006), up to sixth order, in our case, as it was applied in (Kosmopoulos et al., 2011).

The video features, once exported, had a 2 dimensional matrix representation of the form $m \times l$, where $m$ denotes the size of the $1 \times m$ vectors created using Zernike moments, and $l$ the number of such vectors.

Although *m* was constant, *l* varies according to the video duration. In order to create constant size histogram vectors, which would be the system's inputs, the following steps took place:

1. The hyperbolic tangent sigmoid transformation was applied to every video feature. As a result the prices of the 2-d matrices range from -1 to 1.
2. Histogram vectors of 33 classes were created. The number of classes was defined after various simulations. Higher number of classes leads to poor performance due to the small training sample (in our case 48 vectors). Fewer classes also caused poor performance probably due to loss of important information from the original features. Each class counts the frequency of the appearance of a value (within a specific range) for a particular video feature.
3. Finally, each histogram vector value is normalized. Thus, the input vectors were created.

It is clear that each histogram vector describes a specific job among seven different. These histograms, one at a time, are the inputs for a feed forward neural network (FFNN). The target vectors are seven-element arrays. The value at each array will be either one or zero. The number one denotes in which category is categorized the video (e.g., $[0\ 0\ 0\ 1\ 0\ 0\ 0]^T$ corresponds to assembly procedure number four).

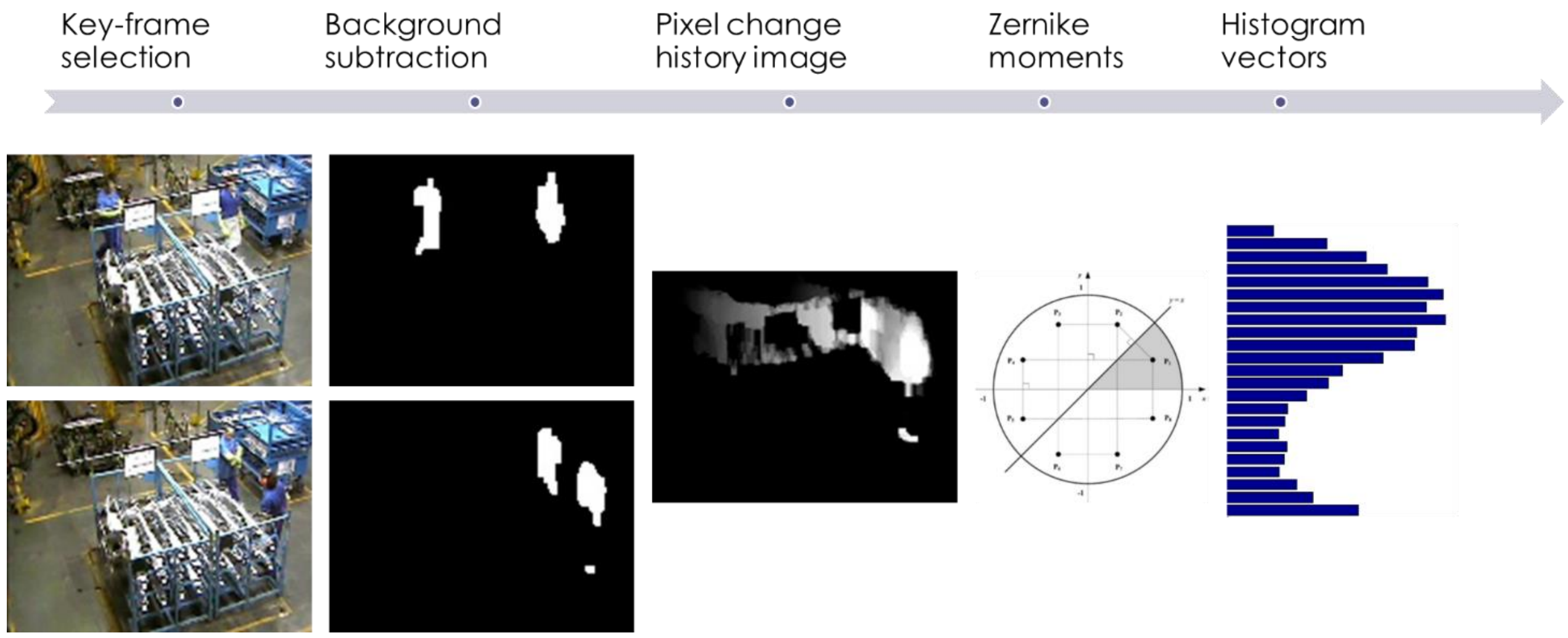

*Figure 6.3. Feature extraction process. Initially, a frame pair is selected and a background subtraction algorithm is applied. Then, a pixel change history image is created and transformed, using Zernike moments, into a histogram vector.*

## 6.4 Performance evaluation

The proposed system was developed on a conventional laptop with i3 CPU, 4GB RAM using MatLab software. The results displayed below are the average numbers after a total of 150 simulations. In every simulation a different set of labeled data was selected.

### 6.4.1 Data set description

The proposed system was tested using the NISSAN video dataset (Voulodimos et al., 2011), which refers to a real-life industrial process videos regarding car parts assembly. Seven different, time-repetitive, workflows have been identified, exploiting knowledge from industrial engineers. Challenging visual effects are encountered, such as background clutter/motion, severe occlusions, and illumination fluctuations.

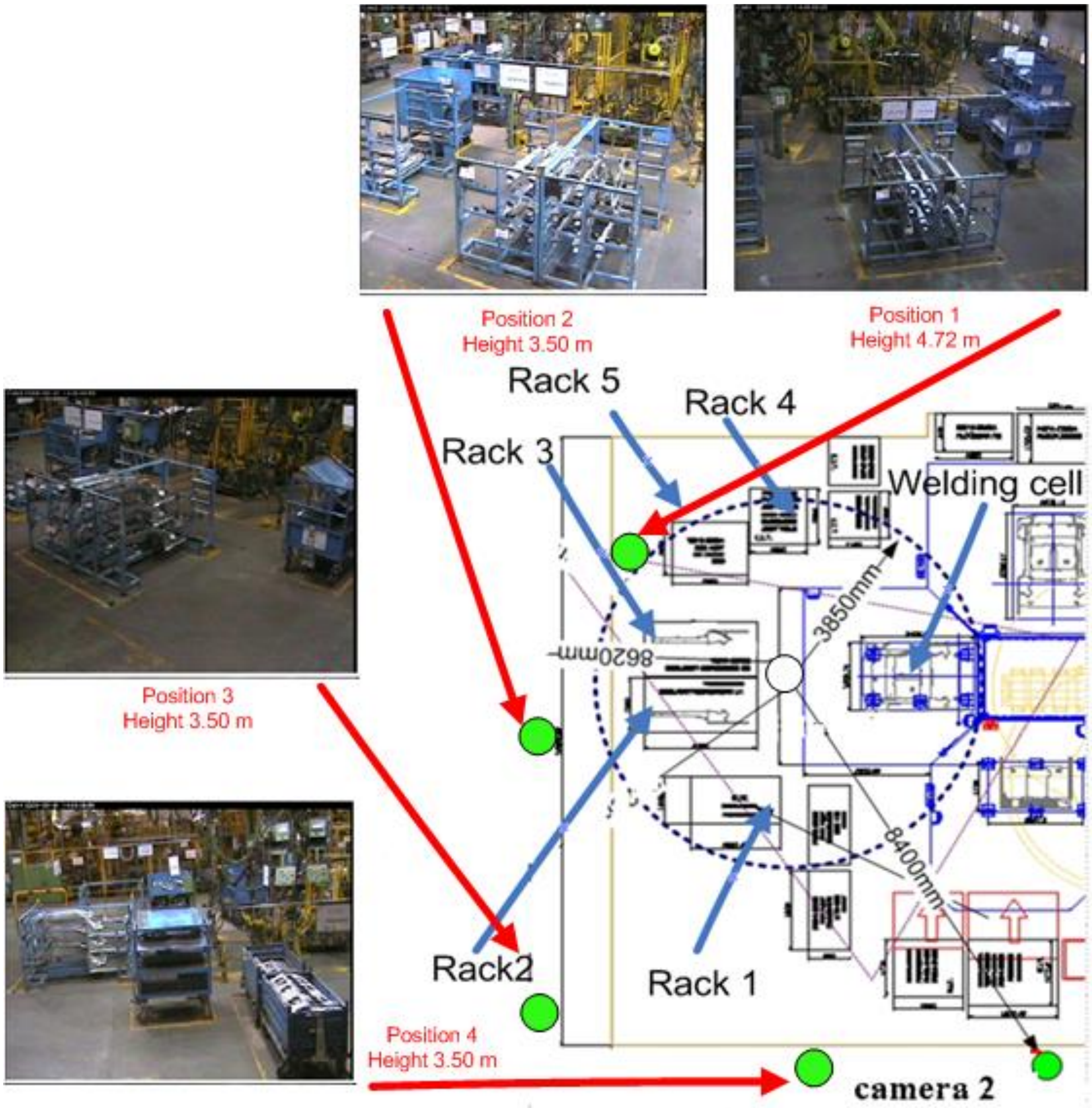

*Figure 6.4. Depiction of a work cell along with the position of camera 1 and the racks #1-5.*

The production cycle on the industrial line included tasks of picking several parts from racks and placing them on a designated cell some meters away, where welding took place. Each of the above tasks was regarded as a class of behavioral patterns that had to be recognized. The behaviors (tasks) we were aiming to model in the examined application are briefly the following:

1. One worker picks part #1 from rack #1 and places it on the welding cell.
2. Two workers pick part #2a from rack #2 and place it on the welding cell.
3. Two workers pick part #2b from rack #3 and place it on the welding cell.
4. One worker picks up parts #3a and #3b from rack #4 and places them on the welding cell.
5. One worker picks up part #4 from rack #1 and places it on the welding cell.
6. Two workers pick up part #5 from rack #5 and place it on the welding cell.
7. Workers were idle or absent (null task).

For each of the above scenarios, 20 videos were available. An illustration of the working facility is shown in Figure 6.4.

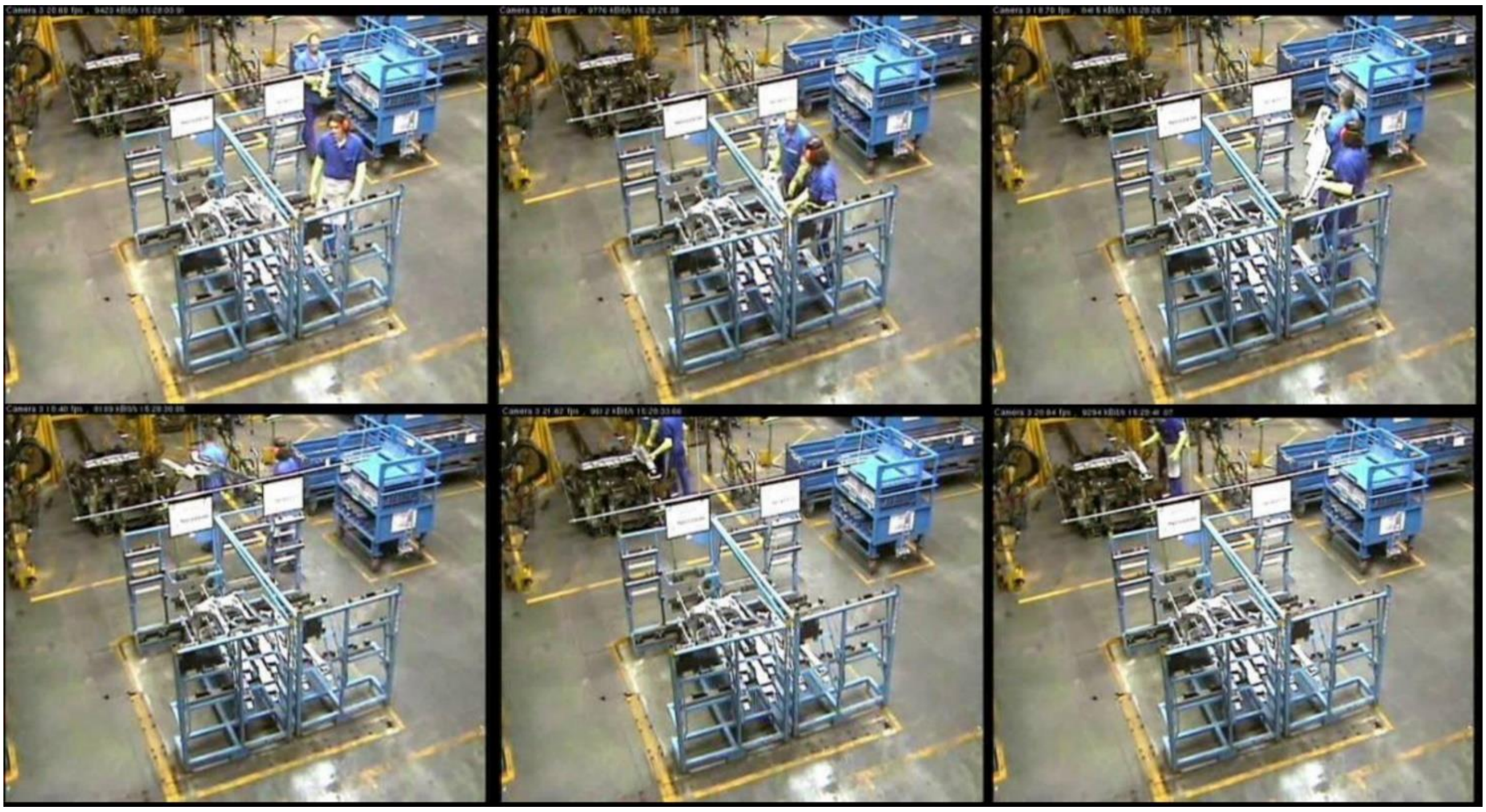

*Figure 6.5. A typical execution of task No 2.*

## 6.4.2 Model initialization and adaptation in new data

Initially, the best possible network is produced using the island genetic algorithm and 40% of the available data. The remaining data are fed to the network, one video at a time, and the overall out of sample performance is calculated. In every case, all the data that activated scenario No 3 is excluded. Then, we reefed the network, one by one, with the rest data. If the network's suggestions were correct it will perform better since more training data (excluding these from scenario No 3) were used for further training. By doing so, the unlabeled data fall below 60% and training data increases further. The above procedure concludes after five iterations. At that time, the ratio between in sample data and out of sample data does not exceed 50%.

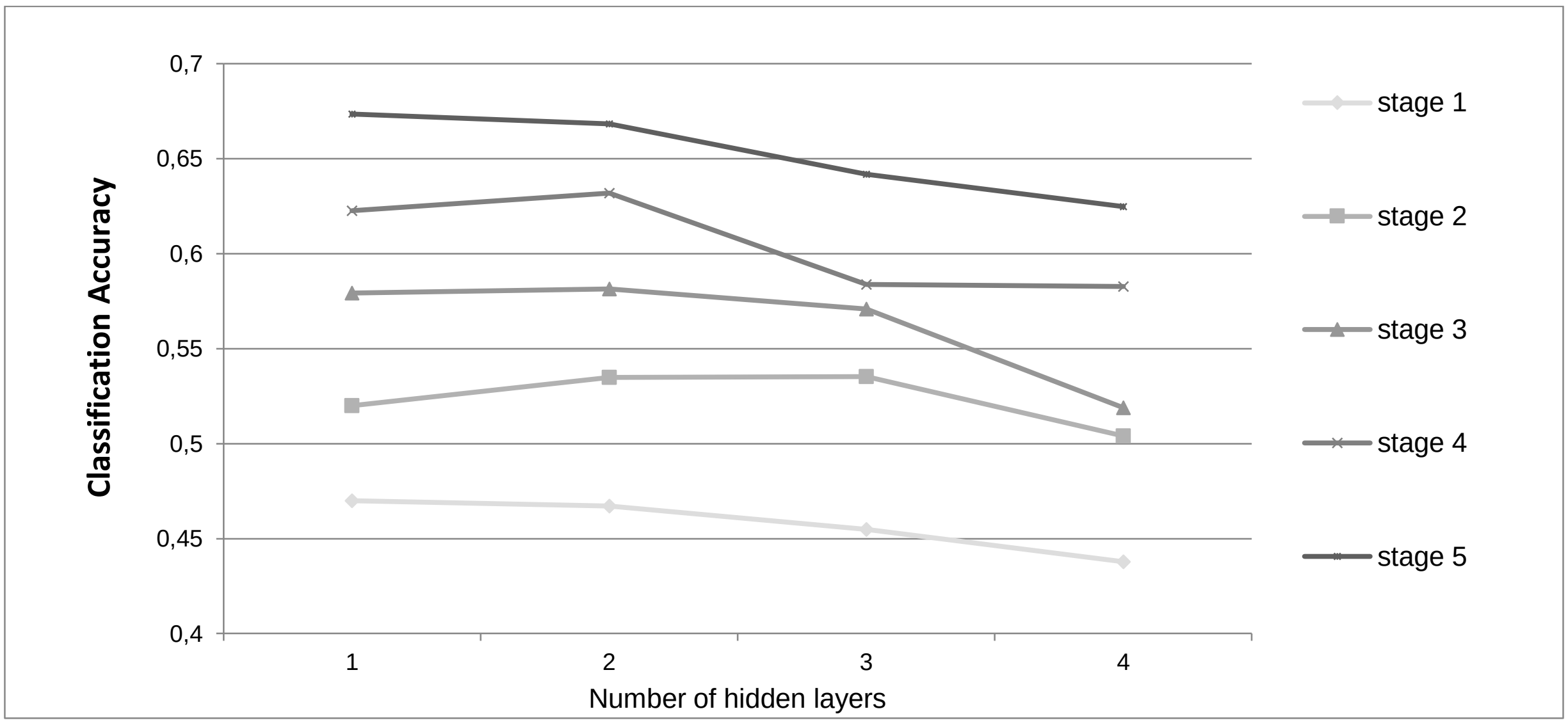

*Figure 6.6. Classification percentages for each of the 5 evaluation stages – test data.*

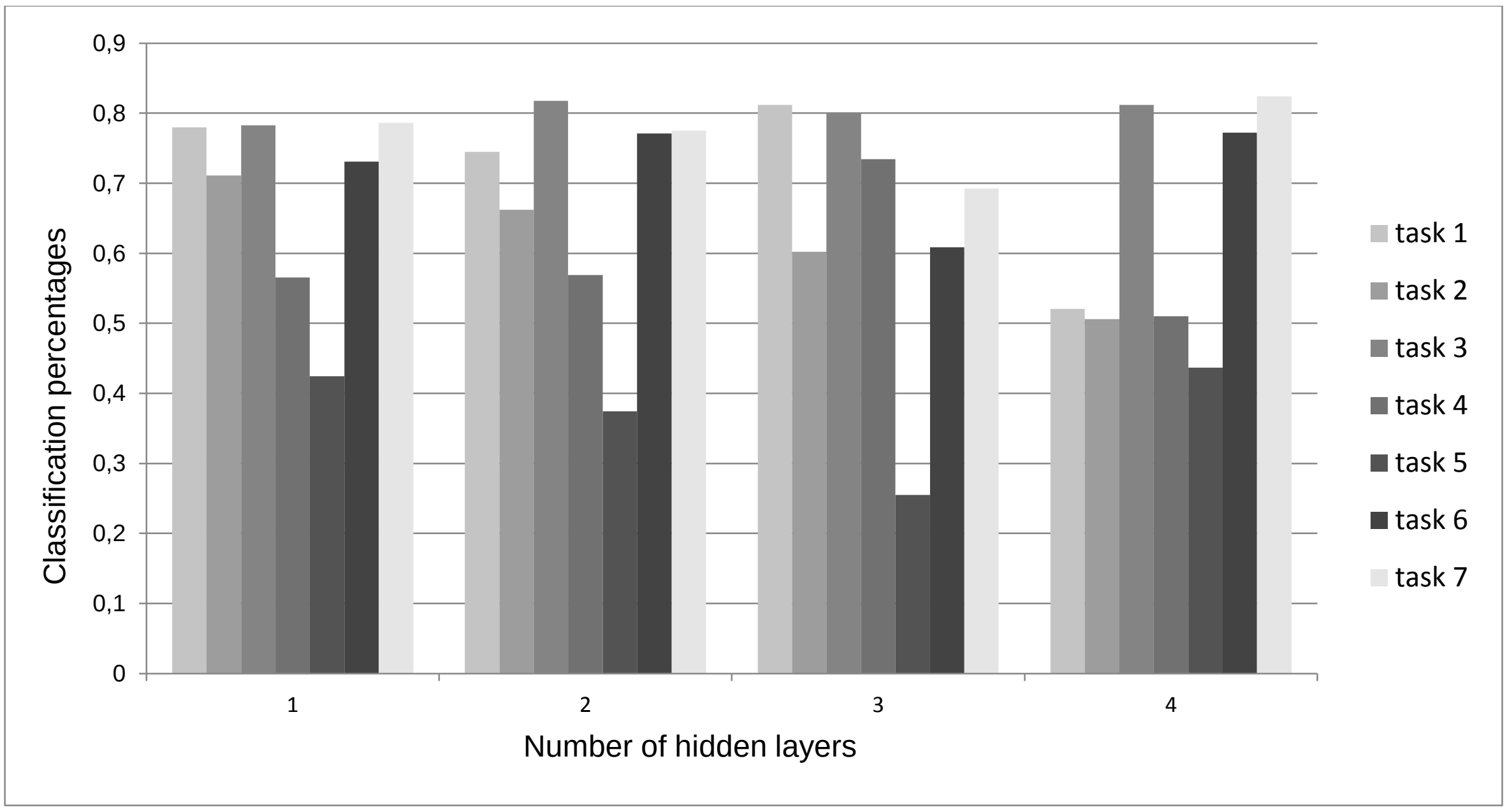

*Figure 6.7 Stage 5 results for each one of the 7 tasks – test data*

### 6.4.3 Performance

It appears that a two hidden layers neural network using hyperbolic tangent sigmoid or log-sigmoid transfer function with an average of nine neurons in each layer is the most suitable solution. The proposed system is able to use the new knowledge to its benefit.

The overall performance increases through iterations, using a small amount of data, as it is shown in Figure 6.6. Actually, by using additionally 10% of the videos, the system reached a 75% correct classification. This is important because the system saves time and resources during the initialization and provides good classification percentages using less than 50% of the available data.

The impact of the training epochs at the overall performance is shown in Figure 6.7. There appear to be a tradeoff between overall and individual task classification. Although 200 up to 300 training epochs provide significant classification accuracy further training increases partially the accuracy only on specific tasks in expense on others.

## 6.5 Conclusions & future Work

In this work, we have proposed a novel framework for behavior recognition in workflows. The above methodology handles with an important problem in visual recognition: it requires a small training sample in order to efficiently categorize various assembly workflows. Such methodology will support the visual supervision of industrial environments by providing essential information to the supervisors and supporting their job.

Improvements at any stage of the system can be made in order to further refine the system's performance. Future work will be based on the usage of different classifiers (e.g. neuro-fuzzy, linear Support Vector Machines) and decision mechanism (e.g. voting-based). In addition, instead of using all frames of a specific task to create classifiers' input, only a subset of them may be used providing equivalent result.

# *Chapter* VII: Supporting the Elder

*If society fits you comfortably enough, you call it freedom.*
*Robert Lee Frost, American poet*

## 7 Life quality improvement for elder people

In this chapter, we create a transductive classifier, in order to identify if we have a person's fall or not. Falls have been reported as the leading cause of injury-related visits to emergency departments and the primary etiology of accidental deaths in elderly. Thus, the development of robust home surveillance systems is of great importance. In the work of (Makantasis et al., 2015b), such a system is presented, which tries to address the fall detection problem through visual cues.

The proposed methodology utilizes a fast, real-time background subtraction algorithm, based on motion information in the scene and pixels intensity, capable to operate properly in dynamically changing visual conditions, in order to detect the foreground object. At the same time, it exploits 3D space's measures, through automatic camera calibration, to increase the robustness of fall detection algorithm which is based on semi-supervised learning approach. The above system uses a single monocular camera.

As such, the system presented in this chapter is characterized by low computational cost and memory requirements, making it suitable for large scale implementations, let alone its low financial cost since simple low resolution cameras are used, making it affordable for large scale implementations.

**SSL is utilized for handling the training data set, supporting a smooth initialization process**. Initially, clusters the data according to various features in order to estimate initial label values. A small subset from each cluster is randomly selected and evaluated by an expert (annotation process), in order to form a training set for the neural classifier. Also, they will be utilized as core samples for the $k$nn classifier, for future comparisons. Given new frame sequences, both classifiers outcomes, are calculated and compared. In case of disagreement an expert is summoned to further investigate.

### 7.1 Introduction

In order to understand the elderly fall problem, and try to prevent fall incidents, someone needs to examine where they occur. Recent studies show that 67% of fall incidents take place inside or in close proximity to patients' home and residential institutions (Government of Canada, 2014), where a medical alert system can be of immediate assistance. Taking into consideration the importance of humans' fall problem and the aforementioned statistics, the development of robust home surveillance systems is necessary. For this reason, a major research effort has been conducted in the recent years for automatically detecting persons' falls.

One common way for automatic fall detection is through the use of specialized devices, such as accelerometers, floor vibration sensors, barometric pressure sensors, gyroscopic sensors, or combination/fusion of them (Le and Pan, 2009; Nyan et al., 2008; Wang et al., 2005; Zigel et al., 2009) or help buttons. However, most of these techniques require specific wearable devices that should be attached to human body. Thus, their efficiency relies on the persons' ability and willingness to wear them. External sensors, such as floor vibration detectors, require a complex setup and are still in their infancy. In the case of a help button, it is useless, if the person is unconscious after the fall.

A more challenging alternative is the use of visual cameras, like the system presented in this chapter, which is however a prime research issue due to the complexity of visual content, i.e. illumination variations, background changes and occlusions, and the fact that a fall incident should be discriminated over other ordinary humans' activities. The emergence of computer vision systems has allowed researchers to overcome the aforementioned problems. Vision based systems are less intrusive, can be installed on buildings and are not worn by users.

Furthermore, cameras can provide a vast amount of information about person and environment making vision based systems suitable for different kind of applications, as they are able to detect several events simultaneously. For example, a vision based system can be used to detect fall incidents, while at the same time, is checking other daily life activities, like medication intake. Although, vision based systems can provide information about human activities, they can preserve persons' privacy by exploiting an event-based design that triggers alarms and/or enables video recording only after the occurrence of specific predefined events. A detailed survey of fall detection methodologies is presented in (Mubashir et al., 2013).

Camera based approaches have also various limitations. Firstly, it is apparent that camera based falls detection is a more challenging process than using other types of sensors, due to occlusion issues, camera resolution, background clutter, visual complexity of the environment, etc. Secondly, camera positioning causes a lot of parallax problems that can affect falls' detection performance. Last but not least, privacy issues should be carefully examined.

### 7.1.1 Place for improvement

A new approach is presented that exploits both 2D image analysis (monocular cameras) and the relationship between camera coordinate system and the real 3D space. Such approach allows fall detection in real-time and in dynamically changing visual conditions, achieving robustness, through camera calibration and inverse perspective mapping, exploiting 3D physical space's measures.

The fall detection scheme consists of a non Linear Warning System (nLWS), based on a semi-supervised learning (SSL) approach. The nLWS utilizes neural networks that are topologically optimized through the use of an Island GA. Once the GA is completed using a small training sample, a self-training procedure is utilized, based on the cluster assumption. The methodology is similar to the work of (Protopapadakis et al., 2012).

In contrast to other 2D fall detection methods, the proposed system is very robust for a wider range of camera positions and mountings and its performance is not affected by the distance between camera and foreground object. Moreover, it extracts 3D features without using computationally expensive stereo-vision mathematics, which are necessary and compulsory for all multiple camera systems.

## 7.2 Proposed methodology

The proposed fall detection mechanism includes three phases: a) foreground/ background extraction and human detection, b) appropriate feature extraction and c) the decision mechanism utilization. System approach is presented in Figure 7.1.

The first step of our algorithm is to detect the persons in a scene for every captured frame. This can be done by applying image segmentation and background subtraction methods. Once the persons have been detected, the system tracks them and estimates their height in order to calculate vertical motion velocity and to analyze their posture.

Once humans have been detected, the system tracks them and estimates their height, in order to calculate vertical motion velocity, and to analyze their posture. Additionally, the primary concern of a fall detection system is to achieve low false negative rates, i.e. its goal is to detect all fall incidents even if some of them are not true (false alarms). Thus, we prefer to exclude the third feature from our analysis.

Finally, a decision mechanism uses the aforementioned features, in order to denote falls and trigger the alarm. The mechanism is initialized offline, using a semi-supervised approach.

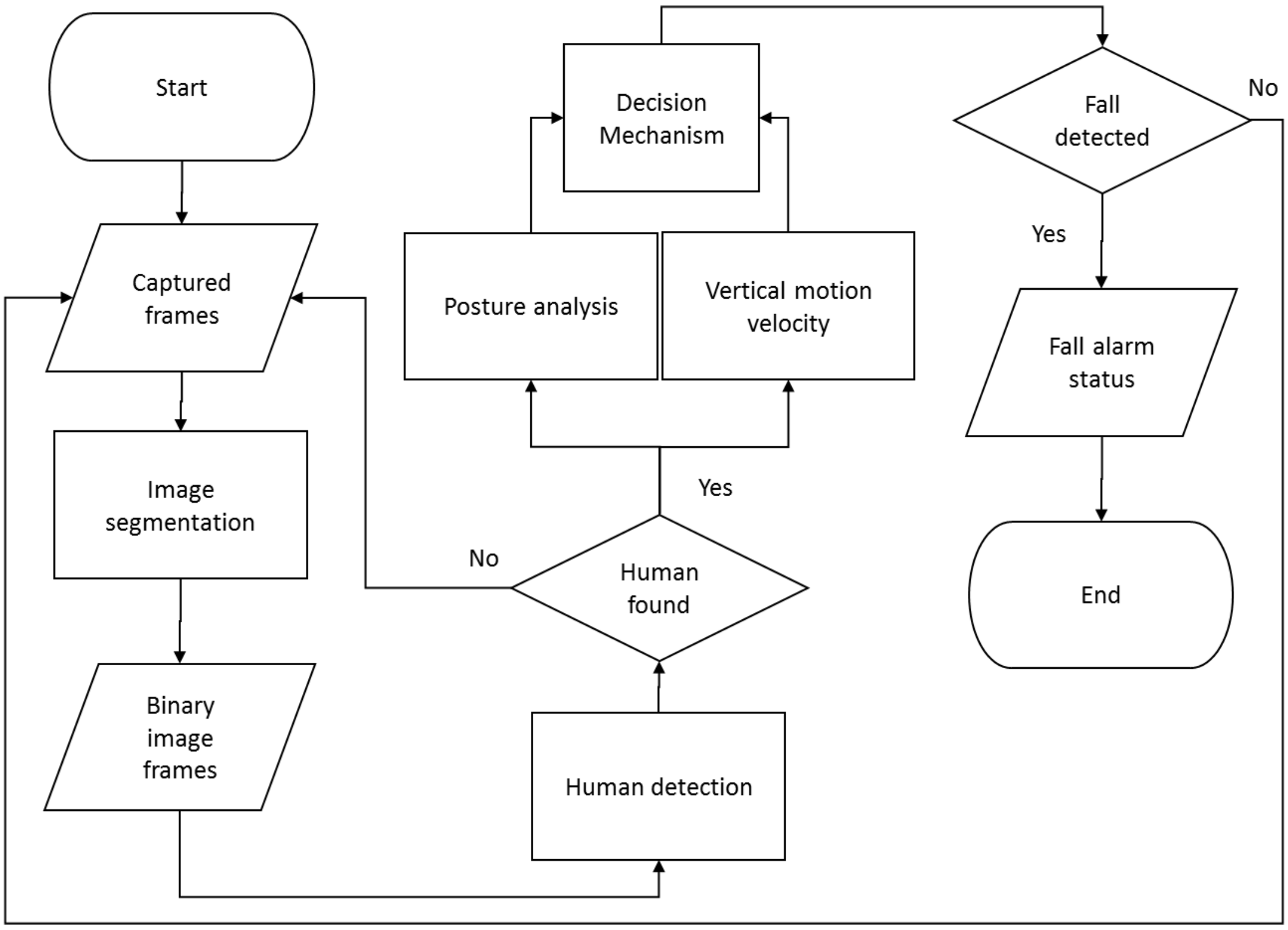

*Figure 7.1. Fall detection proposed methodology flowchart.*

### 7.2.1 The self-training approach

The self-training approach can also be seen as kNN refinement process. The process is, mainly, based on the cluster approach: the data tend to form discrete clusters, and points in the same cluster are more likely to share a label. The overall concept is similar to the work of (Protopapadakis et al., 2012).

Initially, let us capture a sequence of $n$ frames. That set will be the training set $X = \{X_L \cup X_U\}$ where $X_L = \{(x_1, y_1), .., (x_l, y_l)$ is the labeled frames set and $X_U = \{x_{l+1}, \dots, x_n\}$ is the unlabeled frames set. Thus, each frame $i$ is described by a feature vector $x_i$ and its corresponding class (in a vector form), $y_i = [y_{1i}\ y_{2i}]^T$ , if available. In our case, $y_i = [1\ 0]^T$ for non-fall case, or $y_i = [0\ 1]^T$ for the fall case. The classifier is firstly created using the island genetic algorithm (Figure 6.1) over $X_L$, and then further trained using $X_U$ set.

Given an unlabeled instance $x_j$ , $j = l + 1, \dots, n$, FFNN generates an output $\hat{y}_j^{(net)}$. At that time, $x_j$ is also compared to the elements of $X_L$ using a kNN approach, producing a new subset $X_C = \{(x_1^C, y_1^C), \dots, (x_k^C, y_k^C)\}$, $X_C \subset X_L$. Subset $X_C$ contains the $k$ closest instances of $X_L$ to $x_j$. The kNN classifier, then, produces an output $\hat{y}_j^{(knn)}$ according to the following equation:

$$\hat{y}_j^{(knn)} = \begin{cases} [0\ 1]^T, if \max\{\hat{y}_{21}^c, \dots, \hat{y}_{2k}^c\} = 1 \\ [1\ 0]^T\ otherwise \end{cases} \tag{7.1}$$

In other words, even if one of the closest neighbors describe a fall the $j$-th frame, described by $x_j$ , corresponds to a fall incident.

In case that FFNN output, $\hat{y}_j^{(net)}$, and kNN classifier output, $\hat{y}_j^{(knn)}$ , do not agree an expert is summoned to classify the j-th frame in the fall/non-fall category. When the entire set $X_U$ is labeled, the FFNN is retrained using the entire $X$ set.

### 7.2.2 nLWS initialization

Initially, the aforementioned calculated data are separated in two classes, i.e. falls and non-falls, using a combination of various metrics on a k-means algorithm. The variation of metrics includes: (a) squared Euclidean distance, (b) sum of absolute differences and (c) one minus the cosine of the included angle between data points. Then an assumption is made: the fewer element class is considered as the fall class. This assumption is justified by the fact that fall incidents occur far less frequently than other ordinary activities. A FFNN is topologically optimized using as inputs the previously generated data.

Inputs of size 3x1 are used to produce corresponding outputs of size 2x1. Output vector's elements values are associated with the probability an event to be a fall and non-fall. The event is classified according to the higher output element's value. High diference between output element's values suggests a robust performance and solid adaptation to falls and non-falls. A plain weighted sum model is used for the class definition of each observation. Model has the following form:

$$\boldsymbol{y}_c = \sum_{\forall metric} w_i \boldsymbol{y}_i \tag{7.2}$$

where $w_i$ stands for the trade-off value for metric $i$ and $\boldsymbol{y}_i$ is a $2 \times 1$ binary vector whose elements correspond to non-fall and fall classes and their values can be one or zero according to $k$-means clustering, based on metric $i$. Only a small portion of them is used to form the initial training set $\boldsymbol{X}_L$for the nLWS.

The nLWS is then created based on labeled examples, using the island GA. Once the genetic operation is concluded, the fittest FFNN is evaluated and further trained over the entire data set $\boldsymbol{X}_L \cup \boldsymbol{X}_U$. It has to be mentioned that initialization procedure takes place offline.

### 7.2.3 nLWS operation

The operation of nLWS is based on a simple scheme. Inputs are provided for the FFNN, if its output element that corresponds to the fall class presents higher value than the element corresponds to the non-fall class, then a fall incident occurred, or will occur, in the following 0.9 seconds, which, as mentioned before, is the average duration of a fall incident.

### 7.2.4 Possible limitations

In complex environmental conditions, such as the ones encountered in bathrooms, the performance of our system should be further investigated. This is due to the fact that in such environments, severe occlusions are expected (bath curtains, steams) let alone humidity factors that may affect the performance of any visual sensor. We expect that using normal activities, same as within other house rooms (crossing, walking), our system can yield almost the same performance. But for specific bathing conditions special care should be taken into account. One important issue is also the privacy. People are very reluctant to install monitoring devices (especially cameras) in such private spaces.

Please note that other type of sensors (e.g. depth sensors) can potentially improve system performance. However, currently these sensors are applicable for close range cases (few meters), making them not so reliable for monitoring elderly people in their homes, especially for large rooms. In addition, their cost, though very affordable, is few times greater than our low-cost optical devices, implying that our system is much more suitable for wide scale implementations, like health-care premises and elderly nursing homes.

## 7.3 Feature extraction

In this section, we provide a description for both 2D features and 3D features, utilized by the proposed methodology. 2D features include person's projected width-height ratio and person's body orientation, while 3D features include vertical motion velocity based on person's actual height estimation. All features are extracted using a simple monocular camera. Generally, there are no limitations regarding the maximum distance of the camera from the falling scene, neither for the resolution, as long as the human spans an area of more than 40 pixels in each frame, in order to achieve robust segmentation of foreground objects.

### 7.3.1 Image segmentation

The first step is the image segmentation procedure in order to separate foreground (i.e. human) and background (i.e. all else). We have, both, indoor environment and outdoor environment operation requirements. Illumination conditions can dramatically and suddenly change in indoor environments, due to artificial light sources, while in outdoor environments illumination, usually, changes progressively and slow. Furthermore, outdoor environments present more complicated background with higher background motion and more moving objects compared to indoor environments.

We test the following algorithms: a) Iterative Scene Learning algorithm (ISL) presented in (N. Doulamis, 2010), b) Adaptive Student's-t Mixture Model background subtraction (ASMM), presented in (Makantasis et al., 2012) and c) non-Parametric Background Generation (nPBG), presented in (Liu et al., 2007). This choice is justified by the fact that ISL algorithm extracts the foreground by using motion information in the scene, ASMM subtracts the background by using a parametric approach to learn background pixels intensities and nPBG learns the same intensities in a non-parametric way.

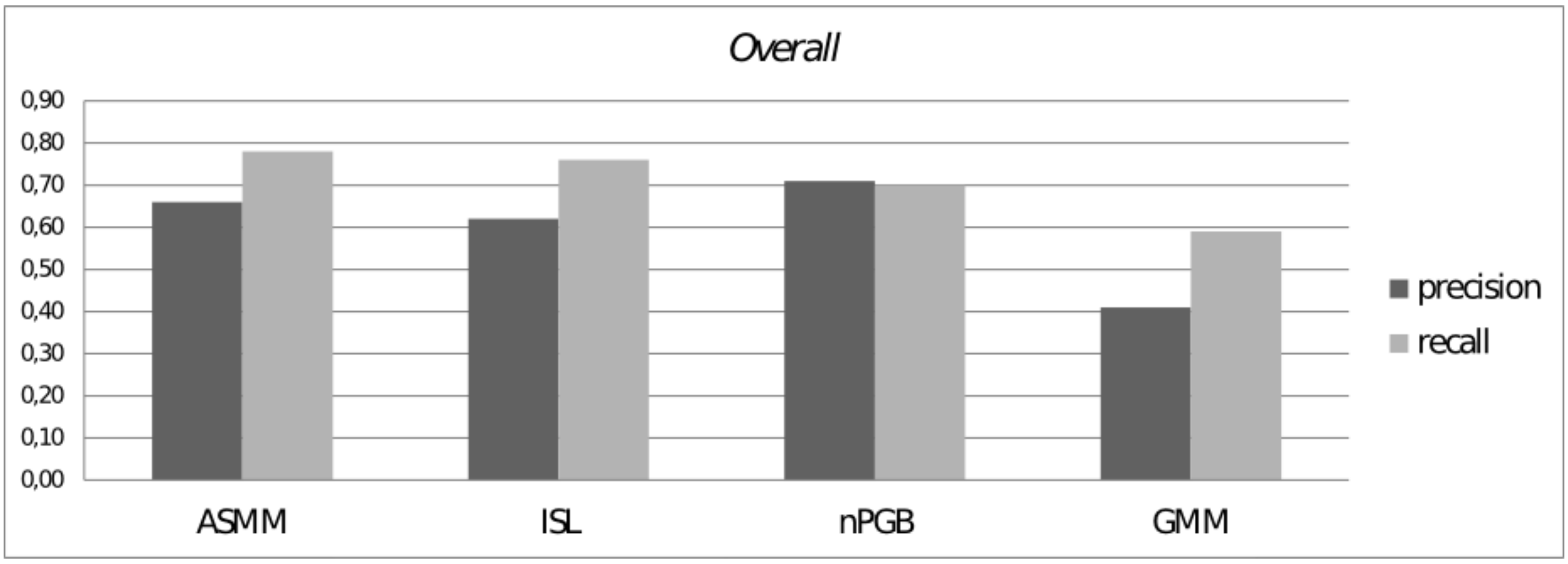

*Figure 7.2. Precision and Recall diagrams for indoor-outdoor environments.*

### 7.3.2 Describing a fall

The features that are used to discriminate fall incidents than other ordinary activities are: vertical motion velocity, based on actual person's height, person's projected width-height ratio and body orientation. All of them are calculated over the foreground pixels. You may find the details of such process in (Makantasis et al., 2015b).

Vertical motion velocity can be defined as the time derivative of human height: $V = \nabla h_{\alpha}(t)$, where $h_{\alpha}(t)$ stands for the actual height of a human in 3D space at a time instance $t$. Denser the change, greater the possibility of a fall. However, height estimation from a single camera requires a self-calibration process. Additionally, In contrast to ordinary human activities, during which human posture changes slowly, during a fall human posture changes suddenly. On the one hand, human posture can be characterized by person's width-height ratio, and this is valid as this ratio is bigger in value when a fall event occurs than the same ratio with the person in standing position, and on the other by the orientation of person's body (Foroughi et al., 2008).

Width-height ratio is determined by person's projected width and height. So, the first step for the computation of this ratio is the estimation of these two measures. Both of these measures can be estimated by the four corners of a minimum bounding box that includes the person. By using the four corners of the minimum bounding box the points $q_{bm}$, $q_{tm}$, $q_{lm}$ and $q_{rm}$, which correspond to foreground object's bottom-most, top-most, left-most and right-most points, can be obtained. By using these four pints, width-height ratio can be expressed as:

$$R = \frac{w_p}{h_p} = \frac{q_{rm} - q_{lm}}{q_{tm} - q_{bm}} \quad (7.3)$$

where $w_p$ and $h_p$ stand for the projected width and height of the foreground object.

Orientation of a person's body can be successfully described by the orientation of an ellipse that best bounds the person. The approximation of such an ellipse requires to define its center $(\bar{x}, \bar{y})$ , its orientation, which is the angle $\phi$ of its major semi-axis and the lengths $a$ and $b$ of its major and minor semi-axes.

As described in (Foroughi et al., 2008), a bounding ellipse can be approximated by image moments. Having estimate the bounding box that contains a foreground object, as shown in (Spiliotis and Mertzios, 1998) the computational cost for computing image moments is linear to the number of foreground object pixels. Thus, the complexity is independent on the size of the image, depending only on the size of the foreground objects.

## 7.4 Experimental results

For the experimentation process, we had to simulate a person’s fall, in every direction according to the camera position and normal every day activities. There are, also, cases that may look like but they are not real falls (Figure 7.3). The fall detection algorithm was tested in dynamically changing visual conditions, including illumination changes, cluttered background and occlusions

In-sample and out-of-sample algorithm's performance is presented in Figure 7.4; results correspond to a video sequence of 1000 frames that contain 19 fall incidents. Since most of the time the actor was walking or standing, we keep few characteristic keyframes of the original footage for visualization purposes. Its performance is affected by the quality of extracted features and subsequently by foreground extraction. For this reason, our system presents more robust performance for indoor environments.

However, it should be mentioned that its performance is not affected by humans' height, because the threshold for discriminating fall incidents than normal activities, is estimated through a learning procedure and, thus, is adapted to individual's height. In addition, the impact of occlusions is being reduced as camera's height is being increased.

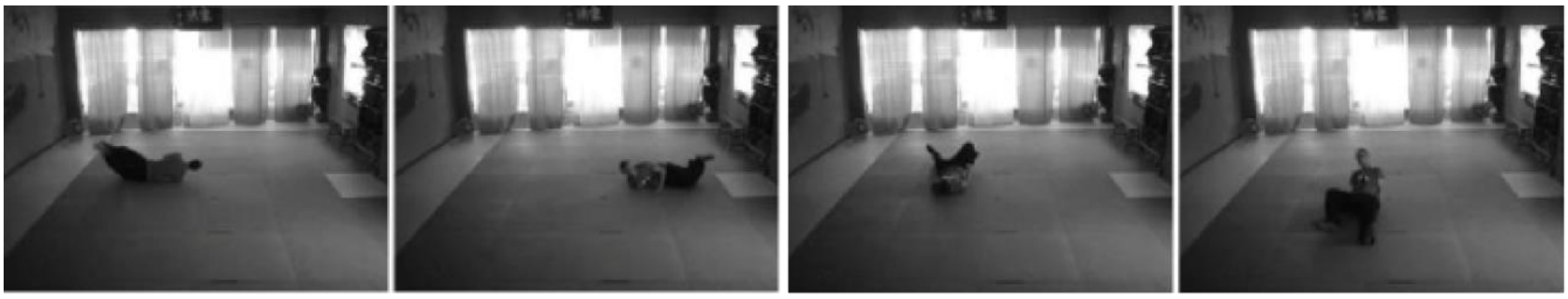

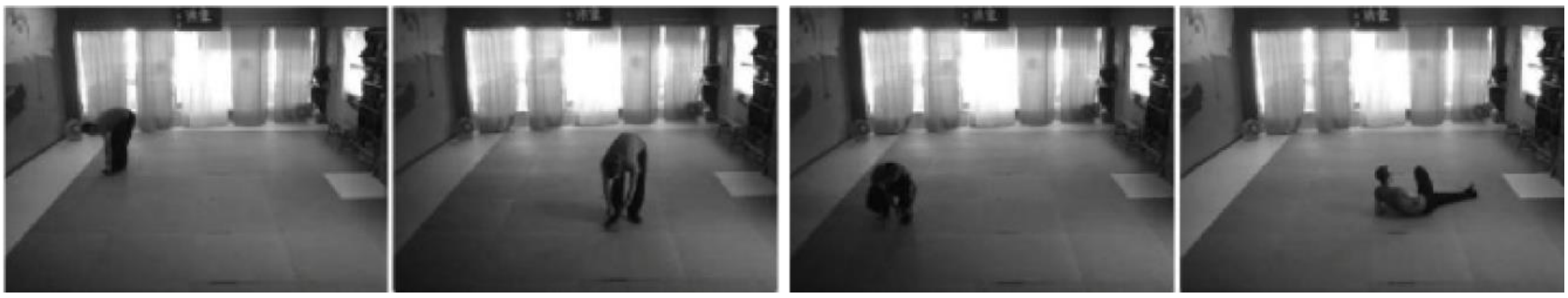

*Figure 7.3. Simulated activities during experimentation process: Falls (above) and normal activities (bellow).*

Finally, in Table 7.2 false positive rates are presented with regard to different activities. The biggest false positive rate is presented when the human lies on the floor, however, this activity cannot be thought as "normal". False positive rates, associated with the "lying on the floor" activity, can decrease by relaxing vertical velocity threshold, used to discriminate fall incidents than normal activities. Relaxing this threshold, however, can increase false negative rates, which are of a primary concern for a fall detection system.

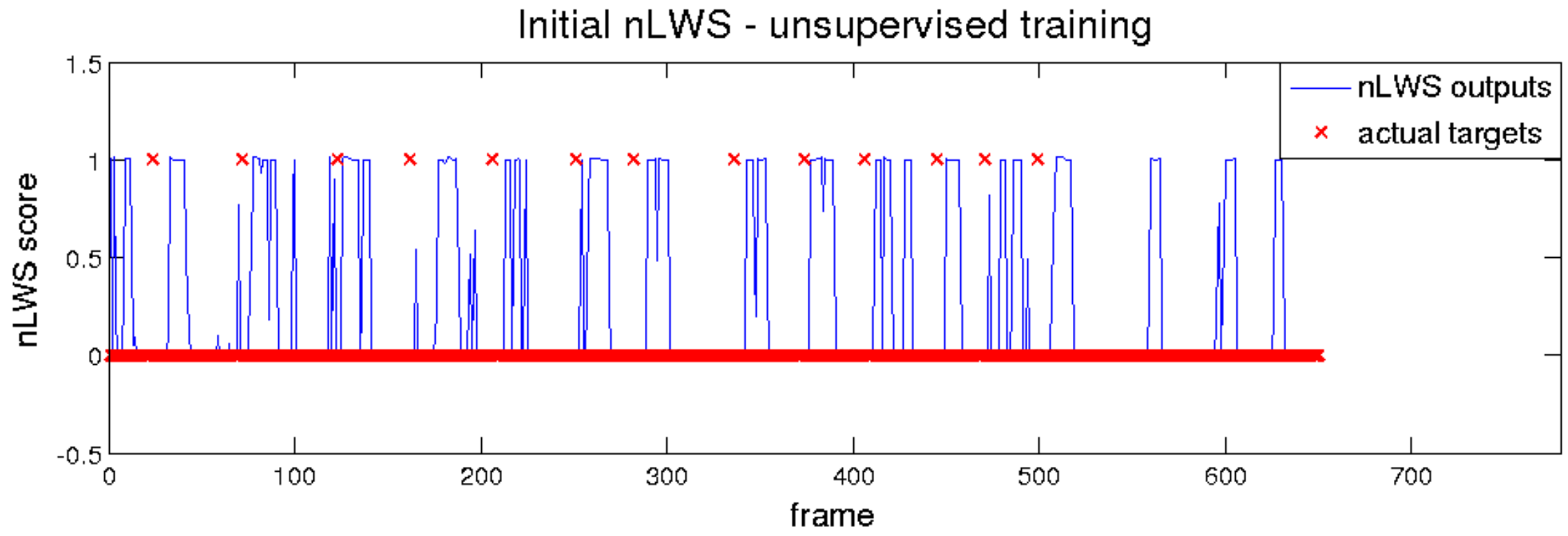

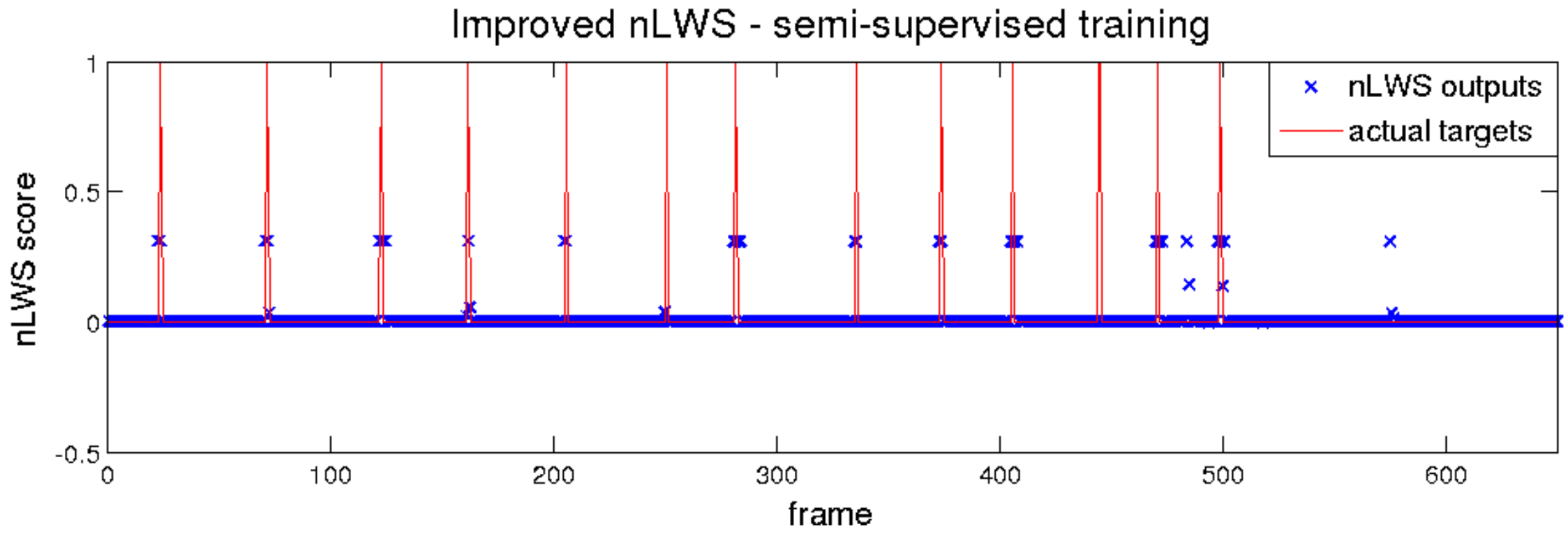

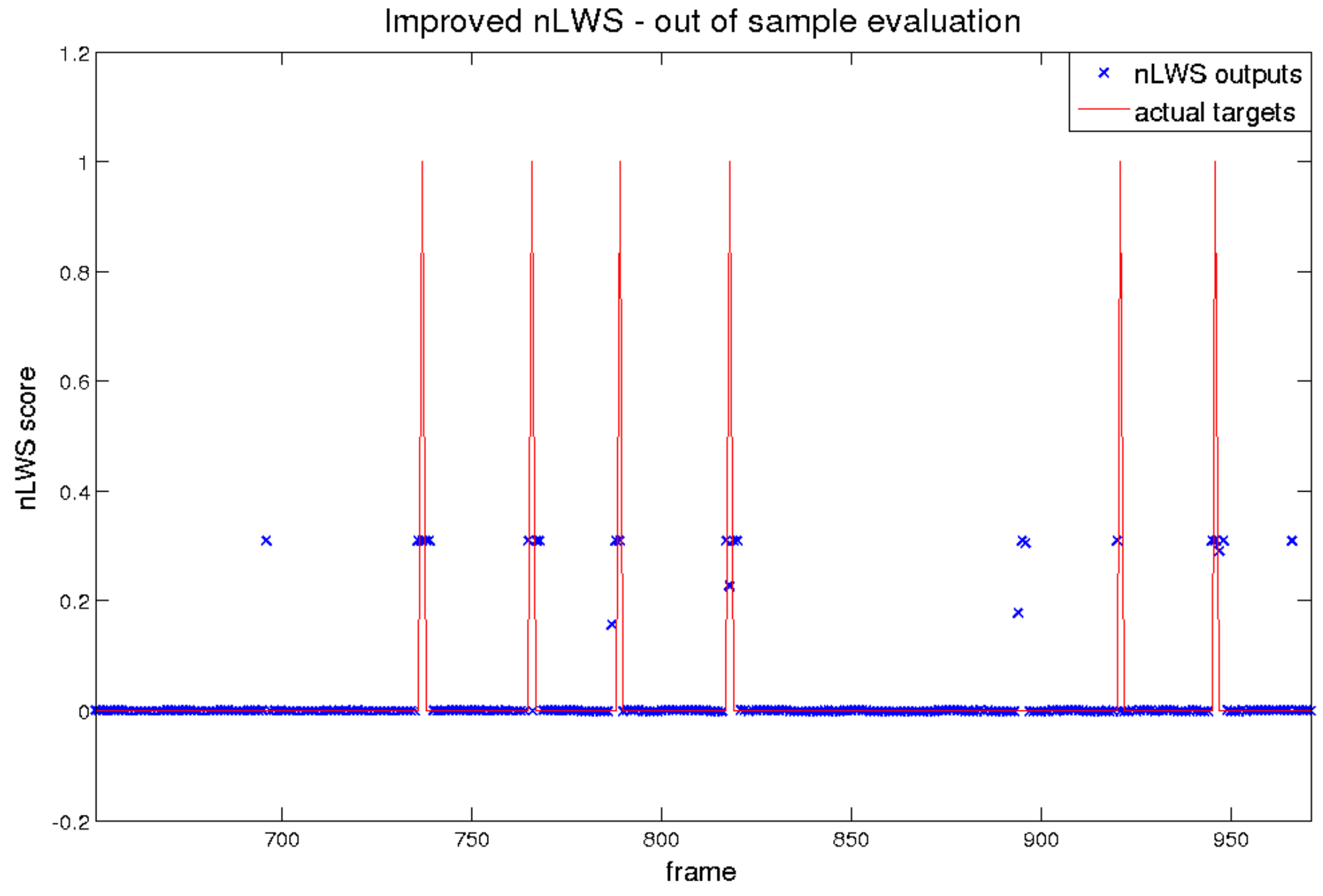

*Figure 7.4. (a) In-sample performance - unsupervised and semi-supervised training results and (b) out-of-sample performance.*

*Table 7.1. Proposed system's overall performance.*

| Camera's Height (cm) | | Proposed system | Indoor No occlusion | Indoor With occlusion | Outdoor |
|---|---|---|---|---|---|
| **40** | Falls detected | 90% | 98% | 76% | 83% |
| | Wrong detections | 4 | 2 | 5 | 9 |
| **220** | Falls detected | 93% | 97% | 92% | 82% |
| | Wrong detections | 6 | 4 | 7 | 8 |
| **260** | Falls detected | 97% | 97% | 94% | 82% |
| | Wrong detections | 3 | 3 | 4 | 6 |

*Table 7.2. Total false positive rate divided in regard to human activities.*

| Activity | Lie down | Sit on the floor | Other |
|---|---|---|---|
| **False Positive** | 62.5% | 25% | 12.5% |

## 7.5 Conclusions & future work

This chapter presents a fall detection scheme that uses a single low-cost monocular camera. Through camera self-calibration and perspective transformations, our system is capable to exploit 3D measures to increase its robustness. It operates in real-time and is capable to detect over 90% of fall incidents in complex and dynamically changing visual conditions, while it presents very low false positive rate. Its low computational cost and memory requirements making it suitable for large scale implementations, let alone its low financial cost since simple low resolution cameras are used, making it affordable for a large scale implementation.

# *Chapter* VIII: Sea Border Surveillance

***Intelligence is the ability to adapt to change.***
***Stephen Hawking, English theoretical physicist***

## 8 Maritime surveillance

This chapter presents a vision-based system for maritime surveillance, using moving PTZ cameras. The proposed methodology fuses a visual attention method that exploits low-level image features appropriately selected for maritime environment, with appropriate tracker. Such features require no assumptions about environmental nor visual conditions. The offline initialization is based on large graph semi-supervised technique in order to minimize user's effort.

In particular, the scalable SSL approach actuates over the initially annotated training data set; it is, therefore, utilized only once, at the initialization offline phase. The main goal is the minimization of possible annotation errors, occurred during the training dataset creation stage.

System's performance was evaluated with videos from cameras placed at Limassol port and Venetian port of Chania. Results suggest high detection ability, despite dynamically changing visual conditions and different kinds of vessels, all in real time. Analytical description for the system implementation can be found in (Makantasis et al., 2015c).

### 8.1 Introduction

Management of emergency situations, known to the maritime domain, can be supported by advanced surveillance systems suitable for complex environments. Such systems vary from radar-based to video based. The former, however, has two major drawbacks (Zemmari et al., 2013); it is quite expensive and its performance is affected by various factors (e.g. echoes from targets out of interest). The latter, consists of various techniques, each one with specific advantages and drawbacks. The majority of such systems are controlled by humans, who are responsible for monitoring and evaluating numerous video feeds simultaneously.

Advanced surveillance systems should process and present collected sensor data, in an intelligent and meaningful way, to give a sufficient information support to human decision makers (Fischer and Bauer, 2010). The detection and tracking of vessels is inherently depended on dynamically varying visual conditions (e.g. varying lighting and reflections of sea). So, to successfully design a vision-based surveillance system, we have to carefully define both its operation requirements and vessels' characteristics.

On the one hand there are minimum standards concerning operation requirements (Szpak and Tapamo, 2011). At first, it must determine possible targets within a scene containing a complex, moving background. Additionally, the system must not produce false negatives and keep as low as possible the number of false positives. Since we are talking about surveillance system, it must be fast and highly efficient, operating at a reasonable frame rate and for long time periods using a minimal number of scene related assumptions.

On the other hand, regardless of vessel type's variation, there are four major descriptive categories. First comes the size, which ranges from jet-skis to large cruise ships. Secondly, we have the moving speed. Thirdly, vessels move to any direction, according to the camera position, and thus their angle varies from $0^o$ to $360^o$. Finally, there is vehicles' visibility. Some vessels have a good contrast to the sea water while others are intentionally camouflaged. A robust maritime surveillance system must be able to detect vessels having any of the above properties.

### 8.1.1 Related work

This chapter focuses on detection and tracking of targets within camera's range, rather than their trajectory patterns' investigation (Lei, 2013; Vandecasteele et al., 2013) or their classification in categories of interest (Maresca et al., 2010). The system's main purpose is to support end-user in monitoring coastlines, regardless of existing conditions.

Object detection is a common approach with many variations; i.e. an-isotropic diffusion (Voles, 1999), which has high computational cost and per forms well only for horizontal and vertical edges, foreground object detection/image color segmentation fusion (Socek et al., 2005). In (Albrecht et al., 2011a; Albrecht et al., 2010) a maritime surveillance system mainly focuses on finding regions in images, where is a high likelihood of a vessel being present, is proposed. Such system was expanded by adding a sea/sky classification approach using HOG (Albrecht et al., 2011b). Vessel classes' detection, using a trained set of MACH filters was proposed by (Rodriguez Sullivan and Shah, 2008).

All of the above approaches adopt offline learning methods that are sensitive to accumulation errors and difficult to generalize for various operational conditions. (Wijnhoven et al., 2010) utilized an online trained classifier, based on HOG. However, retraining takes place when a human user manually annotates the new training set. In (Szpak and Tapamo, 2011) an adaptive background subtraction technique is proposed for vessels extraction. Unfortunately, when a target is almost homogeneous is difficult, for the background model, to learn such environmental changes without misclassifying the target.

More recent approaches, using monocular video data, are the works (Makantasis et al., 2013) and (Kaimakis and Tsapatsoulis, 2013). The former, utilizes a fusion of Visual Attention Map (VAM) and background subtraction algorithm, based on Mixture of Gaussians (MOG), to produce a refined VAM. These features are fed to a neural network tracker, which is capable of online adaptation. The latter, utilized statistical modelling of the scene's non-stationary background to detect targets implicitly.

The work of (Auslander et al., 2011) emphasize on algorithms that automatically learn anomaly detection models for maritime vessels, where the tracks are derived from ground-based optical video, and no domain-specific knowledge is employed. Some models can be created manually, by eliciting anomaly models in the form of rules from experts (Nilsson et al., 2008), but this may be impractical if experts are not available, cannot easily provide these models, or the elicitation cost may be high.

### 8.1.2 Place for improvement

A careful examination of the proposed methodologies suggest that specific points have to be addressed. Firstly, a system needs to combine both supervised and unsupervised tracking techniques, in order to exploit all the possible advantages. Secondly, since we deal with vast amount of available data, we need to reduce, as much as possible, the required effort for the initialization of the system. The innovation of our approach lies in the creation of a visual detection system, able to overcome the aforementioned difficulties by combining various, well tested techniques and, at the same time, minimizes effort during the offline initialization using a Semi- Supervised Learning (SSL) technique, appropriate for large data sets.

In contrast to the approach of (Makantasis et al., 2013), the user has to roughly segment few images, i.e. use minimal effort, in order to create an initial training set. Such procedure is easily implemented using the suggested areas according to the unsupervised techniques' results. Collaboration of visual attention maps, that represents the probability of a vessel being present in the scene, and background subtraction algorithms provides to the user initially segmented parts, over which user further actuates.

Then, SVMs are used as the additional supervised technique, in order to handle new video frames. The significant amount of labelled data for the training process originates from the previously generated roughly segmented data sets. In order to facilitate the creation of such training set and further refine it (i.e. correct some user errors), SSL graph-based algorithms need to be involved. Unfortunately, SSL techniques scale badly as the available data rises. Consequently, a semi-supervised procedure, suitable for large data sets is exploited for the offline initialization, significantly reducing the effort required.

## 8.2 Proposed methodology

The main goal is the real-time detection and tracking of maritime targets. Towards this direction, an appearance-based approach is adopted to create visual attention maps that represent the probability of a target being present in the scene. High probability implies high confidence for a maritime target's presence.

Visual attention maps creation is based exclusively on each frame's visual content, in relation to their surrounding regions or the entire image. Consequently, they do not take into consideration neither the temporal relationship between subsequent frames, nor any motion information presented in the scene. Due to this limitation, high probability is assigned, frequently, to image regions that depict non-maritime targets (e.g. stationary land parts). In order to overcome such drawback, our system exploits the temporal relationship between subsequent frames.

Concretely, video blocks, containing a predefined number, $h$, of frames and covering a time span, $T$, are used to model the pixels' intensities. Thus, the temporal evolution of pixels intensities is utilized to estimate a pixel-wise background model, capable to denote each one of the pixels of the scene as background or foreground. By using a background modelling algorithm, system can efficiently discriminate moving from stationary objects in the scene. In order to model pixels' intensities, we use the background modelling algorithm presented in (Zivkovic, 2004). This choice is justified by the fact that this algorithm can automatically fully adapt to dynamically changing visual conditions and cluttered background.

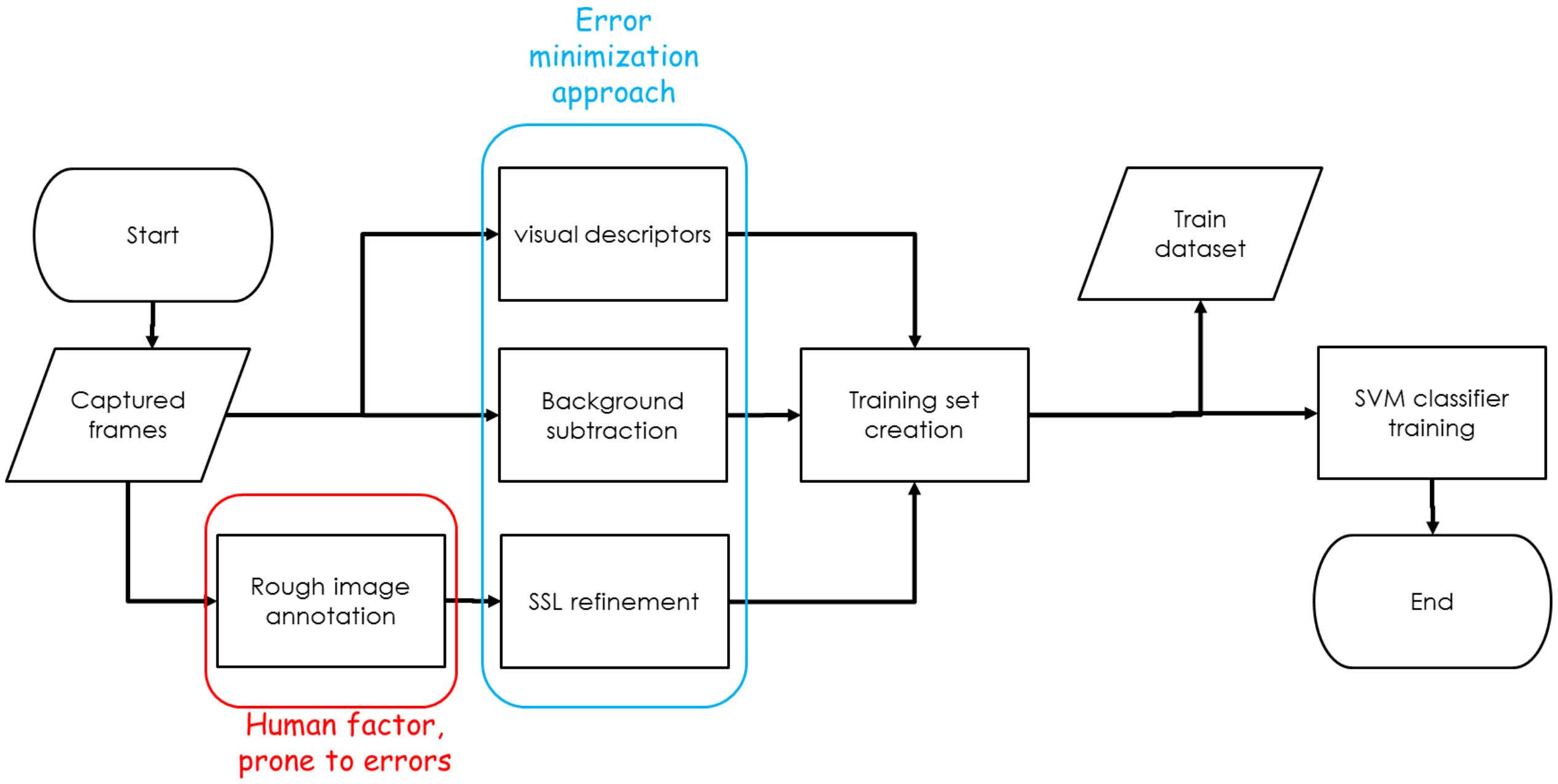

*Figure 8.1. The offline initialization process. The SSL approach tries to minimize annotation errors due to human factor.*

### 8.2.1 Problem formulation

Maritime target detection can be seen as an image classification problem. Thus, we classify each one of the frame's pixels in one of two classes, $C_T$ and $C_B$. If we denote as $l_{xy}^{(i)}$ the label of pixel $p_{xy}^{(i)}$, then, for a frame $i$, the classification task can be formulated as:

$$l_{xy}^{(i)} = \begin{cases} 1 \; if \; p_{xy}^{(i)} \in C_T \; for \; x = 1, \dots, w \; and \; y = 1, \dots, h \\ -1 \; if \; p_{xy}^{(i)} \in C_B \; for \; x = 1, \dots, w \; and \; y = 1, \dots, h \end{cases} \tag{8.1}$$

where $h$ and $w$ stand for frame's height and width.

Although, a binary classifier, SVM in our case, can successfully transact the classification task, a classifier training process should precede. Training process requires the formation of a robust training set composed of appropriate pixel descriptors, along with their associated labels. Such a set can be formed by the user, through a rough segmentation of a frame $t$ into two regions, that contain positive and negative samples, i.e. pixels that belong to $C_T^{(t)}$ class, labelled as 1, and pixels that belong to $C_B^{(t)}$ class, labelled as -1. The union of $C_T^{(t)}$and $C_B^{(t)}$ consists the initial training set $S$.

### 8.2.2 Possible limitations

Low level features extraction consists the main computational bottleneck of our system. However, the proposed approach can be expanded for video frames of greater resolution, since each feature can be extracted independently. Thus, feature extraction process can be easily parallelized using multiple processing units (e.g. multiple CPU threads or GPU implementation).

The vision approach is able to detect targets in a similar way to the human eye. As long as the camera is able to capture a vessel (i.e. spans an area of more than 40 pixels in the frame) the system will likely detect it, regardless the weather conditions (e.g. rain, fog, waves etc.). Apparently, system's performance declines badly in cases of low luminosity due to sensor related sensitivity constraints. Better sensors can partially deal with such issues, but resulting in greater hardware costs.

Finally, the SSL is used for refinement of the initial training test; it is not an image annotation mechanism. Therefore, if initial annotations are poorly made, the refinement process will not work.

## 8.3 Feature extraction

The feature extraction procedure needs first to address the scale variance issue. Potential targets in maritime environment vary in sizes, either due to their physical size or due to the distance between them and the camera. Despite that, most of the feature detectors operate as kernel based method and thus they prefer objects of a certain size. As presented in (Alexe et al., 2010; T. Liu et al., 2011) images must be represented in different scales in order to overcome this limitation. In our approach, a Gaussian image pyramid is exploited in order to provide scale invariance and to take into consideration the relationship between adjacent pixels.

Then, we have the traditional Low-level features analysis. As described in (Albrecht et al., 2011a, 2011b) different low-level image features respond to different attributes of potential maritime targets. Thus, a combination of features should be exploited in order to reveal targets' presence. The selected low-level features do not require a specific format for the input image. These are: edges, horizontal and vertical lines, frequency, color and entropy.

The density of image edges can successfully describe the overall structure of an image, horizontal and vertical lines are able to denote man-made structures, making the system able to suppress large image regions, depicting sea and sky. Frequency can successfully emphasize objects in noisy conditions, such as vessels in a wavy sea. Color feature can successfully emphasize objects colored different than sea and sky and, finally, entropy quantifies the amount of information coded in an image.

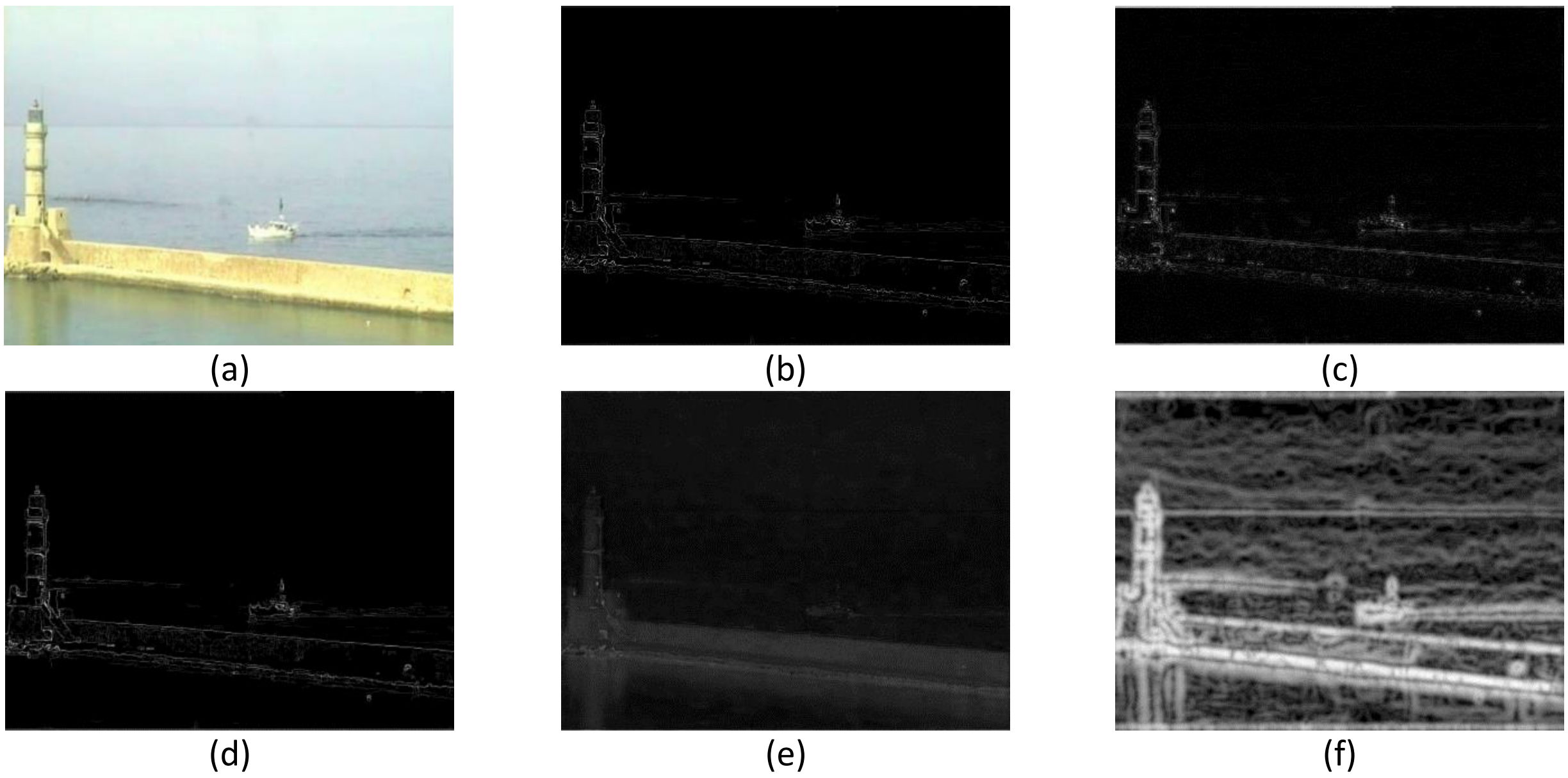

*Figure 8.2. Original captured frame (a) and feature responses (b)-(f); (b) edges, (c) frequencies, (d) vertical and horizontal lines, (e) color and (f) entropy. All feature responded to the land part and the boat (maritime target).*

Utilization of the exported low-level features leads to the creation of visual descriptors. Visual descriptors are computed to encode visual information of captured images. These descriptors are utilized for constructing the visual attention maps. Their computation, instead of pixel-wise, takes place block-wise, in order to reduce the effect of noisy pixels during low-level features extraction. Three different descriptors are computed:

a. Local descriptors that take into consideration each one of the image pixels separately. Local descriptors indicate the magnitude of local features for each one of image pixels.
b. Global descriptors that are capable to emphasize pixels with high uniqueness compared to the rest of the image. To achieve this they indicate how different local features for a specific pixel are, in relation with the same features of all other image pixels.
c. Window descriptors that compare local features of a pixel with the same features of its neighboring pixels.

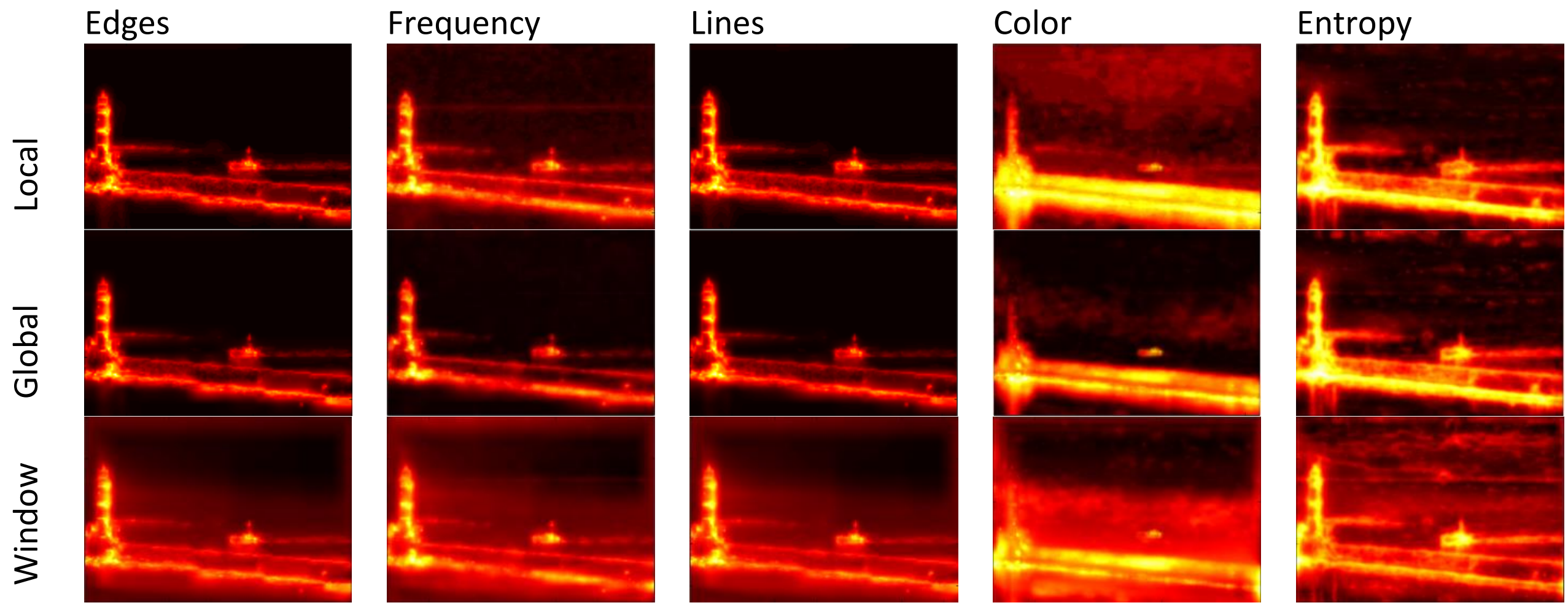

*Figure 8.3. Visual attention maps for each local, global and window descriptors. Using give low level features and three descriptor, each one of the frames pixels is described by a 15-dimensional vector. The presented visual attention maps correspond to the original frame of Figure 8.2.*

### 8.3.1 Local descriptors

One local descriptor is computed for each one of the extracted low-level features. Let us denote as $F$ the feature in question, which can correspond to image edges, frequency, horizontal and vertical lines, color or entropy. For the feature $F$, the computation of local descriptor is derived by feature's response image. As mentioned before, descriptors are computed block-wise. So, firstly the feature's response image is divided into $B$ blocks of size $8 \times 8$ pixels. Then, the local descriptor for a specific block $j$ is defined as the average magnitude of the feature $F$ in the same block. More formally, for a block $j$, with $b_h$ height and $b_w$ width, the local descriptor of feature $F$ is computed as follows:

$$lF_j = \frac{1}{b_h \cdot b_w} \sum_{(x,y)\in j} F(x,y) \tag{8.2}$$

where $F(x,y)$ is the response of feature in question at pixel $(x,y)$. This kind of descriptor is capable to highlight image bocks with high feature responses.

### 8.3.2 Global descriptors

The local descriptors handle each image block separately and, thus, are insufficient to provide useful information when features' responses are quite similar along all image blocks. Consider, for example, an image full of edges, like a wavy sea. In this case, local descriptor $l\mathcal{E}$, which is associated with the image edges, is not able to provide useful information about salient objects in the scene, since all blocks will present high edge responses. The proposed system can overcome this problem by using global descriptors.

Uniqueness of a block $j$ can be evaluated by the absolute difference of the feature response between this block and the rest blocks of the image. The global descriptor for a feature $F$ and image block $j$ is defined as:

$$gF_j = \frac{1}{B} \sum_{i}^{B} \left| lF_j - lF_i \right| \tag{8.3}$$

### 8.3.3 Window descriptors

Local and global descriptors are capable to emphasize image blocks that are highly distinctive, in terms of features' responses, or have a unique presence in the image. However, if potential targets are presented in more than one block the aforementioned descriptors will emphasize the most dominant target and will suppress the others. In order to overcome this problem, system exploits a window descriptor, which compares each image block with its neighboring blocks.

Window descriptor for an image with $N \times M$ blocks uses an image window $W$, which is spanned by the maximum symmetric distance, $d_h$ and $d_v$ along horizontal and vertical axes respectively. Symmetric distances are defined as $d_h = \min(l, k_h, N - k_h)$ and $d_v = \min(l, k_v, M - k_v)$, where $l$ is the default symmetric distance, 3 blocks in our case, and $k_h$ and $k_v$ stands for block coordinates on image plane along horizontal and vertical axes respectively. The window descriptor for a feature $F$ and image block $j$ with coordinates $(j_1, j_2)$ is defined as:

$$wF_j = \frac{1}{2d_h \cdot 2d_v} \sum_{k=-d_h}^{d_h} \sum_{l=-d_v}^{d_v} \left| lF_j - lF_{j_1+k, j_2+l} \right| \tag{8.4}$$

### 8.3.4 Background subtraction

For the maritime surveillance case, most state-of-the-art background modeling algorithms, like (Doulamis and Doulamis, 2012; Makantasis et al., 2012), fail either due to their high computational cost or due to the continuously moving background, and moving cameras. However, if the background modeling algorithm output is fused in a unified feature vector with the previously constructed visual attention maps, our system will be able to emphasize potential threats and at the same time to suppress land parts that may be appeared in the scene by implicitly capture motion presence.

The proposed system uses the Mixtures of Gaussians (MOG) background modeling technique, presented in (Zivkovic, 2004). This choice is justified by the fact that MOG is fast, robust to small periodic movements of background, and easy to parameterize algorithm. An illustration of MOG technique is shown in Figure 8.4.

By fusing together the outputs of visual attention maps and the output of a background modeling algorithm, camera motion temporarily increases false positives detections, but false negatives, that comprises the most important characteristic of a maritime surveillance system, are not affected.

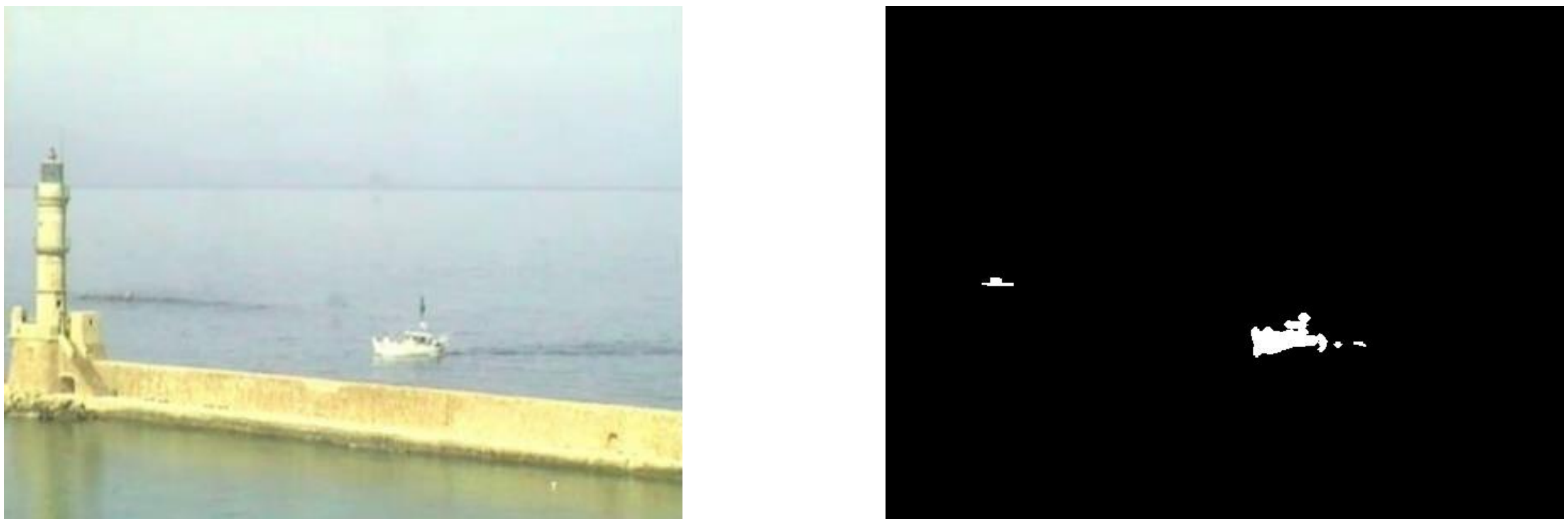

(a) (b)

*Figure 8.4. Original frame (a) and the output of background modeling algorithm (b).*

## 8.4 Training set creation, classifier initialization and adaptation

In order to be able to exploit a binary classifier, a process of classifier training should be preceded. Training process requires the formation of a robust training set which contains pixels along with their associated labels. Let us denote as $Z^{(t)}$ the set that contains all the pixels of frame $t$, $C_T^{(t)}$ the set that contains pixels that depict some part of a maritime target and as $C_B^{(t)}$ the set that corresponds to background.

The creation of a training set $S$ requires from the user to roughly segment the frame $t$ into two regions, which contain positive and negative samples (i.e. pixels that belong to $C_T^{(t)}$ and $C_B^{(t)}$ class respectively). This segmentation results in a set $S = \left\{\left(p_{xy}^{(t)}, l_{xy}^{(t)}\right)\right\}$, and labels are defined as:

$$l_{xy} = \begin{cases} 1 \; iff \; p_{xy} \in C_T \\ -1 \quad else \end{cases} \tag{8.5}$$

where $p_{xy}$ is a pixel at location $(x, y)$.

The next step is the accurate description of any pixel $p_{xy}$. Based on image low level features (described in sec. 8.3), we create visual attention maps that indicate the probability a pixel to depict a part of a maritime target. In addition, based on the observation that a vessel must be depicted as a moving object, we implicitly capture the presence of motion by exploiting a background modeling algorithm. Using the output of visual attention maps and the background modeling algorithm we can form appropriate feature vectors.

By using three descriptors and five low-level image features, each image block is described by a 1×15 feature vector. Each feature of this vector corresponds to a different visual attention map. For blocks of size 8×8 pixels the visual attention maps are sixty four times smaller than the original captured frame. In order to create a pixel-wise feature vector, we rescale visual attention maps to the same dimensions with the original captured frame. Thus, each pixel, $p_{xy}^{(i)}$, of a frame $i$, at location $(x, y)$ on image plane, is described by an appropriate feature vector $\boldsymbol{f}_{xy}^{(i)}$:

$$\boldsymbol{f}_{xy}^{(i)} = \left[f_{1,xy}^{(i)}, \dots, f_{k,xy}^{(i)}\right]^T \tag{8.6}$$

where $f_{1,xy}^{(i)}, \dots, f_{k-1,xy}^{(i)}$ stand for scalar features that correspond to the probabilities assigned to the pixel $p_{xy}^{(i)}$ by different visual attention maps, while $f_{k,xy}^{(i)}$ is the binary output of background modeling algorithm, associated with the same pixel. Thus, the training set $S$ can be transformed to:

$$S = \left\{\left(\boldsymbol{f}_{xy}^{(t)}, l_{xy}^{(t)}\right)\right\} \tag{8.7}$$

However, human centric labeling, especially of large image data sets, is an arduous and inconsistent task, mainly due to the complexity of the visual content and the huge manual effort required. To overcome this drawback, we refine the initial training set through a semi-supervised approach. SSL allow the rough initial segmentation facilitating the expert and reducing the time required.

In order to refine the initial user-defined training set, we partition the set $S$ into two disjoint classes, $R$ and $U$. The class $R$ contains the most representative samples of $S$, i.e. the samples that can best describe the classes $C_T$ and $C_B$, while class $U$ contains the rest of the data. Samples of class $R$ are considered as labeled, while samples belonging to $U$ are considered as unlabeled. Then, via a semi-supervised approach the samples of $R$ are used for label propagation through the ambiguously labeled data of $U$. In the following we describe in detail the labeling process.

### 8.4.1 Representative data selection through simplex volume expansion

In order to select the most representative samples for each one of the classes$C_T$ and $C_B$, we consider each sample as a point into an $\mu$-dimensional space. In our case $\mu$ is equal to 16, because the dimension of the space is equal to the dimension of the feature vectors that describe the pixels. The process for representatives selection is conducted twice, once for class $C_T$ and once for $C_B$. In the following, we describe the process for representative samples selection for one of the classes, let's say $C_T$. Exactly the same process is followed for selecting representatives for the other class.

We assume that the $\mu$ -dimensional volume formed by a simplex with vertices specified by the most representative points (pixels), belonging to class $C_T$, should be larger than that formed by any other combination of points of the same class. Let us denote as $\boldsymbol{v}^{(i)}$ the $i^{\text{th}}$ representative sample, as $\beta$ the number of representatives to be generated, as $C_{T,R} = \{\boldsymbol{v}^{(i)}, \dots, \boldsymbol{v}^{(\beta)}\} \subseteq \mathrm{C_T}$ the set that contains the representative samples and as vector $\boldsymbol{w}^{(j)} = \boldsymbol{v}^{(j)} - \boldsymbol{v}^{(i)}$ for $j = 1, \dots, \beta$. Then the volume, $V(C_{T,R})$, of the simplex whose vertices are the points $\boldsymbol{v}^{(i)}$ for $i = 1, \dots, \beta$ can be computed as:

$$V(C_{T,R}) = \frac{|\det(\boldsymbol{W}\boldsymbol{W}^T)|^{1/2}}{(\beta - 1)!} \tag{8.8}$$

where $\boldsymbol{W}$ is an $(\beta - 1) \times \mu$ matrix whose rows are the row vectors $\boldsymbol{w}^{(j)}$.

The estimation process involves several steps. Initially the set $C_{T,R}$ is constructed by randomly selecting $\beta$ samples from set $C_T$ and the volume of the simplex, formed by the elements of $C_{T,R}$, is calculated. Then, an iterative approach is adopted to test every sample in the set $C_T$ as a candidate representative. Each one of the samples of $C_{T,R}$ is replaced, one at a time, with a sample $\hat{\boldsymbol{v}}$ from $C_T$ that is being tested as candidate representative. Then, the algorithm evaluates if replacing any of the elements, of $C_{T,R}$ with the sample being tested, results in a larger simplex volume. If this is true, let's say for the point $\boldsymbol{v}^{(i)} \in C_{T,R}$, then the $\boldsymbol{v}^{(j)}$ point is replaced by the image point $\hat{\boldsymbol{v}}$ and the process is repeated again until each one of the samples of $C_T$ set is evaluated.

The selection method is *scalable* to large datasets, using an incremental approach. Let us assume that β representatives are known. Then, the problem of selecting $\beta + 1$ representatives can be reduced to finding $\beta + 1$ representatives given $\beta$ of them. This way, only the volumes of simplices formed by the elements of the sets $C_{T,R} \cup \boldsymbol{x}^{(i)}$ for $\boldsymbol{x}^{(i)} \in C_T$ need to be evaluated.

### 8.4.2 Graph-based semi-supervised label propagation

The aforementioned procedure results to two sets of representative samples, $C_{T,R}$ and $C_{B,R}$, one for each class. The samples of $C_{T,R}$ and $C_{B,R}$ are considered as labeled, while the rest samples of the classes $C_T$ and $C_B$ are considered as ambiguously labeled. More formally, we have:

$$\begin{gathered} R = C_{T,R} \cup C_{B,R} \\ U = S - C_{T,R} - C_{B,R} \end{gathered} \tag{8.9}$$

At this point, we need to refine the initial training set, $S$, using a suitable approach for the label propagation, through the ambiguously labeled data.

Thus, we need to estimate a labeling prediction function $g\colon \mathbb{R}^{\mu} \rightarrow \mathbb{R}$ defined on the samples of $S$, by using the labeled data $R$. Let us denote as $\boldsymbol{r}_i$ the samples of set $R$ so that $R = \{\boldsymbol{r}_i\}_{i=1}^{m}$, where $m$ is the cardinality of the set $R$. Then, according to (Liu et al., 2010), the label prediction function can be expressed as a convex combination of the labels of a subset of representative samples:

$$g(\boldsymbol{f}_i) = \sum_{k=1}^{m} Z_{ik} \cdot g(\boldsymbol{l}_k) \tag{8.10}$$

where $Z_{ik}$ denotes sample-adaptive weights, which must satisfy the constraints $\sum_{k=1}^{m} Z_{ik} = 1$ and $Z_{ik} \geq 0$ (convex combination constraints). By defining vectors $\boldsymbol{g}$ and $\boldsymbol{a}$ respectively as $\boldsymbol{g} = [g(\boldsymbol{f}_1), \dots, g(\boldsymbol{f}_n)]^{\boldsymbol{T}}$ and $\boldsymbol{a} = [g(\boldsymbol{r}_1), \dots, g(\boldsymbol{r}_m)]^{\boldsymbol{T}}$ eq. (8.10) can be rewritten as $\boldsymbol{g} = \boldsymbol{Z\alpha}$ where $\boldsymbol{Z} \in \mathbb{R}^{n \times m}$.

The designing of matrix $\boldsymbol{Z}$, which measures the underlying relationship between the samples of $U$ and representative samples $R$ (were $R \subset U$), is based on weights optimization; actually non-parametric regression is being performed by means of data reconstruction with representative samples. Thus, the reconstruction for any data point $\boldsymbol{f}_i$, $i = 1, \dots, n$ is a convex combination of its closest representative samples. In order to optimize these coefficients the following quadratic programming problem needs to be solved:

$$\begin{gathered} \min_{\boldsymbol{z}_i \in \mathbb{R}^S} h(z_i) = \frac{1}{2} \|\boldsymbol{f}_i - \boldsymbol{R}_S \cdot z_i\|^2 \\ s.t.\, \boldsymbol{1}^T z_i = 1, z_i \geq 0 \end{gathered} \tag{8.11}$$

where, $\boldsymbol{R}_S \in \mathbb{R}^{\mu \times s}$ is a matrix containing as elements a subset of $R = \{\boldsymbol{r}_1, \dots, \boldsymbol{r}_m\}$ composed of $s < m$ nearest representative samples of $\boldsymbol{f}_i$ and $z_i$ stands for the $i^{\text{th}}$ row of $\boldsymbol{Z}$ matrix.

Nevertheless, the creation of matrix $\boldsymbol{Z}$ is not sufficient for labeling the entire data set, as it does not assure a smooth function $\boldsymbol{g}$. As mentioned before, a large portion of data are considered ambiguously labeled. Despite the small labeled set, there is always the possibility of inconsistencies in segmentation; in specific frames the user may miss some pixels that depict targets. In order to deal with such cases the following SSL framework is employed:

$$\min_{\boldsymbol{A}=[\boldsymbol{a}_1,\dots,\boldsymbol{a}_c]} Q(\boldsymbol{A}) = \frac{1}{2}\|\boldsymbol{Z}\cdot\boldsymbol{A}-\boldsymbol{Y}\|_F^2 + \frac{\gamma}{2} trace\left(\boldsymbol{A}^{\boldsymbol{T}}\hat{\boldsymbol{L}}\boldsymbol{A}\right) \tag{8.12}$$

where $\hat{\boldsymbol{L}} = \boldsymbol{Z}^T \cdot \boldsymbol{L} \cdot \boldsymbol{Z}$ is an memory-wise and computationally tractable alternative of the Laplacian matrix $\boldsymbol{L}$. The matrix $\boldsymbol{A} = [\boldsymbol{a}_1, \dots, \boldsymbol{a}_c] \in \mathbb{R}^{m\times c}$ is the soft label matrix for the representative samples, in which each column vector accounts for a class. The matrix $\boldsymbol{Y} = [\boldsymbol{y}_1, \dots, \boldsymbol{y}_c] \in \mathbb{R}^{n\times c}$ is a class indicator matrix on ambiguously labeled samples with $Y_{ij} = 1$ if the label $l_i$ of sample $i$ is equal to $j$ and $Y_{ij} = 0$ otherwise.

In order to calculate the Laplacian matrix $\boldsymbol{L}$, the adjacency matrix $\boldsymbol{W}$ needs to be calculated, since $\boldsymbol{L} = \boldsymbol{D} - \boldsymbol{W}$, where $\boldsymbol{D} \in \mathbb{R}^{n\times n}$ is a diagonal degree matrix (defined in sec. 0). In our case $\boldsymbol{W}$ is approximated as $\boldsymbol{W} = \boldsymbol{Z} \cdot \boldsymbol{\Lambda}^{-1} \cdot \boldsymbol{Z}^T$, where $\boldsymbol{\Lambda} \in \mathbb{R}^{m\times m}$ is defined as: $\boldsymbol{\Lambda} = \sum_{i=1}^{n} Z_{ik}$. The solution of the eq. (8.12) has the form of:

$$\boldsymbol{A}^* = \left(\boldsymbol{Z}^T\boldsymbol{Z} + \gamma\hat{\boldsymbol{L}}\right)\boldsymbol{Z}^T \cdot \boldsymbol{Y} \tag{8.13}$$

Each sample label is, then, given by:

$$\hat{l}_i = arg \max_{j\in\{1,\dots,c\}} \frac{\boldsymbol{Z}_i \cdot a_j}{\lambda_j} \tag{8.14}$$

where $\boldsymbol{Z}_i \in \mathbb{R}^{1\times m}$ denotes the $i$-th row of $\boldsymbol{Z}$, and the normalization factor $\lambda_j = \boldsymbol{1}^T \boldsymbol{Z} \boldsymbol{\alpha}_j$ balances skewed class distributions.

### 8.4.3 Target detection

Having constructed a training set, $S = \{(\boldsymbol{f}_i, l_i)\}_{i=1}^{n}$, a binary classifier, capable to discriminate pixels that depict some part of a maritime target from pixels that depict the background, can be trained. Here, we choose to utilize linear SVMs to transact the classification task.

In the framework of maritime detection, SVM must be able to handle unbalanced classification problems, due to the fact that maritime target usually occupy the minority of captured frames' pixels let alone their total absence from the scene for large time periods. To address this problem, the misclassification error for each class is weighted separately. This means that the total misclassification error of eq. (3.10) is replaced with two terms:

$$c\sum_{i}^{n} \xi_i \rightarrow c_p \sum_{\{i|l_i=1\}} \xi_i + c_n \sum_{\{i|l_i=1\}} \xi_i \tag{8.15}$$

where $c_p$ and $c_n$ are constant variables that weight separately the misclassification errors for positive and negative examples. The solution of eq. (3.10) with the classification error of eq. (8.15) results to a trained SVM, which is capable to classify the pixels of new captured frames.

### 8.4.4 Detector adaptation to new visual conditions

A robust maritime surveillance system must retain high performance for long time periods. Thus, an SVM adaptation mechanism has to be developed, allowing the classifier to be adapted to dynamically changing visual conditions.

Let us denote as $V_1, \dots, V_{15}$ the fifteen visual attention maps described in sec. 8.3. We define the *average* visual attention map, $V_{avg}$, as:

$$V_{avg} = \frac{1}{15}\sum_{i=1}^{15} V_i \tag{8.16}$$

The elements' value, from eq. (8.16), express the overall probability a pixel to depict some part of a maritime target. Then, we define the refined visual attention map $V_r$ as the outcome of the element-wise multiplication between $V_{avg}$ matrix and background modeling algorithm output $B$.

Classifier adaptation process is triggered by an automated decision mechanism. Let us define as $V_{r,n}$ and $T_n$ the refined visual attention map and the output of the classifier, respectively, at frame $n$. When the difference between $V_{r,n}$ and $T_n$ exceeds a predefined threshold the decision mechanism triggers the adaptation process.

During the adaptation process the SVM classifier is retrained. We form a new training set that contains as elements the support vectors of the previously trained classifier, the $\kappa$ elements of $V_r$ that present the highest probability (positive samples) and have denoted as belonging to the negative class, and the $\kappa$ elements of $V_{avg}$ that present the lowest probability and have been denoted as background by the background modeling algorithm (negative samples).

Finally, we assume that visual conditions in a maritime environment are smoothly and gradually changing. This implies that the values for $\boldsymbol{w}$, $b$ and $\boldsymbol{\xi}$ of the adapted classifier should be close to the estimated values, $\overline{\boldsymbol{w}}$, $\bar{b}$ and $\bar{\boldsymbol{\xi}}$, of the previously trained classifier. To reduce the time required for classifier retraining, the aforementioned assumption, allows us to speed up the convergence of the optimization algorithm, which seeks for a solution to the problem defined in eq. (3.10), by restricting the feasible solutions region (set the initial values of the under optimization parameters to the values of $\overline{\boldsymbol{w}}$, $\bar{b}$and $\bar{\boldsymbol{\xi}}$).

## 8.5 Experimental results

Most of the algorithms were developed exclusively in C++ to achieve high performance; the overall system works almost in real time, 17fps, for frames with dimensions 384 × 288 pixels. There is, also, code in Python[8], concerning visual attention maps construction, available to download. The performance of each system's component have been checked separately; extracted features were evaluated in terms of discriminative ability and importance, semi-supervised labeling for the predicting outcome and, finally, the binary classifier for its performance.

The data sets describe real life scenarios, in various weather conditions. Data consists of recorded videos from cameras mounted at the Limassol port, Cyprus and Chania old port, Crete, Greece. Monocular cameras were recording videos streams depicting maritime traffic for over one year. Unfortunately, for the vast majority of the video frames, maritime targets are absent from the scene. In order to deal with such cases, we manually edited the videos and kept only the tracks that depict intrusion of one or more targets in the scene. Then, we manually labeled the pixels of key video frames, *keyframes*, to create a ground truth dataset for evaluating our system.

Keyframes originate from raw video frames, on a constant time span equals to $t$ frames i.e. frames that correspond to time instances $t, 2t, 3t, \ldots$ . The time span is selected to be 6 seconds, which means that one frame out of 150 is denoted as keyframe. We followed this approach for practical reasons. Firstly, it would be impossible to manually label all video frames at a framerate of 25 fps. Also, the time interval of 6 seconds is small enough to allow the detection of the intrusion of a maritime target in the scene. At this point it has to be clarified that feature extraction task, as well as the binary classification are performed for all frames of a video track. Keyframes are used only for system's performance evaluation.

### 8.5.1 Evaluation of extracted features

In this section, we examine if the extracted features are able to describe appropriately frame's pixels and, consequently, provide useful information that will facilitate the classification task. In addition, the extent that each one of the features affects the classification task (i.e. how important a feature is) is examined. Results concerning the importance of features, may allow us to discard some of them, in order to speed up system's performance. To evaluate features information, we utilized the keyframes' ground truth data. The feature extraction task results in a 16-dimensional feature vector for each pixel in a frame. The quality of features' information is evaluated through dimensionality reduction and samples plotting, in order to visually examine their distribution in space, see Fig.5. The two classes, as shown in Fig.5, are linearly separable, which suggests high quality features. The small amount of positive samples, which lie inside the region of the negative class, correspond to maritime targets' contours and probably occurred due to segmentation errors during manual labeling.

[8] https://github.com/kmakantasis/poseidon_features.git

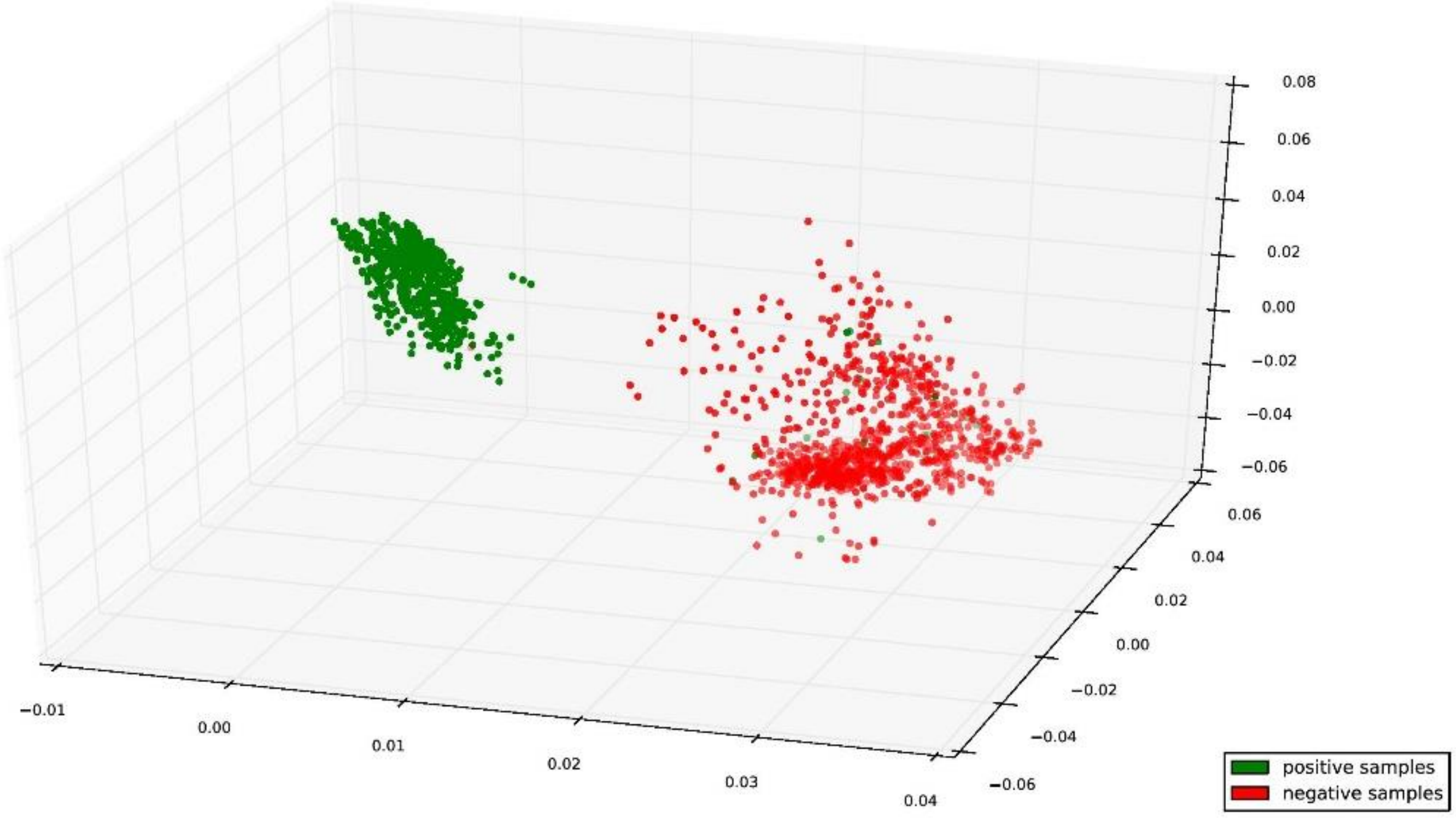

*Figure 8.5. Positive and negative samples plotted in 3-dimensional space. Randomized PCA was used to extract the three dominant components of the dataset. The class containing positive samples can be linearly separated from the class containing negative samples. The small amount of positive samples that lie inside the region of the negative class, correspond to maritime targets' contours.*

### 8.5.2 Evaluation of semi-supervised labeling

In order to evaluate semi-supervised labeling, we assume that manual labeling of keyframes contains no segmentation errors. The ratio of the representative samples in relation with the ambiguously labeled samples is the only factor that affect the performance of labeling algorithm.

As shown in Figure 8.6, the labeling error is lower than 2% when the ratio of the representative samples in relation with the ambiguously labeled samples is over 40%. When the ratio is smaller than 40% the labeling error is linearly increasing and it reaches the value of 5.7% when the ratio of representative samples is 10%.

The choice for an appropriate value for the ratio of representatives is inherently dependent on the quality of human based labeling. If labeling is the result of a rough image segmentation, a lot of the labeled pixel will carry the wrong label. In such cases the aforementioned ratio must be set to a small value. The most representative samples from each class is assumed that carry the right label, while the labels of the rest of the samples must be reconsidered. In our case, we required the user to segment the frame in a very careful way, which implies that the vast majority of the pixels will carry the right label. For this reason we set the ratio value to 40%. The semi-supervised labeling algorithm with 40% of representatives is expected to re-label 1.7% of the samples.

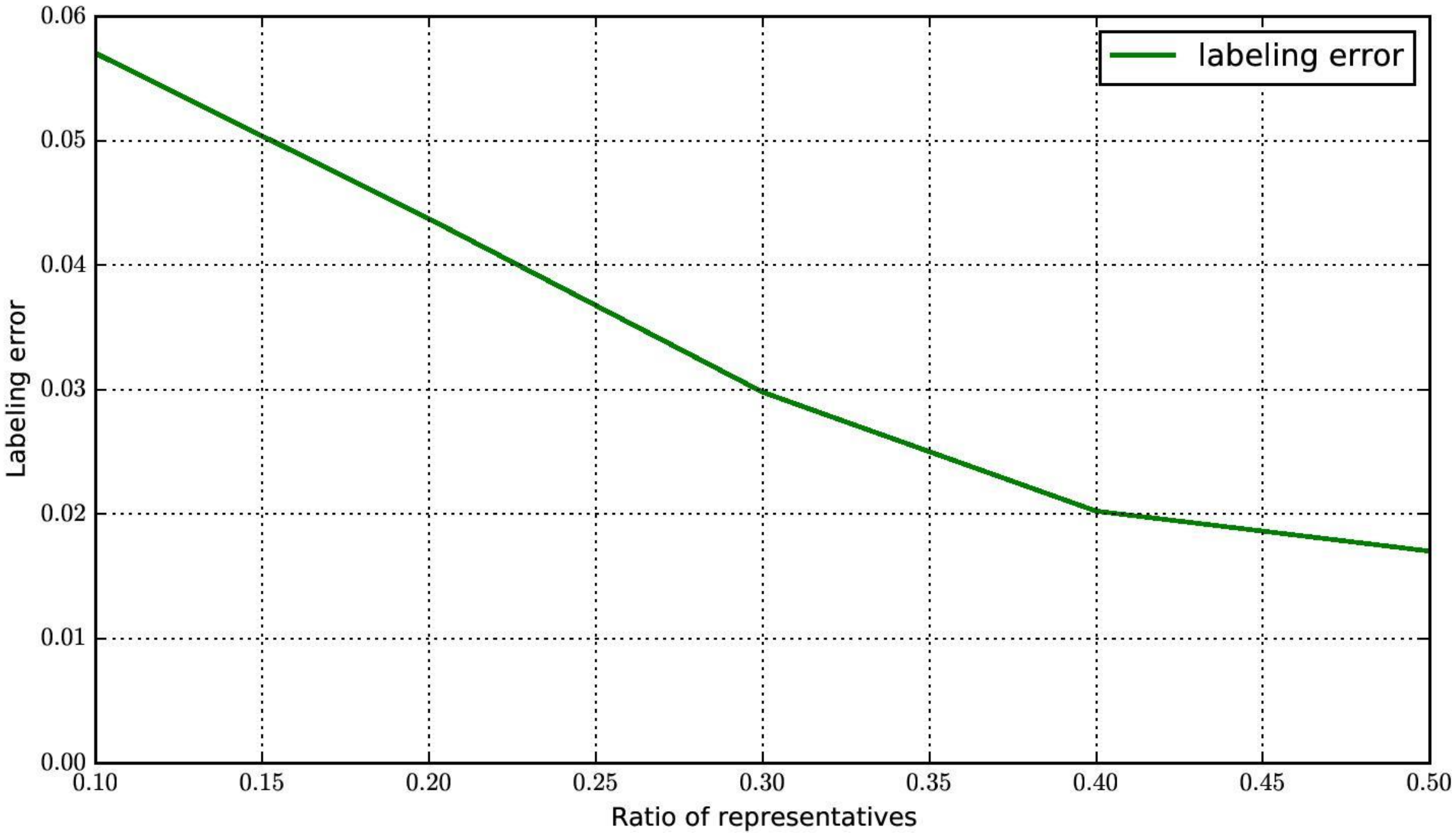

*Figure 8.6. Semi-supervised labeling performance. When ratio of representative samples is over 40% the labeling error is lower than 2%. When the ratio of representatives is lower than 40% the error is linearly increasing till the value of 5.7% for 10% of representatives.*

## 8.6 Conclusions & future work

A vision based system, using monocular camera data, is presented. The system provides robust results by combining supervised and unsupervised methods, appropriate for maritime surveillance, utilizing an innovative initialization procedure. The system offline initialization is achieved through graph based SSL algorithm, suitable for large data sets, supporting users during segmentation process. Another advantage is the automated adaptation of the system to new environments, in real time.

Extensive performance analysis suggest that the proposed system performs well, in real time, for long periods without any special hardware requirements and the without any assumptions related to scene, environment and/or visual conditions. Such system is expected to significantly support the local authorities, or anyone interested in maritime surveillance without any significant additional cost.

Further extensions of the system are possible in many fields. At first, incorporation of a target recognition system would be intriguing. Another idea would be the operation in parallel with other proposed methodologies in a voting-based system.

# *Chapter* IX: Searching for Foundation Flaws

*I do not fear computers. I fear the lack of them.*

*Isaac Asimov, American author*

## 9 Non-destructive flaws detection in foundation piles

In this chapter, we exploit the unlabeled data in order to improve detection accuracy for defect detection in foundation piles. In order to do so, we employ a graph based approach and noise modeling techniques for the mapping of the waveforms onto a new manifold. Given a set of waveforms, an experts' help is required only for a small subset (i.e. ≤ 40% of the available samples).

The proposed approach reduces labeling effort, which is both costly and time consuming. Such an approach encourages the data sampling, since larger the data base better the classification accuracy. Data availability was not an issue, since we used numerical simulation in order to create various pile types with defects.

### 9.1 Introduction

Structures foundation in the form of concrete piles is a commonly adopted approach in many cases. These piles are usually built by using precast and cast-in-situ techniques. Sometimes, "necks" or "bulbs" may be created in the process of drilling. These defects may affect the bearing capacity of the piles. Hence the structural evaluation and monitoring of new and existing piles are becoming increasingly important. In this chapter we deal with the detection of such defects using a graph based approach. Piles' surfaces' oscillations, produced through numerical simulation, serve as row data for the detection mechanism.

Surface oscillation is non-destructive testing (NDT) approach (Garnier et al., 2011), adopted for many practical reasons. The test is based on wave propagation theory; the impact generates a compression wave that travels down the pile at a constant wave speed. Changes in cross sectional area (e.g. reduction in diameter) produce wave reflections. Engineers would desire the use of an intelligent software tool, able to automatically analyze these complex waveforms generated as a result of a pile integrity test (PIT) testing and produce classification outputs, regarding piles' condition states.

Towards that direction, an innovative work for the NDT of piles is employed using a mixture of state of the art soft computing techniques, appropriate feature extraction and data generation procedures. Regarding the classification process, a graph based label propagation approach is adopted over a graph, which is constructed under SSL assumptions (Zhu and Goldberg, 2009b). The innovation of the current methodology is the fully automatic post processing technique, which results in high classification performance, easy implementation and noise tolerance, using a limited training sample.

The results have been obtained on experimental data originating from numerical experiments. These data, as described below, simulate as much as possible real-life phenomena of "neck" or "bulb" type structural defects. Application of novel intelligent classification algorithms for defects' prediction should be first experientially validated and tested under laboratory conditions to guarantee the successful performance of the classifier and then to be validated on real-data, which requires huge financial effort while it is also risky in such infrastructures.

### 9.1.1 Related work

Surface reflection techniques are a common approach in the foundation assessment field. The work of (Huang and Ni, 2012) investigates the relative performance of the sonic echo (SE), impulse response (IR), and parallel seismic (PS) tests using a field constructed pile foundation incorporating simulated defects. The work of (Hola and Schabowicz, 2010) presents the state of the art approaches of NDT on building structures. Such studies suggest the NDT approach appropriateness.

Numerical simulation of NDT cases (Haddad, 2010) is an alternative approach, which allow the investigation of various defect types, providing accurate results very close to real life situations. The work of (Huang et al., 2010) focuses on drilled shaft defects identification. Other approaches focuses on the testing conditions, such as the effects of the source on wave propagation (Chai et al., 2010). Generally, the quality of numerically generated waveforms is close to the actual ones. Thus, accurate interpretation of such waveforms could be extremely beneficial.

There are many methods for the analysis of such waveforms; artificial neural networks (ANNs) is a common one. Relating work on inverse analysis and defect identification problems solved by optimization and ANNs can be found in (Stavroulakis et al., 2003, 2004; Stavroulakis, 2000). Relative work can be, also, found in (Tam et al., 2004; Zhang and Zhang, 2009). These approaches exploit relatively simple ANN topologies, using a few selected inputs. Therefore, the users must have quite extended experience in order to choose the measurements to use, while the effectiveness of the neural network cannot be guaranteed or optimized.

## 9.2 Proposed methodology

The proposed approach is suitable for low strain integrity tests, carried out in time domain. In time domain reflectometry, the wave is generated by a hand held hammer blow impact and the response as a function of time is picked up by multiple accelerometers, placed on piles' head and around it, on a circle base, close to the location of hammer blow. Monitoring and analysis of these reflections form the basis of PIT (Schauer and Langer, 2012).

In our case similar tests, with the ones performed in the laboratories, are modelled by employing a coupled finite element method (FEM) together with scaled boundary finite element method (SBFEM) approach (Schauer et al., 2012). Numerical simulation is used for the data generation. Graph label propagation (Wang and Zhang, 2008) is, then, used for the defect identification. Data generation involves the generated waveforms (time domain), while graph detectors provide results regarding the integrity testing, by exploiting the information of *all available data* provided, following a specific feature extraction.

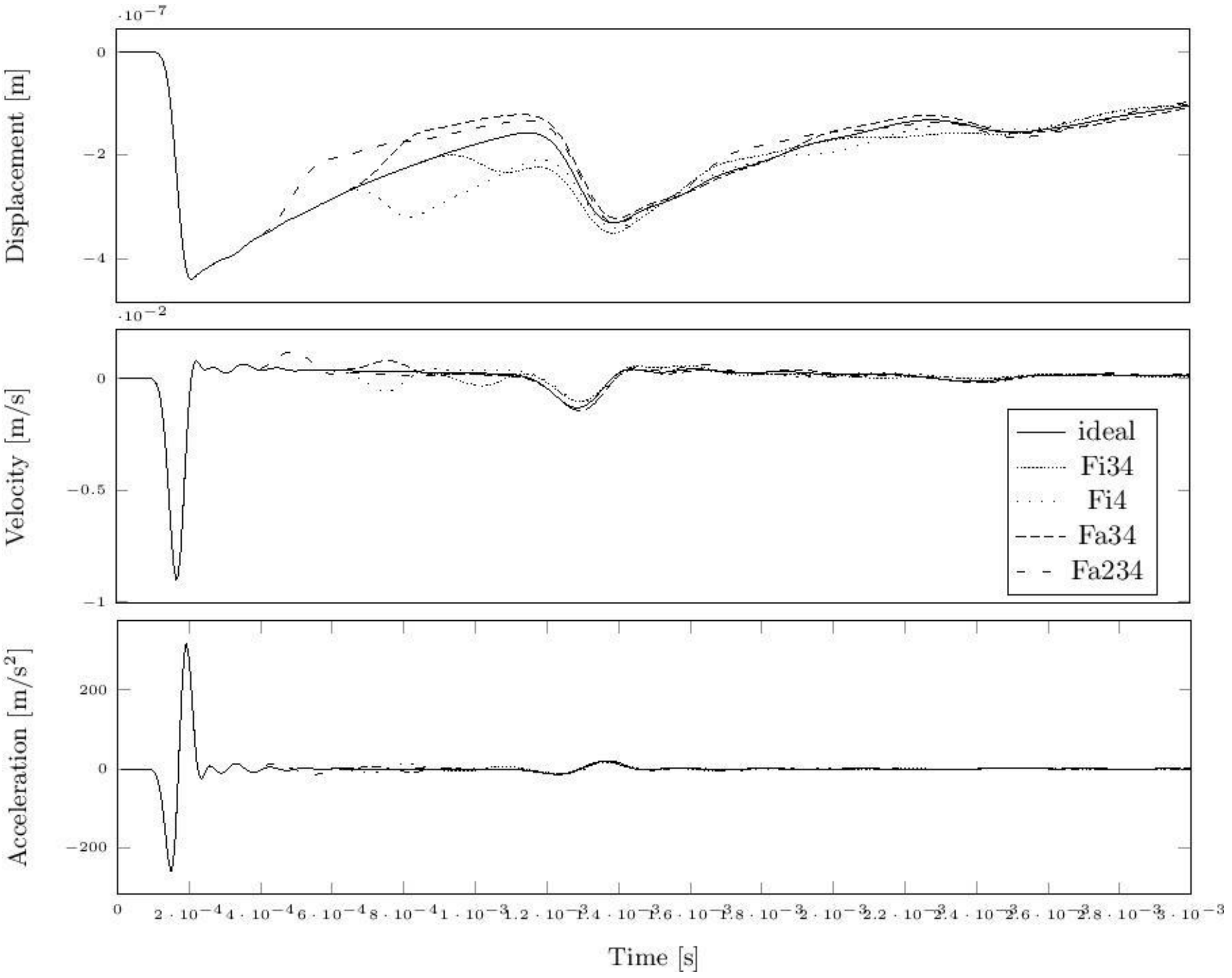

*Figure 9.1.Time dependent plots for displacement, velocity and acceleration for pile configuration: ideal, Fi34, Fi4, Fa34 and Fa234. More information about piles' defects can be found in sec. 3.1.*

### 9.2.1 Piles numerical simulation

In order to simulate the wave propagation through the piles a coupled FEM and SBFEM approach is used. This approach satisfies Sommerfeld's radiation condition and allows simulating an infinite half space. This ensures that the applied impulse will not be reflected at the artificial boundary which is introduced by the boundary of the numerical discretization. The coupled approach proposed here requires only the discretization of a small domain compared to a purely FEM-based approach.

FEM and SBFEM are used to model the near field and far field, respectively. The equation of motion at an arbitrary time step can be written as:

$$\begin{bmatrix} \boldsymbol{M}_{nn} & \boldsymbol{M}_{nf} \\ \boldsymbol{M}_{fn} & \boldsymbol{M}_{ff} \end{bmatrix} \ddot{\boldsymbol{u}} + \begin{bmatrix} \boldsymbol{K}_{nn} & \boldsymbol{K}_{nf} \\ \boldsymbol{K}_{fn} & \boldsymbol{K}_{ff} \end{bmatrix} \boldsymbol{u} = \begin{bmatrix} \boldsymbol{p}_{nn} \\ \boldsymbol{p}_{ff} \end{bmatrix} - \begin{bmatrix} \boldsymbol{0} \\ \boldsymbol{p}_{b} \end{bmatrix} \tag{9.1}$$

where the vector $\boldsymbol{u}$ represents the nodal displacement, $\ddot{\boldsymbol{u}}$ the nodal acceleration, and $\boldsymbol{p}$ denotes the applied nodal forces. $\boldsymbol{M}$ is the mass matrix and $\boldsymbol{K}$ stands for the stiffness matrix. Here, matrix blocks with subscript $nn$ contain the nodes of the near field while blocks with subscript $ff$ comprise the nodes of the far field. The coupling of near and far field nodes is reflected in those blocks subscribed with $nf$ and $fn$. Vector $\boldsymbol{p}_b$ denotes the far field influence on the near field, so that the behavior of the infinite half space can be applied to the FEM sub-domain as a load.

The far field is represented by the forces of the far field $\boldsymbol{p}_b$ at the interface given by the convolution integral:

$$\boldsymbol{p}_b(t) = \int_0^t \boldsymbol{M}^{\infty}(t-\tau)\, \ddot{\boldsymbol{u}}(\tau) d\tau \tag{9.2}$$

where $\boldsymbol{M}^{\infty}$ is the acceleration unit-impulse response matrix. Detailed information on how the acceleration unit-impulse response matrices are assembled are published in (Schauer et al., 2012; Wolf et al., 2003). An illustration of the generated waveforms at the piles' surface is shown in Figure 9.1.

### 9.2.2 Feature extraction

Once a load $p$ is applied, at the top and center of the pile an oscillation occurs as a result of wave propagation through the piles' structure. For a predefined time duration $T_t$ the oscillating patterns $\boldsymbol{O}_{p,i}$ are recorded for every node $i$. These patterns have the form of:

$$\boldsymbol{O}_{p,i} = \begin{bmatrix} x_{d,\mathrm{i}} & y_{d,\mathrm{i}} & z_{d,\mathrm{i}} \\ x_{v,i} & y_{v,\mathrm{i}} & z_{v,\mathrm{i}} \\ x_{a,\mathrm{i}} & y_{a,\mathrm{i}} & z_{a,\mathrm{i}} \end{bmatrix} \tag{9.3}$$

where d, v and a stand for displacement, velocity and acceleration respectively. So all information regarding a piles' behavior is expressed by:

$$\boldsymbol{S}_{pile} = \left[\boldsymbol{O}_{p,1} \cdots \boldsymbol{O}_{p,m}\right] \tag{9.4}$$

where $\boldsymbol{S}_{pile}$ denotes the available information about the waveform in any of the $m$ nodes for a total time $T_t$.

A waveform that describes the oscillating behavior (or recorded observation 340 for simplicity) for each node represents the base for our analysis. However, the recorded observations for each of the piles' nodes are too large to be processed and it is suspected to be non-informative after a time period. The simplification of the amount of resources, required to describe a large set of data accurately, is possible taking *two main assumptions* into account:

1. Ideal pile behavior is known. That can be achieved through CAD models and numerical simulation. Every one of the investigating nodes has its corresponding ideal waveform. As we will see that is of major importance during the feature extraction of the data.
2. There is a transient period with sufficient information, for every observed node. In other words, a short period of time includes most of the important signal variations, needed by the model in order to recognize the type of the defect.

The feature extraction is based on signal subtraction, a common technique applied in noise modelling (Lipponen and Tarvainen, 2013). Thus, the first step is the subtraction stage, so that:

$$\boldsymbol{S}_{pile}^{new} = \boldsymbol{S}_{pile}^{check} - \boldsymbol{S}_{pile}^{ideal} \tag{9.5}$$

where $\boldsymbol{S}_{pile}^{ideal}$ denotes the generated signal from a pile without defects. We also define:

1. The transient period time $T_{trans}$. During this period the wave propagates from the pile's top to the bottom and then to the top again. After $T_{trans}$, waveforms get a complicated form due to the waves deflections and reflections.
2. Feature space dimension, $n_v$. We map each oscillation pattern to a new space $\mathbb{R}^{n_v \times 1}$. Values of $n_v$ less than 6 are unable to create descriptive feature vectors, while values greater than 40 may require additional training (labeled) data for a smooth classification performance.
3. The mapping function from oscillation space to feature space. In our approach, we used three alternatives: Mean absolute error, mean square error and difference in specific time steps.

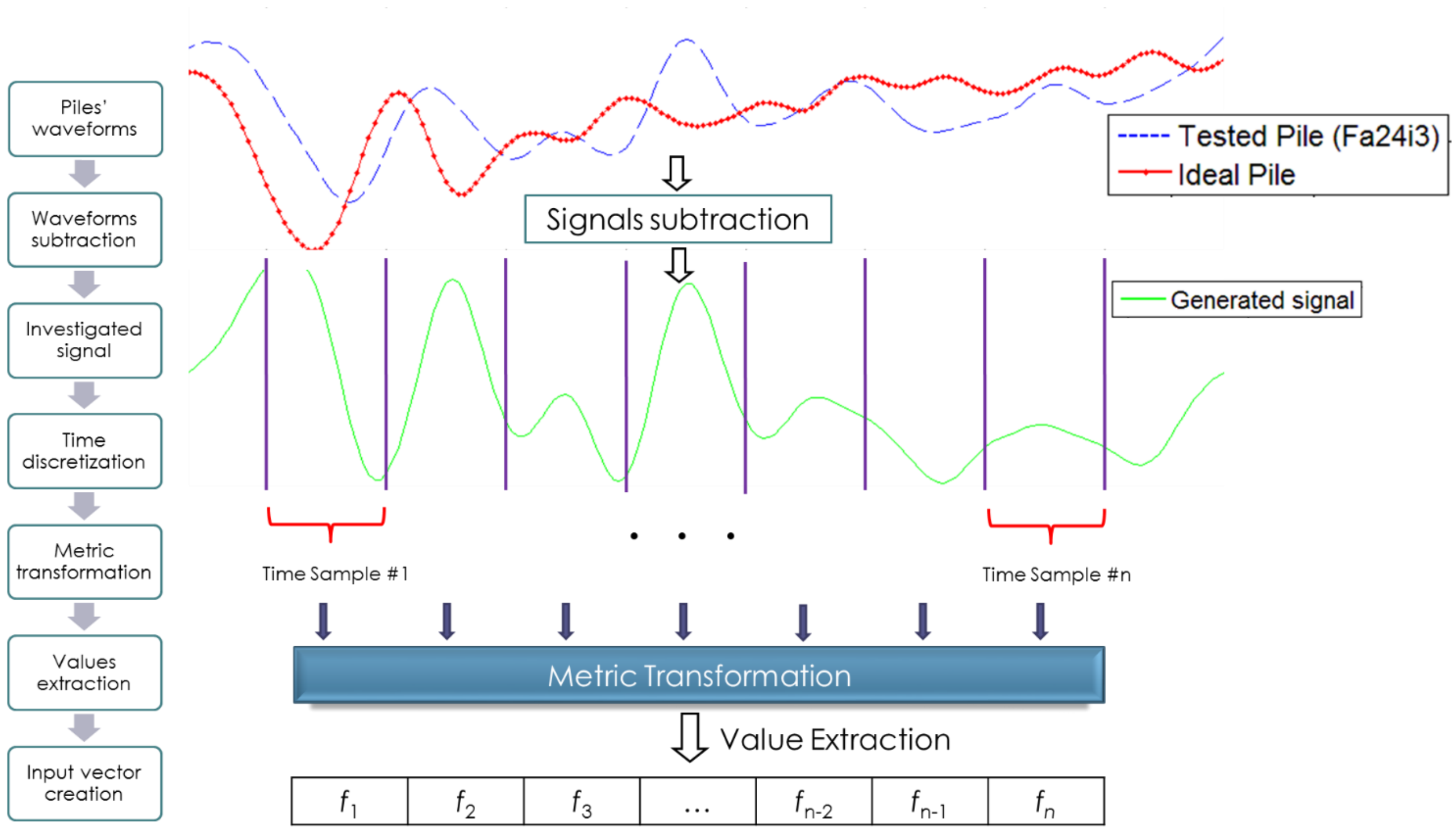

*Figure 9.2. Input vector creation process for the neural detector. The process exploits the differences between the investigated and the ideal pile.*

### 9.2.3 Possible limitations

The entire process is based on signals comparisons and spans a very limited range of defect cases. Thus, there is the possibility of feature ineffectiveness in different defect scenarios. Additionally, this chapter case study involves numerically simulated pile signals; no noise is expected as a result of minor voids, surface fluctuations, etc., which is a common case in actual foundation piles. The adopted feature extraction process (sec. 9.2.2) could partially deal with such noise via averaging operators. However, this has not been tested on actual field.

## 9.3 Experimental results

The defect recognition can be seen as a classification problem. Assume that a pile is separated in $q$ parts. Also, the available information is limited in few waveforms, recorded on the pile's surface. We have to classify each of these parts in one out of three categories. The "neck"-category indicates the existence of a "neck", i.e. smaller radious than expected. The "bulb"-category indicates the existence of a "bulb", i.e. greater radious than expected. There is, also, the "no-defect"-category, where there are neither "bulbs" nor "necks". Feature extraction and defect recognition routines are written in MatLab code.

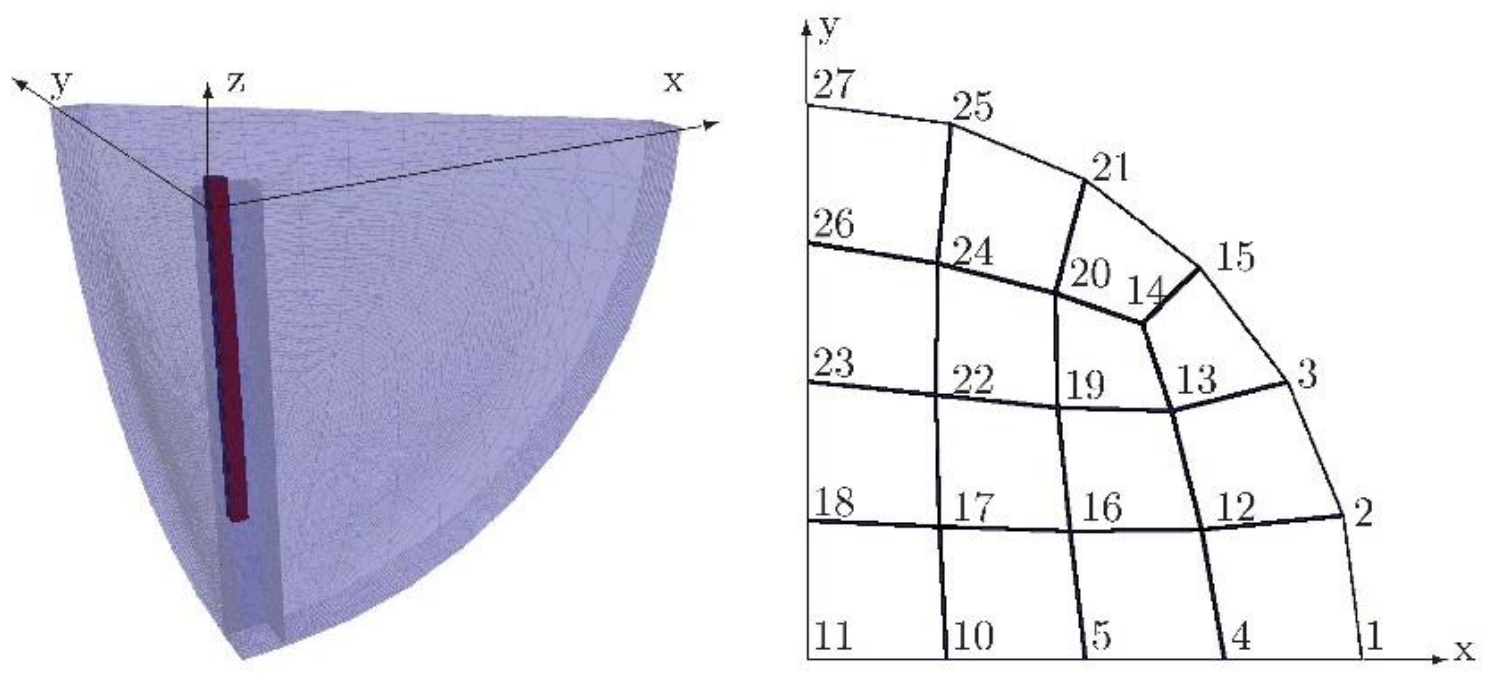

*Figure 9.3. Left: The FEM near field discretization includes pile and surrounding soil; Right: Top view at the piles surface and the corresponding node numbers.*

### 9.3.1 Piles set description

Different three-dimensional pile configurations are analyzed. All investigated piles are modelled as floating piles, since no bedrock is taken into account. One clean pile without defects, Figure 9.4 (ideal), is discretized. Length $l_0$ and radius $r_0$ are chosen as $2.1m$ and $0.1m$, respectively. The surface of the ground is defined at $0.0m$, the pile's head is located $+0.1m$ over the surface, while the piles' toe is at $-2.0m$ in the ground. Additional piles with defects are discretized as well, the geometries of these modified piles are shown in Figure 9.4. The finite element mesh is shown in Figure 9.3. The near field is discretized by 116974 tetrahedral elements and 28140 hexahedral elements and 151833 degrees of freedom. The attached far field is discretized by 549 quadrangular elements and 1794 degrees of freedom. All elements are using linear finite and scaled boundary finite elements. The minimum elements length is $l_{min} = 7.10347 \times 10^{-3}\ m$ and the maximum length is $l_{max} = 6.46424 \times 10^{-1}\ m$.

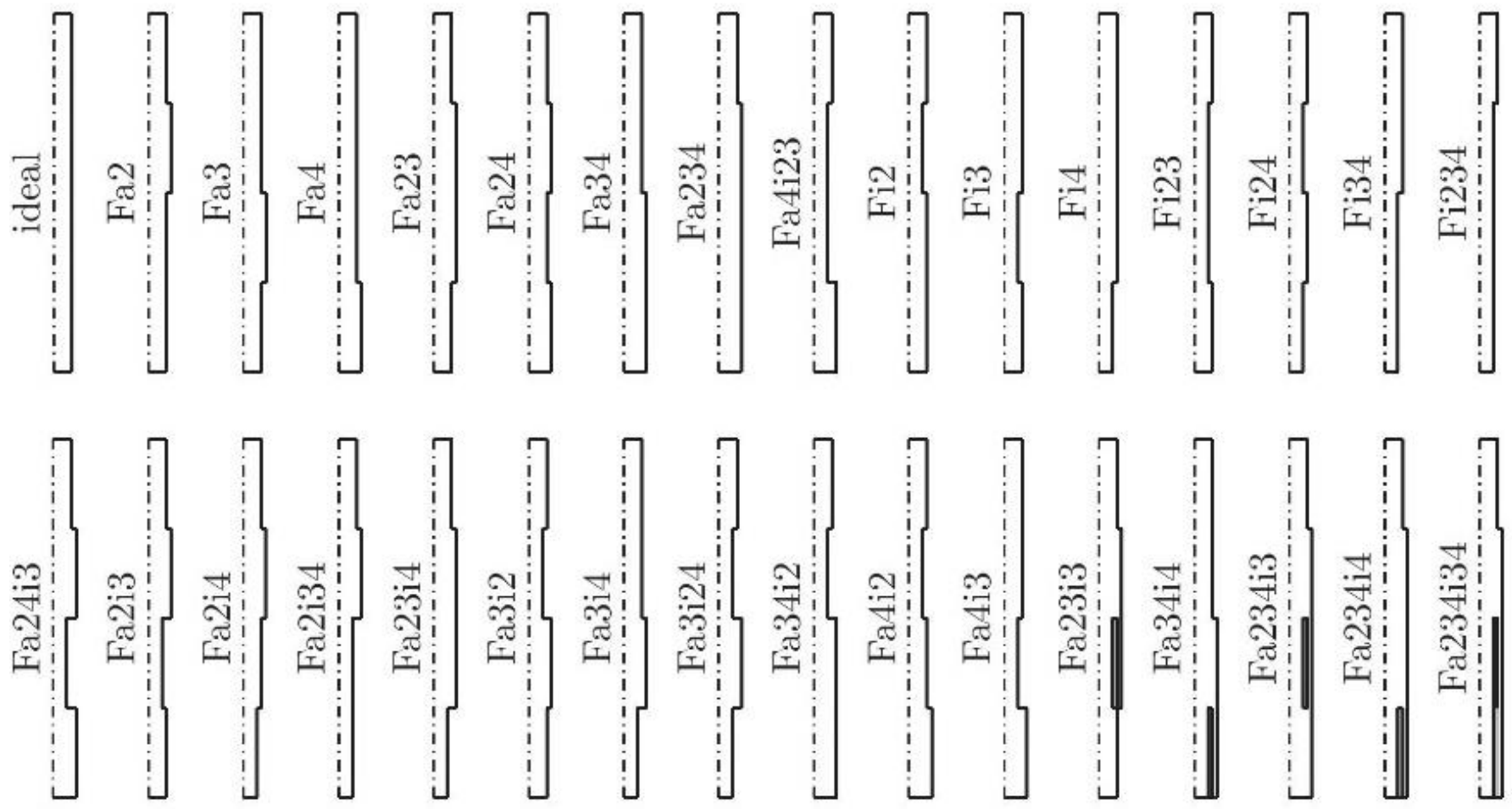

*Figure 9.4. Pile geometries and naming of the cases, half the longitudinal section of the pile is pictured. The pile is divided into 4 sections enumerated from 1 to 4 top down. The "a" stands for additional pile material and the "i" inner distortion of pile cross section. So, "Fa2" gives additional material in section 2. The last five geometries in the second row have an internal distortion, which is represented with a soil inclusion in the concrete body of the pile.*

### 9.3.2 Training, validation and evaluation sets

Let us first focus on the data creation. The data, for the graph construction, originates from the subtraction between two signals (i.e. the ideal and the examined pile) in the process shown in Figure 9.2. Although the signal duration is 2000 time steps (or 0.012 seconds), only the first 400 time steps of the transient period were utilized, since after that period the wave signal is backward propagated causing interference in the signal altitude. This way, we would create an input signal of size $400 \times 1$, which causes misclassification issues, due to the high dimensionality. To handle this problem, we equally downsample the input signal by 10. Thus, input vector is of size $40 \times 1$.

For every input vector there is a corresponding output vector of size $k \times 1$, where $k$ denotes the number of pillar parts that are investigated (in our case $k = 4$). The number of parts was selected to facilitate the numerical simulations, in terms of computational complexity. However, division into greater number of parts is feasible, for simulating more complex structures. The first part is located above the ground.

Specific nodes were used to form the training data, while the remaining formed the evaluation data. The two different data sets are described in Table 9.1. For each node, waveforms of nine different cases were available:

1. $x$-axis: displacement ($x_d$), velocity ($x_v$) and acceleration ($x_a$)
2. $y$-axis: displacement ($y_d$), velocity ($y_v$) and acceleration ($y_a$)

3. $z$-axis: displacement ($z_d$), velocity ($z_v$) and acceleration ($z_a$).

However, due to modelling assumptions, regarding the boundary conditions of the FEM-SBFEM approach, specific nodes oscillation patterns had to be excluded from the data generation (as non-informative waveforms). These oscillations refer to $x$ and $y$ axes, but not to $z$-axis. Due to symmetry boundary conditions these $x$ and $y$ data should be equal to zero. If not they are practically zero.

*Table 9.1. Training and evaluation nodes for each of the created data sets.*

| Data set | Training nodes | Evaluation nodes |
|---|---|---|
| **TS1** | {1, 15, 25 } | { 2, 3, 4, 5, 10, 11, 12, 13, 14, 16, 17, 18, 19, 20, 21, 22, 23, 24, 26, 27 } |
| **TS2** | {17, 25 } | { 1, 2, 3, 4, 5, 10, 11, 12, 13, 14, 15, 16, 18, 19, 20, 21, 22, 23, 24, 26, 27 } |
| **TS3** | {2, 19 } | { 1, 3, 4, 5, 10, 11, 12, 13, 14, 15, 16, 17, 18, 20, 21, 22, 23, 24, 25, 26, 27 } |
| **TS4** | {1, 21 } | { 2, 3, 4, 5, 10, 11, 12, 13, 14, 15, 16, 17, 18, 19, 20, 22, 23, 24, 25, 26, 27 } |

### 9.3.3 Classification performance

The proposed approach has been evaluated against two well-known classification approaches: $k$ nearest neighbors (Bhatia and Vandana, 2010) ($k$NN) and ANNs (Tan et al., 2011). All methods are characterized as soft labeled; corresponding outputs are not integers. Thus, we adopt a mapping procedure. For a given pile, a specific output vector, $\boldsymbol{P}_i = [p_1, p_2, p_3, p_4]$, is generated. The $p_i$ values, $i = 1, \dots, 4$, correspond to a certain defect type, $d_f$ , according to the following transformation:

$$d_f = \begin{cases} -1 & , p_i \in (-\infty, -0.5) \\ 0 & , p_i \in [-0.5, 0.5] \\ 1 & , p_i \in (0.5, \infty) \end{cases} \tag{9.6}$$

For the $i$-th part, value $df = -1$ suggests the existence of a “neck”, while value $d_f = 1$ suggests the existence of a “bulb”. Value $df = 0$ corresponds to non-detection of any defect. Range selection for the value intervals in eq. (9.6) stems from equal division of the detectors’ interval range of $[-1, 1]$ into three examined defect types.

*Table 9.2. Classification accuracy over the evaluation set. The simulation’s results correspond to average values over the different data sets, as described in Table 9.1.*

| | **ANN** | | | ***k*nn** | | | **Harmonic** | | |
|---|---|---|---|---|---|---|---|---|---|
| **Descriptor** | **MSE** | **MAE** | **Diff** | **MSE** | **MAE** | **Diff** | **MSE** | **MAE** | **Diff** |
| $\boldsymbol{x_d}$ | 0.366 | 0.369 | 0.362 | 0.359 | 0.363 | 0.370 | 0.356 | 0.377 | **0.382** |
| $\boldsymbol{x_v}$ | 0.224 | 0.225 | 0.222 | 0.247 | 0.259 | **0.265** | 0.252 | 0.255 | 0.260 |
| $\boldsymbol{x_a}$ | 0.345 | 0.361 | 0.376 | 0.364 | 0.381 | **0.400** | 0.347 | 0.362 | 0.375 |
| $\boldsymbol{y_d}$ | 0.468 | 0.490 | 0.486 | 0.488 | 0.500 | 0.508 | 0.498 | 0.508 | **0.515** |
| $\boldsymbol{y_v}$ | 0.287 | 0.300 | **0.310** | 0.292 | 0.307 | 0.310 | 0.241 | 0.242 | 0.238 |
| $\boldsymbol{y_a}$ | 0.452 | 0.480 | **0.502** | 0.442 | 0.454 | 0.481 | 0.380 | 0.382 | 0.394 |
| $\boldsymbol{z_d}$ | 0.706 | 0.715 | 0.723 | 0.764 | 0.743 | 0.738 | 0.831 | **0.839** | 0.837 |
| $\boldsymbol{z_v}$ | 0.725 | 0.736 | 0.758 | 0.704 | 0.706 | 0.710 | 0.817 | 0.819 | **0.830** |
| $\boldsymbol{z_a}$ | 0.689 | 0.691 | 0.696 | 0.735 | 0.736 | 0.732 | 0.787 | 0.764 | **0.789** |

ANNs and kNN were available through MatLab toolboxes. Harmonic label propagation function and weight matrix creation functions where provided by (Zhu et al., 2003) and (Liu and Chang, 2009) respectively. A random search among a variety of ANN topologies was performed. ANNs parameters' range is shown Table 6.1. The number of nearest neighbors was set to 4, for both *k*NN and Harmonic function approaches. Indicative classification results are shown in Table 9.2.

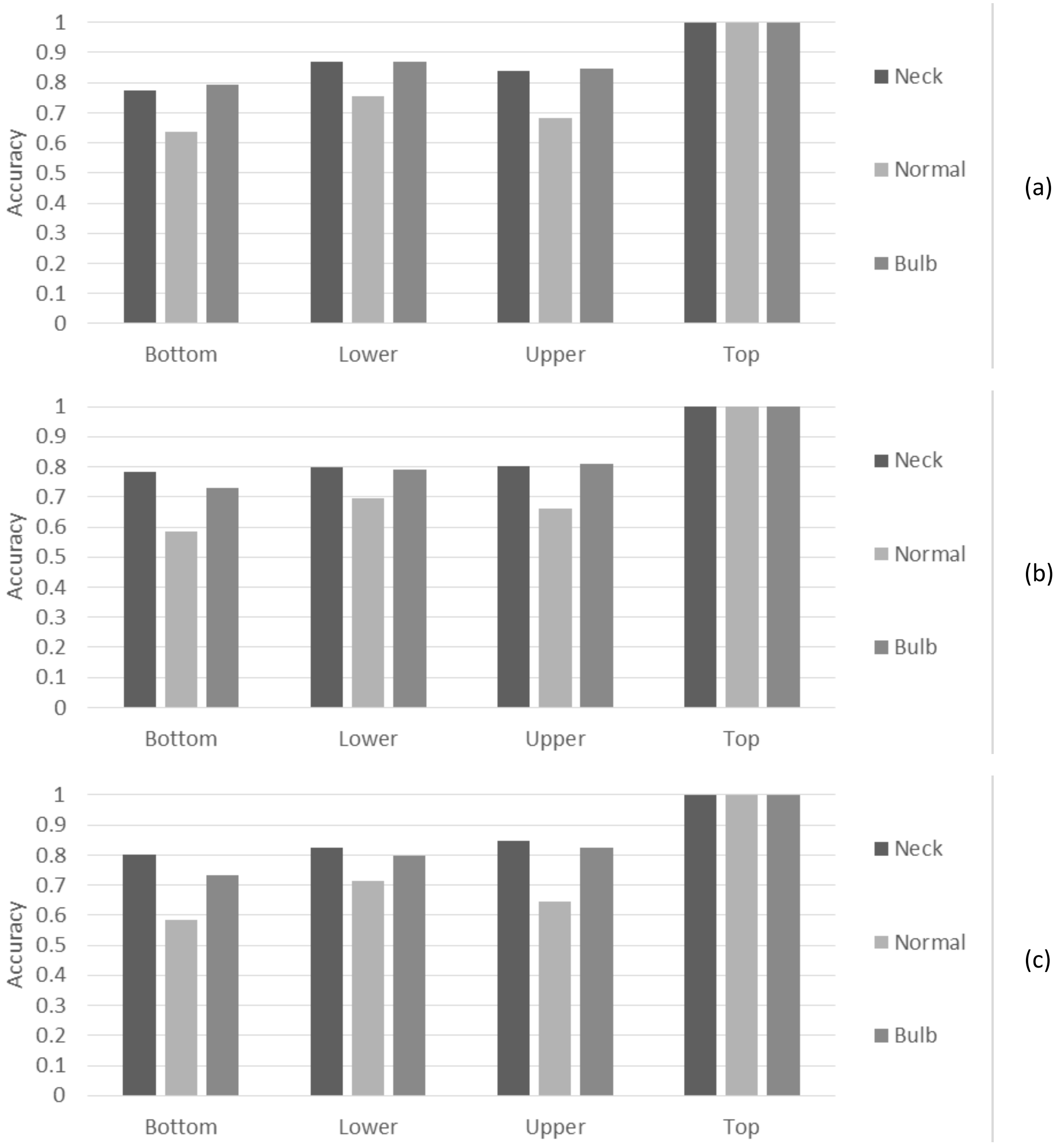

*Figure 9.5. Defect Classification accuracy for each one of the 4 pile's parts. Results are based on displacement observations in $\boldsymbol{x}$,$\boldsymbol{y}$,$\boldsymbol{z}$ axes using: (a) Difference in values (b) MAE and (c) MSE as quality metric.*

SSL graph based approaches limitations do not apply in our case, although such methods scale badly as the number of data raises (i.e. $O(n^2)$). In this particular field, we expect no more than few hundreds of samples (i.e. waveforms), even in real life cases. Another possible limitation is the transductive nature of the approach; graph based approaches are unable to handle new data. In such case, we have to recreate the graph and all the corresponding matrices.

The proposed approach performs better than the other commonly used techniques. Compared to the ANN, harmonic function needs a considerably smaller amount of labeled data and requires less heuristic approaches for the best topology definition. Generally, ANN performance highly depends on various parameters regarding the detector structure, such as number of hidden layers, neurons and training epochs. Compared to the $k$NN, Harmonic function exploits additional information of the unlabeled data. Thus, as labels propagate, edges' labels match the $k$ closest ones from the entire data space.

Results suggest that bulb or neck cases are easily identifiable. However, normal radius is classified either as neck or bulb more frequently (see Figure 9.5). The misclassification of non-defective pile parts can be partially explain by the waveform, which has a specific form defined by the defect(s) type. Such form doesn't change significantly if the defective pile part is followed by a non-defective part. Thus, we have a waveform, indicating a defect, crossing a non-defective part. According to the similarity mechanism of the proposed methodology, such form is more likely to indicate a defection, resulting in wrong classification. The misclassification rates grow as more defects appear in the same pile.

## 9.4 Conclusions & future work

One method to assess the behavior of a pile is to apply non-destructive testing through the use of low strain integrity tests in time domain. The wave is generated by a hand held hammer blow impact and the response, as a function of time, is picked up by multiple accelerometers, placed on pile head and around it, on a circle base, close to the location of hammer blow. Then we need to apply signal processing methods on the waveforms generated in order to detect the defects.

Initially, appropriate features were extracted in order to map piles' waveforms to meaningful short-length signals. Then, these features form a graph, where the labels propagate among the edges utilizing both labeled and unlabeled data. The performed experiments provide very promising results; the defect recognition rate is above 80%, when z-axis observations are used. On the contrary, the performance based on x and y axes behavior patterns is severely low. The problem formulation can easily be expanded in piles with more divisions (i.e. more than 4) and varying defects' diameters. Although, a greater misclassification error is likely, as long as we deal with “neck” or “bulb” detection, the performance is expected to remain high.

Finally, we observe that high detection rates are achievable using only a handful set of samples for training. The detection rates are also affected by the shape and the depth; deeper the defect harder to locate. Piles with more complex structure will be evaluated in future work. Furthermore, new adaptation strategies and detection techniques will be investigated to handle non-stationary waveforms cases.

# *Chapter* X: Adaptive Filtering in Accordance to User's Needs

***In the end, the character of a civilization is encased in its structures.***
***Frank Gehry, American architect***

## 10 Image meta-filtering techniques in cultural heritage applications

In this chapter a novel idea for additional filtering (meta-filtering) is proposed. The main purpose is the further refinement of existing data sets, obtained using various CBIR techniques. The reason behind the proposed mechanism lies in the need of *multiple uses of the same data sets in different applications* (e.g. subsets of the same data can be used for 3D reconstruction, tourism promotion, book publications, etc.). The model utilize a SSL approach for the creation of an appropriate distance metric, which is used for the filtering. User's feedback is involved only for a minor set of data, defined using OPTICS algorithm and sparse modeling representative selection. Such approach facilitates the refinement of retrieval results always under the scope of the end user needs.

### 10.1 Introduction

Cultural heritage (CH) digitization is a complex task that involves data retrieval and filtering from many alternative sources. Our work, described below, examines the filtering abilities of an innovative system (described below), over a given set of acquired images. The system exploits both user feedback and SSL assumptions, in order to capture the user's needs and minimize annotation effort.

effectiveness of any content-based image retrieval (CBIR) system (Valle and Cord, 2009) is severely depended on the selection of the appropriate distance metric. A common approach may involve, the well-known, Euclidean metric for computing distances between images, which are represented in some vector space. Unfortunately, such metric is often inadequate because of the well-known semantic gap between low-level features and high-level semantic concepts (Datta et al., 2005).

Currently, there are many approaches trying to deal with the semantic gap (Kumar K. and Gopal, 2014; Tang et al., 2012; Wang et al., 2008). One of them is the user feedback, which may be inefficient or exhausting in large data bases. In addition, user's needs change constantly, even over the same data base. In cultural heritage applications, there is always the need for data variations, especially when talking about 3D (Manferdini and Galassi, 2013) or 4D (Ioannides et al., 2013) reconstruction. Definitely, there are specific techniques to deal with the selection of appropriate images for reconstruction (e.g. (Konstantinos Makantasis et al., 2013)). Nevertheless, term "appropriate" is always defined by someone's needs.

User's needs or preferences may vary, from simple object detection (A. Doulamis, 2010; Lalos et al., 2014) (e.g. images of a specific statue), complex human motions (Doulamis, 2014) (e.g. images of children running around the fountain). Nowadays, such images, most likely, will be available given a large image data set retrieved from the internet using various techniques. However, it is very difficult, for the user, to search the entire data set, in order to select the specific images. In this chapter, we propose a suitable approach towards the retrieval of the appropriate images, given an initial data set, with respect to the user's preferences.

A mechanism that actuates over an initially retrieved set, given the user's preferences, can be beneficial in many ways. At first, such mechanism allows data sets to be further enriched/modified a priori, by using state of the art techniques and thus exploiting existing knowledge. Secondly, the user can change the features utilized, by the filtering mechanism, at any time in order to get more personalized results. Thirdly it handles outliers; chances are that totally irrelevant images will be in the originally retrieved set. In that case, it is very likely to be detected, during representative objects selection, and excluded. Finally, the semi-supervised approach minimizes user's feedback and, at the same time, captures current needs.

### 10.1.1 Place for improvement

A novel idea for additional filtering (meta-filtering) is proposed. The main purpose is the further refinement of data sets, obtained using various CBIR techniques. The reason behind the proposed mechanism lies in the need of *multiple uses of the same data sets in different applications* (e.g. subsets of the same data can be used for 3D reconstruction, tourism promotion, book publications, etc.).

Meta-filtering creates an appropriate image ranking mechanism, according to the user's requirements. The entire data set is presented to the user through few representative samples. The user has to annotate some of them, either as relevant or irrelevant to his current search. Then, in a fully automated way, an appropriate distance metric is calculated. Such metric is used by a ranking approach to assign a score for every image of the set. Images with high scores are presented to the user as the most relatives to his/her needs.

All the images of the set are employed to the calculation of the appropriate distance metric according to a graph based SSL scheme (a very common approach in SSL (Belkin et al., 2004; Goldberg et al., 2007; Goldberg and Zhu, 2006; Liu et al., 2010)). Thus, non-annotated images support the regularization creating a smooth distance metric (i.e. images with similar features will have similar ranking, given the user preferences as constraint). Such an approach minimizes users' efforts and is easily implemented in any dataset, given appropriate feature vectors.

## 10.2 Proposed methodology

The meta-filtering approach is based on three main phases. The first stage of the methodology aims at the detection of representative samples and their annotation, by the user, as relevant or irrelevant. The second stage involves the distance metric learning, according to the user defined relevance sets. The final stage ranks the rest of the images using both similarity and dissimilarity rankings, based on the previously stated distance metric.

### 10.2.1 Data collection and feature extraction

Initially, a large data set of images is collected from Flickr. The data retrieval was based in various parameters (including tags, location, etc.). Once the data set for a specific monument is gathered, additional features from the images are extracted. Feature selection is crucial; low-quality features can lead to low performance, since we utilize Euclidean based approaches.

Three MPEG-7 visual descriptors have been employed, as in (Kyriakaki et al., 2014), for the purposes of this research: Color Layout Descriptor (CLD), Scalable Color Descriptor (SCD) and Edge Histogram Descriptor (EHD). The specific descriptors were chosen due to their simplicity and small size, high processing speed, robustness, scalability and interoperability (Serna et al., 2011).

The CLD is a very compact one that captures the spatial layout of dominant colors of an image with coefficients of the Discrete Cosine Transform. The SCD is derived from an HSV color histogram with fixed color quantization, and its coefficients are encoded through a Haar transformation, while the EHD is a texture descriptor that detects edges of different angles by dividing the image into smaller blocks.

### 10.2.2 Mathematical formulation

The meta-filtering approach, based on a total raking approach for every image $x_j$, is described by the following equation:

$$r_j = \sum_{\substack{i=1 \\ i \neq j}}^{|P|} \frac{1}{\boldsymbol{w}_i^p d_{\boldsymbol{A}}(x_i, x_j)} - \sum_{\substack{i=1 \\ i \neq j}}^{|N|} \frac{1}{\boldsymbol{w}_i^n d_{\boldsymbol{A}}(x_i, x_j)} \tag{10.1}$$

where $r_j$is the overall ranking score for an image $j$, given its feature vector $x_j$, $|P|$ and $|N|$ denotes the size of user annotated images as positive and negative to current search respectively, $\boldsymbol{w}_i^p$ ($\boldsymbol{w}_i^n$) is a weight value for the importance of the *i*-th annotated positively (negatively) image, and $d_{\boldsymbol{A}}(x_i, x_j)$ is a distance metric defined in both user's annotated and the non-annotated images of the data set.

For any two given data points $x_i$ and $x_j$, let $d(x_i, x_j)$ denote the distance between them. To compute that distance, let $\boldsymbol{A} \in R^{m \times m}$ be a symmetric matrix, we can then express the formula of distance measure in a generic form as in eq. (3.11).

Similar to the approach of (Hoi et al., 2008), the distance metric learning (DML) problem is to learn an optimal $\boldsymbol{A}$ from a collection of data points $C$ on a vector space $R^m$ together with a set of similar pairwise constraints $S$ and a set of dissimilar pairwise constraints $D$. Both sets of constraints should be provided by the user as a relevance feedback in order to guide the problem to an acceptable solution.

At first, consider two sets of user defined pairwise constraints among data points:

$$\begin{gathered} \mathcal{S} = \{(x_i, x_j) | x_i \ relevant\ to\ x_j\} \\ \mathcal{D} = \{(x_i, x_j) | x_i \ irrelevant\ to\ x_j\} \end{gathered} \tag{10.2}$$

Given the above sets we can formulate a loss function of the form:

$$\min_{\boldsymbol{A}} \gamma_{\mathcal{S}} \sum_{q=1}^{d} \sum_{(x_i, x_j) \in \mathcal{S}_q} \|x_i - x_j\|_{\boldsymbol{A}}^2 - \gamma_{\mathcal{D}} \sum_{q=1}^{d} \sum_{(x_i, x_j) \in D_q} \|x_i - x_j\|_{\boldsymbol{A}}^2 + tr(\boldsymbol{XLX}^T \boldsymbol{A}) \tag{10.3}$$

The last term is a regularizer defined on the unlabeled data, where ***L*** = ***D***−***W*** is the Laplacian matrix, ***D*** is a diagonal matrix whose diagonal elements are equal to the sums of the row entries of ***W***, and *tr* stands for the *trace* function. The weight matrix, ***W***, is defined over all *n* images of the data set as:

$$W_{ij} = \begin{cases} 1, x_i \in \mathcal{N}(x_j) \ or\ x_j \in \mathcal{N}(x_i) \\ 0, \quad otherwise \end{cases} \tag{10.4}$$

where $\mathcal{N}(x_j)$ denotes the nearest neighbor list of the data point $x_j$. The aforementioned problem can be, alternatively, formulated as (Hoi et al., 2008):

$$\begin{gathered} \min_{A} t + \gamma_s tr(\boldsymbol{A} \cdot \boldsymbol{S}) - \gamma_d tr(\boldsymbol{A} \cdot \boldsymbol{S}) \\ s.t.\ tr(\boldsymbol{XLX}^T \boldsymbol{A}) \le t \\ \boldsymbol{A} \in S_+ \end{gathered} \tag{10.5}$$

Thus, the DML problem has been approached as a semi-definite problem (SDP), which can be solved efficiently with global optimum using existing convex optimization packages.

### 10.2.3 User involvement

Instead of providing random samples to the user, or letting him scan quickly the retrieved images, a more suitable, yet complicated, approach is adopted for the best paradigms selection. Thus, OPTICS algorithm (Daszykowski et al., 2004) is employed for finding density-based clusters in spatial data. The density values are used for the identification of sub clusters within retrieved data, by searching for local maxima as illustrated in Figure 10.1 (a). OPTICS algorithm requires as parameter, $k$, the number of objects in a neighborhood. We used the heuristic rule: $k = \lceil \log(num\ of\ images) / \log(2) \rceil$. Operator $\lceil(\cdot)\rceil$ rounds the value of $(\cdot)$ to the nearest integer greater than or equal to $(\cdot)$.

Each sub cluster is expected to be a large data collection of images. In order to extract the most important (descriptive) ones, the work of (Elhamifar et al., 2012) around sparse modeling for finding representative objects is employed. Their work is described in sec. 3.2.3.

The representative objects retrieved (*Figure 10.2*) are shown to user, who can define the relevance to his current search. The selection order defines, also, the importance of the datum; ranking scores depend on the order of selection through the $\boldsymbol{w}_i^p$ and $\boldsymbol{w}_i^n$ weights values. User can select any number of the representative images for annotation, as long as there is at least one relevant and one non-relevant image in the end. The small amount of images, provided by the sparse modeling approach, requires minimal effort for the annotation, even if user decides to annotate all the suggested images.

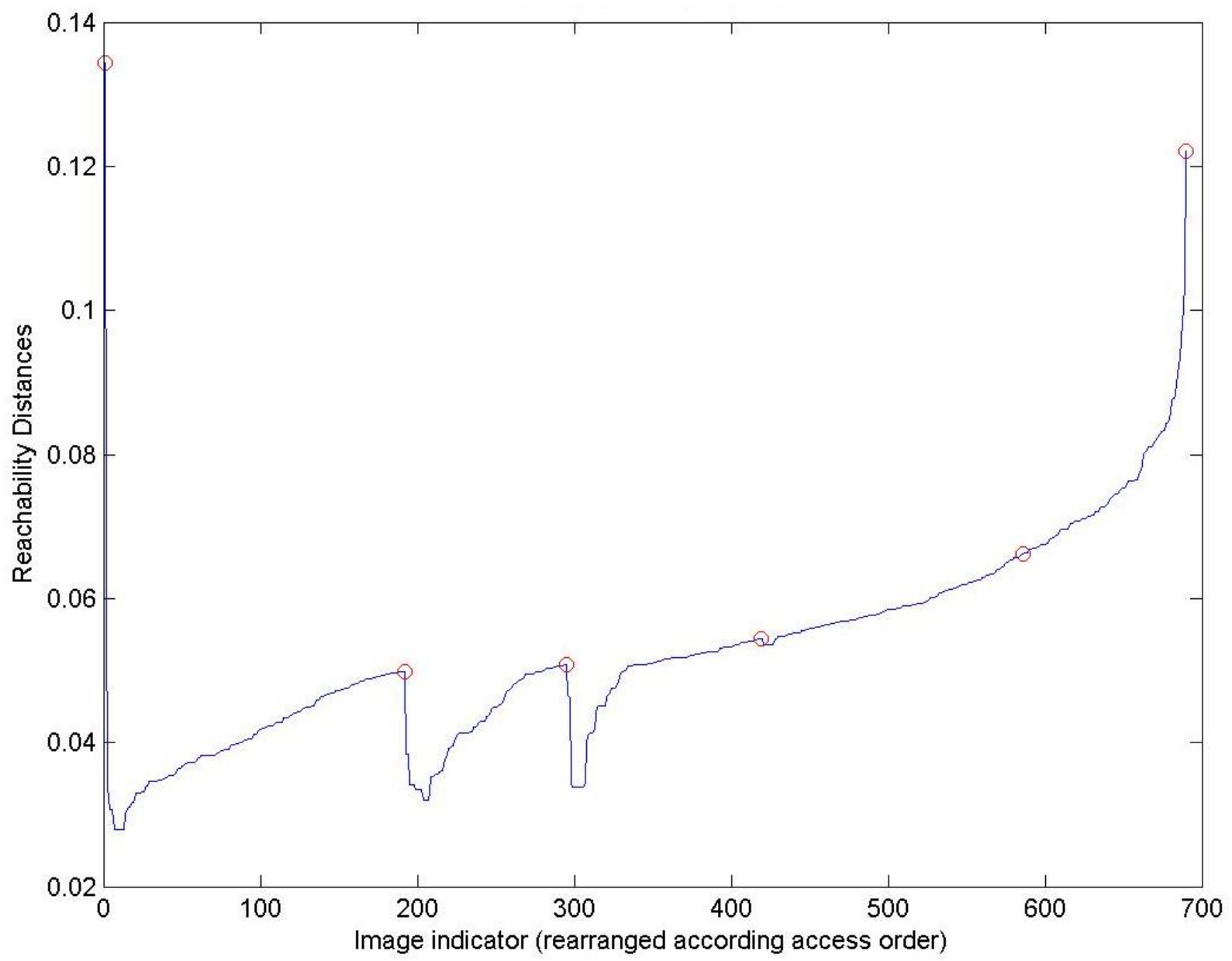

(a)

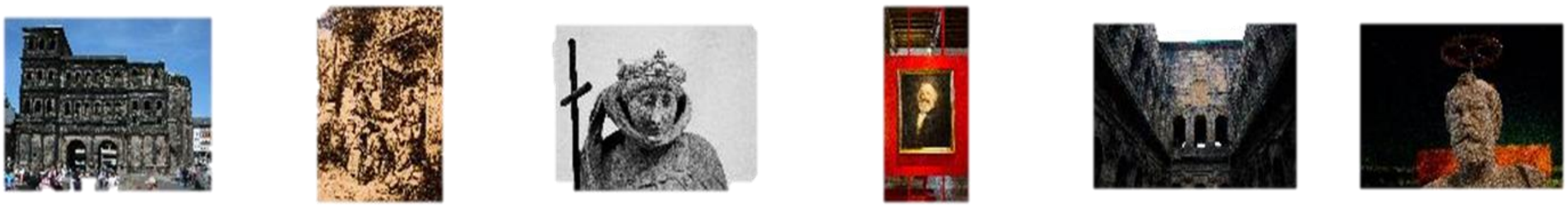

(b)

*Figure 10.1. (a) Optics algorithm results on the original data set of Porta Nigra. Points with 'o' mark the separation between sub-clusters. (b) The corresponding images of the marked points.*

Given the two sets of user defined pairwise constraints, the SDP's solution of eq. (10.5) provides the appropriate distance metric that is used in eq. (3.11). The ranking score sums both accordance and discordance to the user's selection in order to deal with features similarity issue (e.g. color histogram may be the same in two totally different images). Images with ranking close to zero are considered ambiguous, since those are similar to both relevant and non-relevant user defined images.

### 10.2.4 Possible limitations

Generally, it is very difficult for to distinguish among different images (in terms of content) with similar feature vectors. The proposed methodology partially deals with such problems, given that the features used can describe the user's needs. The feature selection is left to the user; there is a variety from low-level (K. Makantasis et al., 2013) to high-level (Makantasis et al., 2014) or combination of both.

However, if data lies in high density manifolds, due to bad feature selection, distance matrix $\boldsymbol{A}$ will ***not*** capture adequately the user's requirements. In the opposite occasion (where features can describe the user's needs), we have a high retrieval rate as described in (Protopapadakis et al., 2014). Finally, it is possible that the representative images do not show what user searches for. In that case, we repeat the process using different image features or different parameters' values for the selection algorithms.

## 10.3 Experimental results

The code has been implemented using MatLab software. The code, for all the utilized methods, is available online. A typical quad-core, 8GB RAM, desktop PC was used. The evaluation data sets were initially obtained using the work of (Ioannides et al., 2013).

### 10.3.1 Dataset description

Evaluation data is specifically build around three cultural monuments; Knossos, Porta Nigra and Fontana dei Quatro Fummi. Knossos is the largest Bronze Age archaeological site on Crete, Greece and is considered Europe's oldest city. The set consists of 1392 images and the special category refers to wall drawings. Porta Nigra (black gate) is a large Roman city gate in Trier, Germany. It is today the largest Roman city gate north of the Alps. The set contains 690 images and the special category refers to interior images. Fontana dei Quatro Fummi (Fountain of the Four Rivers) is a fountain in the Piazza Navona in Rome, Italy. It was designed in 1651 by Gian Lorenzo Bernini for Pope Innocent X. The set contains 133 images and the special category refers to night shots and grayscale images.

For every monument, four cases of image filtering are simulated. The four filtering scenarios can briefly described as: a) need for exterior images of the monument, b) special attributes (depending on the monument), c) people around the monument and d) various images (e.g. animal pictures, night sky, signs, etc.). In every scenario the relevant images are taken from one category and the non-relevant from the rest three in order to construct the pairwise constraints shown in eq. (10.2). In every case, the ratio was 6 relevant to 18 irrelevant. Leading to user feedback of 24 images in total.

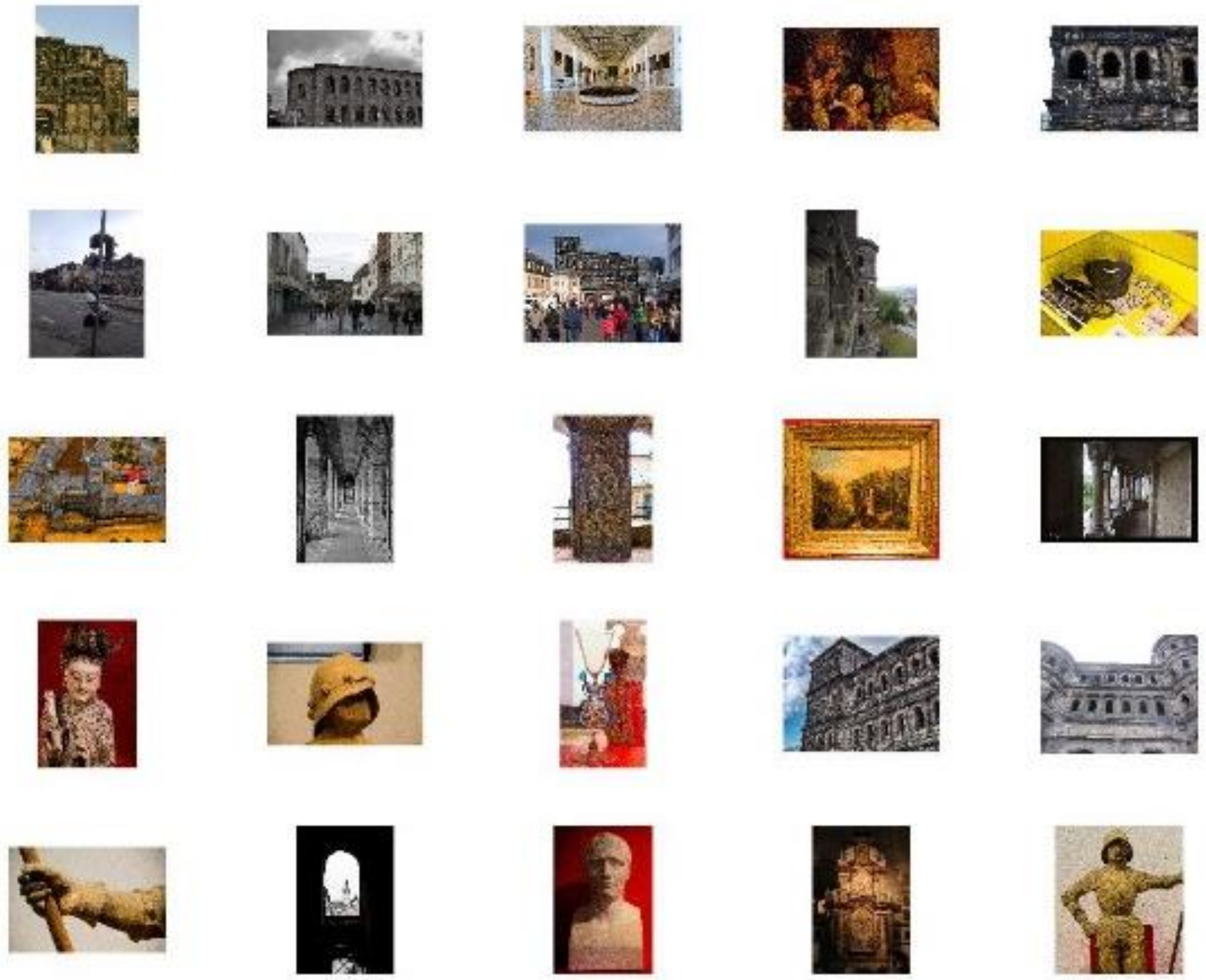

*Figure 10.2. Illustration of the representative objects retrieved, using sparse modeling approach, on Porta Nigra for the 3rd sub-cluster. User may select any number and annotate them as relevant or non-relevant.*

### 10.3.2 Performance scores

Retrieval evaluation, in order to test the filtering capabilities, consist of four scenarios, in three different locations (Table 10.1). It is intriguing that, despite the multiple feature descriptors utilized, a great variance in the content appears; such values suggest low retrieval abilities for the traditional, Euclidean distance based, approaches.

*Table 10.1. Average precision of top ranked images in each of the monuments; 6 images were defined as relevant and 18 as irrelevant through user feedback.*

| | Precision results | Application Scenario | | | | |
|---|---|---|---|---|---|---|
| | | Exterior | Special | People | Various | Overall |
| Monument | Knossos | 0,56 | 0,50 | 0,44 | 0,63 | 0,53 |
| | Porta Nigra | 0,38 | 0,25 | 0,31 | 0,50 | 0,36 |
| | Fontana dei Quatro Fummi | 0,69 | 0,44 | 0,44 | 0,38 | 0,48 |
| | Overall | 0,54 | 0,40 | 0,40 | 0,50 | |

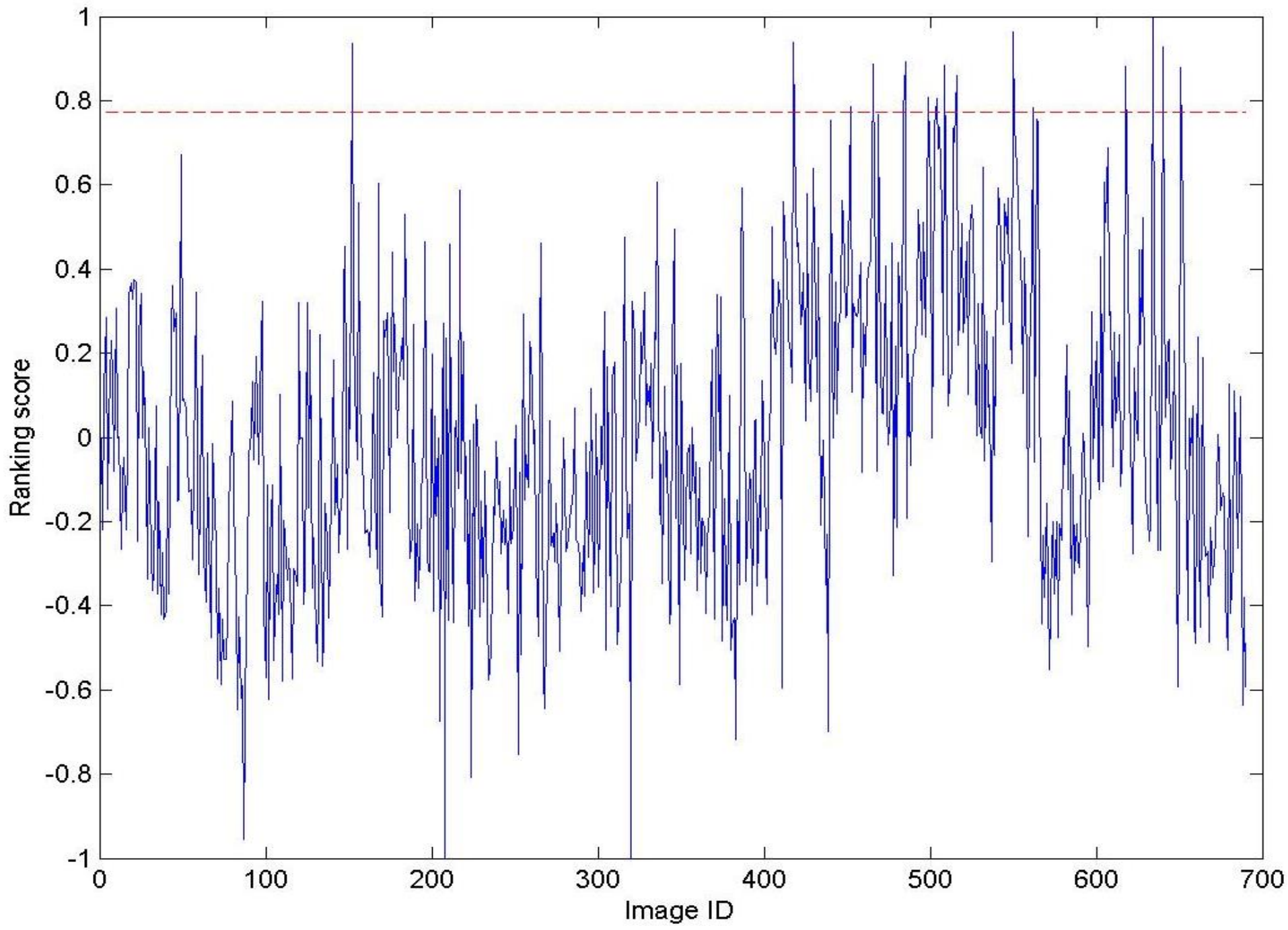

*Figure 10.3. Final ranking score using both accordance and discordance, based on user's selections. A predefined number of images with the higher values are retrieved as relevant. Dashed line indicates minimum value among these images.*

The threshold for the retrieval is set as the minimum ranking value among the *n* images with the higher score. In our case $n = 16$ images. The weight value for an image *i* was defined using a formula of the form:

$$w_i = \frac{r_i}{|R|} \quad (10.6)$$

where $r_i$ denotes *i*-th image ranking among the $|R|$ user-annotated relevant images. The final values are normalized so that $\sum_{i=1}^{|R|} w_i = 1$. The same scheme is employed for the non-relevant images.

Given a set of relevant and non-relevant user defined images (Figure 10.4), the relevant results are defined as the *n* images with higher score over the final ranking (Figure 10.3). The results of retrieval are shown in Figure 10.5.

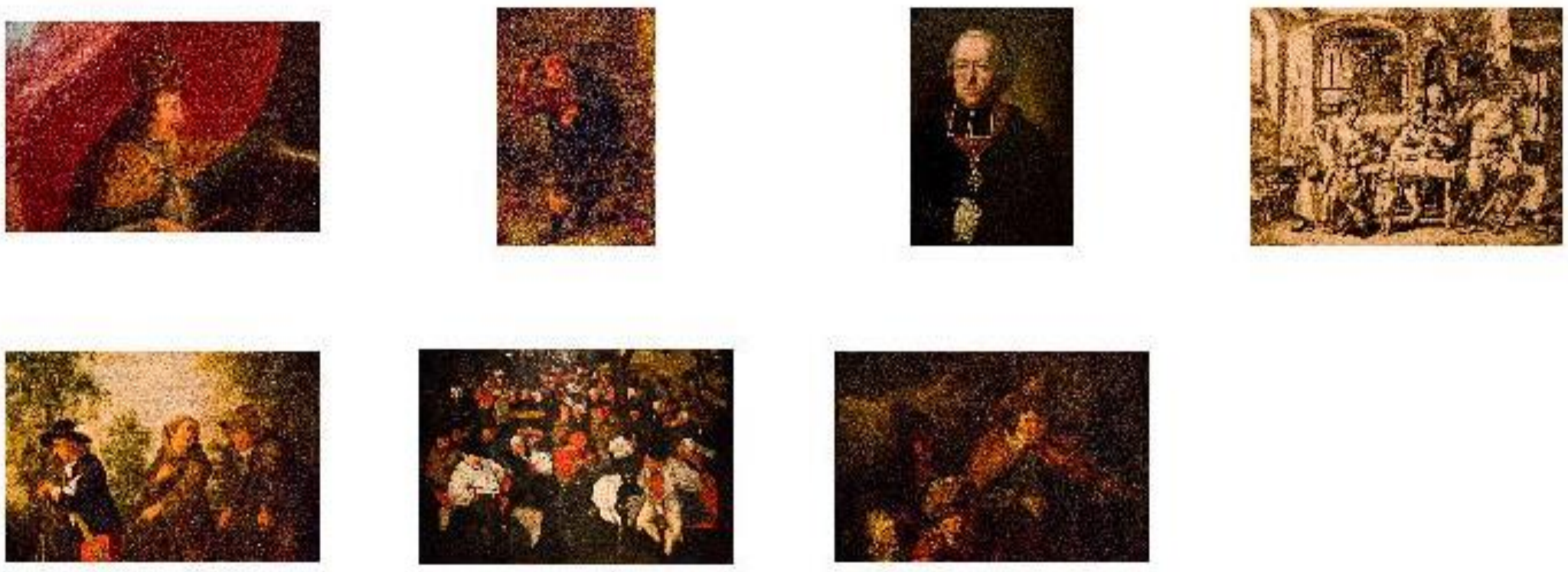

(a)

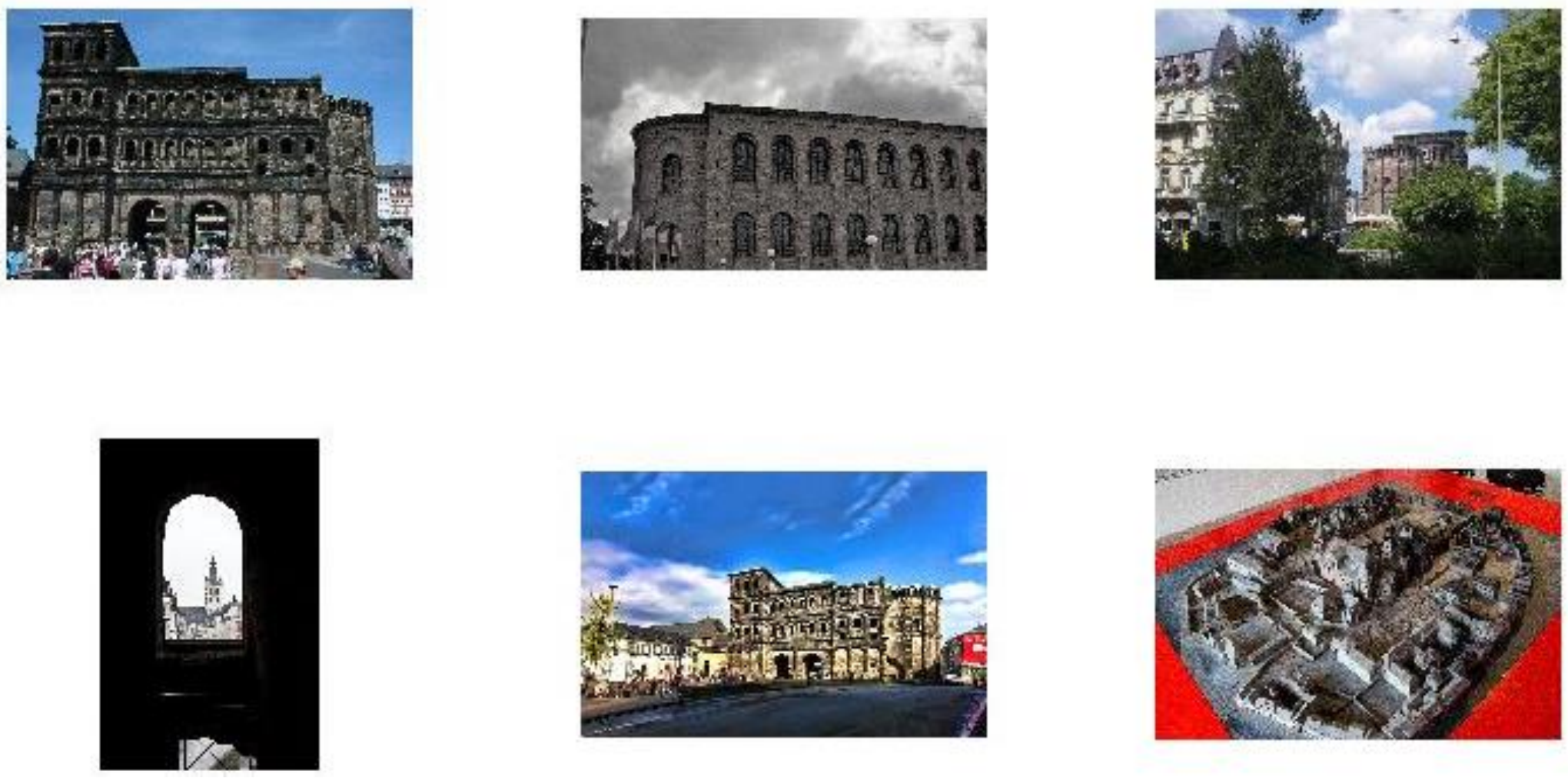

(b)

*Figure 10.4. Illustration of user defined sets of images as relevant (a) to current search results and non-relevant (b), for the Porta Nigra monument. In this case user's interests appears to be around drawings (category: special attributes).*

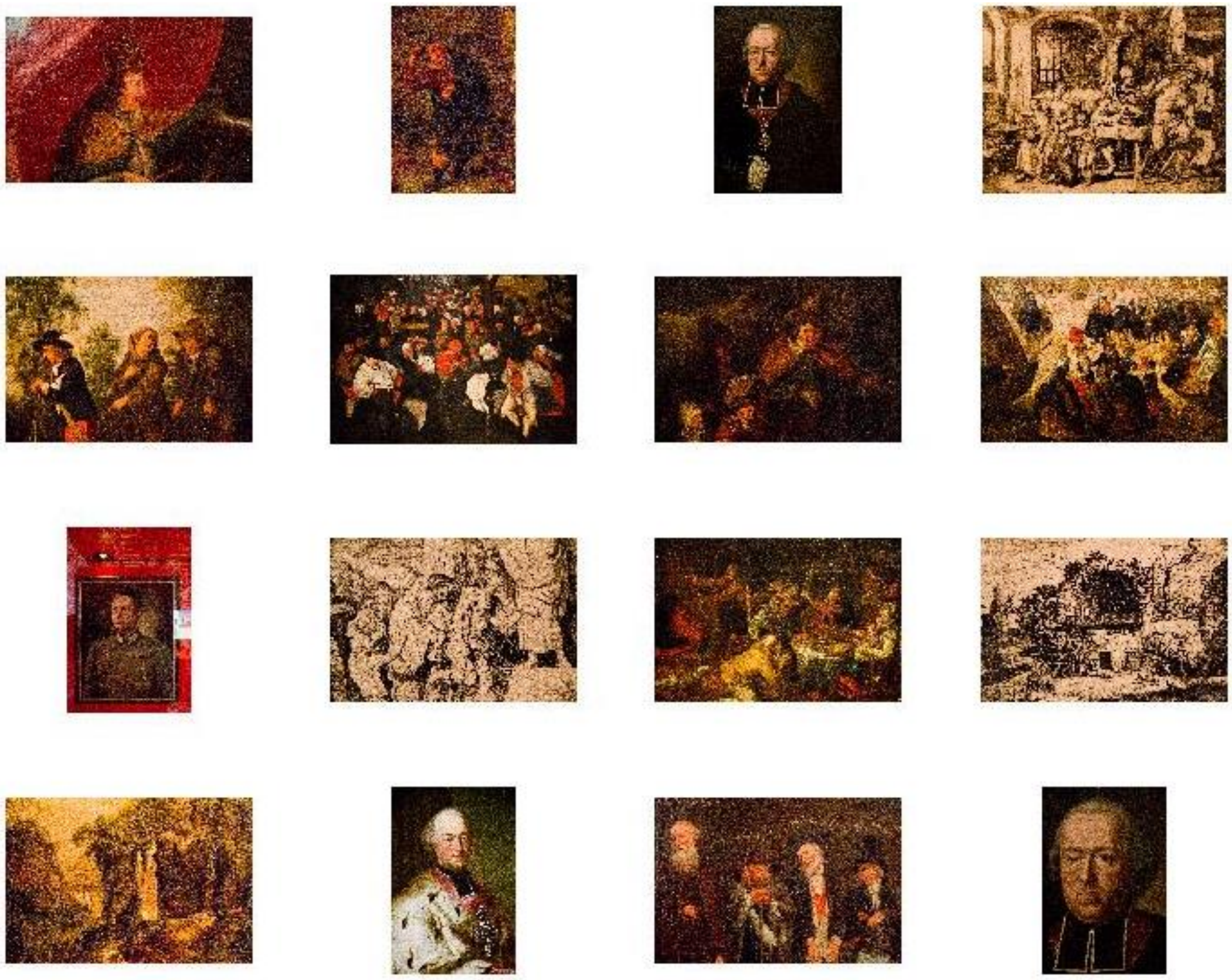

*Figure 10.5. Meta-filtering results given by the system, using the user defined sets of Figure 10.4. It is an extreme case of high retrieval ratio, which is explained mainly due to the distinctive feature values regarding texture.*

## 10.4 Conclusions & future work

In this chapter, a meta-filtering approach, supporting multiple uses of the same vast data sets in Cultural heritage applications, is presented. The problem is formulated as an SDP problem, appropriately adjusted for the field of semi-supervised learning. The system exploits robust, and already tested, techniques in order to support a smooth understanding of user's behavioral patterns.

At first, user selects among the most descriptive data, selected by the system. Then, few of the data are labeled as relevant or irrelevant and are used for the construction of an appropriate distance metric. Once the feedback is concluded, system is able to distinguish the data set's variations and produce results that comply with user's needs.

The impact of the feature vectors is crucial; CLD, SCD and EHD are easily implemented but may be insufficient. Therefore, alternative descriptors should be tested in future approaches. Additionally, future work will emphasize in more sophisticated distance learners in order to capture more complex behavioral patterns of the user.

# *Chapter* XI: In the End

*Life is the art of drawing sufficient conclusions from insufficient premises.*
*Samuel Butler, English poet*

## 11 Concluding remarks

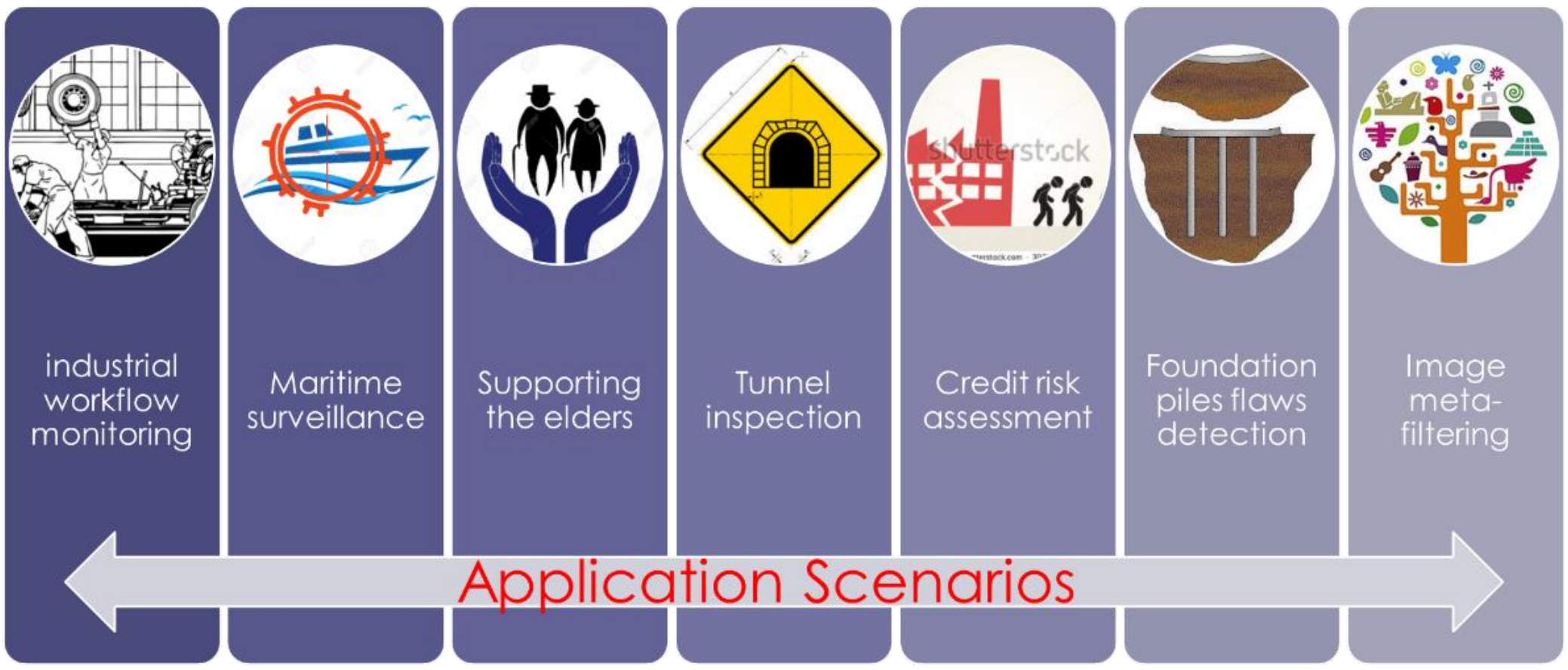

*Figure 11.1. SSL applicability in a wide range of scenarios, demonstrated in this thesis.*

This thesis emphasized on the applicability of SSL techniques in a variety of practical applications (Figure 11.1). The applications fields were: (a) industrial assembly lines monitoring, (b) sea border surveillance, (c) elders' falls detection, (d) transportation tunnels inspection, (e) concrete foundation piles defect recognition, (f) commercial sector companies financial assessment and (g) image advanced filtering for cultural heritage applications.

The main contribution lies in the complex synergistically schemes created from scratch, depending on the application scenario. SSL approaches favor the development of hybrid models; they can be used with, almost, any traditional machine learning approach, facilitating the creation of robust DSSs. Additionally, SSL techniques were used for the system initialization, the detection mechanism formulation, feedback schemes setup, etc. Such advantages are ideal for holistic, user-feedback, information systems. Minimal effort, from user's side is required during trivia steps such as annotation, selection and data labeling. At the same time, minor mistakes can be tolerated by the proposed systems.

The first approach is a hybrid, self-training approach for the industrial workflow recognition. The core mechanism is a topologically optimized feed forward neural network, created using less than 40% of the available data. Then, given new data the classifiers labels them and retrains itself using the most appropriate ones. In order to avoid labeling errors a secondary similarity based classifier is activated, when specific criteria are met. If classifiers did not agree, an expert was summoned via a feedback scheme.

Sea border surveillance is another vision based approach. In that case SSL was used exclusively for training data set creation purposes. The system utilizes a pixel level classification mechanism based on SVMs. In order to correct man made annotation mistakes, which could jeopardize the SVM classifier performance, the initial annotated image data set was re-annotated according to a scalable graph based SSL approach.

Elders' falls detection is an important topic. In this case, the SSL techniques were used for both system initialization and adaptation. At first, under the cluster assumption, extracted feature vectors form two classes: fall and non-fall. Then the classifier is initialized and adopt to the new data using a self-training approach, similar to the industrial monitoring one, previously described. Despite being a vision based approach, no actual video over 3 seconds duration is recorded, preserving the people's privacy.

Applicability over extremely complex data cases was evaluated in tunnel surface defect identification scenarios. It is a typical two class identification problem; SSL approaches could not compete other state of the art techniques in detection performance neither for execution times. There was three major drawbacks: low feature quality, long execution times and hardware requirements. Deep learning hierarchical schemes (i.e. convolutional neural networks) detection abilities outperformed all other approaches.

The structural integrity of foundation piles was also evaluated using graph based approaches. Using wave propagation theory and noise modeling the entire set of foundation piles can be assessed simultaneously. In particular the similarity of the waveforms is projected in a nearest neighbor graph. Then, given the status of few piles, the information propagates through the edges to all connected nodes. The proposed approach reduces labeling effort, which is both costly and time consuming. Such an approach encourages the data sampling, since larger the data base better the classification accuracy.

The labeled data selection impact was evaluated, using data from the Greek commercial sector. A great variety of sampling approaches are used to evaluate the descriptive abilities of small training sets, given a classifier raging from traditional models, e.g. logistic regression, to advanced soft computing techniques, e.g. artificial neural networks. Simulation outcomes suggest that no optimal choice, regarding the data sampling, neither for the classification approach, exists.

Finally, a novel idea for image meta-filtering is proposed. An appropriate distance metric is calculated via SSL assumptions which models user's behavior over image selection. Such an approach allows multiple uses of the same data sets in different applications (e.g. subsets of the same data can be used for 3D reconstruction, tourism promotion, book publications, etc.). User's feedback is involved only for a minor set of data. The proposed scheme facilitates the refinement of retrieval results always under the scope of the end user needs.

Experimental outcomes are in accordance with existing literature suggestions: The unlabeled data can be used in order to improve models' performance (i.e. accuracy, precision, etc.). Yet, there are no optimal solutions regarding the model and feature selection, nor for the data sampling techniques. Heuristic approaches or empiric rules are not always sufficient, although they provide a good starting point during the first stages of a DSS development.

However, there is no "free-lunch". In particular, in real applications the data abundance creates many implementation issues[9]; among them are the memory issue and the time issue. The former directly affects the scalability of the models in datasets with thousands of entries, although there are studies dealing with such an issue. The latter, limit the models applicability in on-line systems (e.g. pixels classification in video sequences). Finally, low feature quality can jeopardize models performance; in the SSL framework bad features result in even more severe performance loss. The model propagates the knowledge to new data and adapt itself to the new wrong labels.

*The end.*

[9] Generally speaking, since the SSL field spans a great variety of methodologies.